\documentclass{article}

\PassOptionsToPackage{numbers}{natbib}

\usepackage[final]{neurips_2023}

\usepackage[utf8]{inputenc} 
\usepackage[T1]{fontenc}    
\usepackage{hyperref}       
\usepackage{url}            
\usepackage{booktabs}       
\usepackage{amsfonts}       
\usepackage{nicefrac}       
\usepackage{microtype}      
\usepackage{xcolor}         
\usepackage{graphicx}
\usepackage{wrapfig}

\usepackage{amsmath}
\usepackage{amssymb}
\usepackage{mathtools}
\usepackage{amsthm}
\usepackage{thmtools, thm-restate}

\usepackage[capitalize,noabbrev]{cleveref}
\usepackage{xspace}
\usepackage{comment}
\usepackage{algorithm}
\usepackage{algpseudocode}
\usepackage{pgfplots}
\usepackage[font=small]{caption}
\usepackage{subcaption}
\usepackage{tikz}
\usepackage{longtable}
\usepackage{adjustbox}
\usetikzlibrary{decorations.pathreplacing,calligraphy}

\theoremstyle{plain}
\newtheorem{theorem}{Theorem}[section]

\newtheorem{lemma}[theorem]{Lemma}

\newtheorem{definition}[theorem]{Definition}
\newtheorem{assumption}[theorem]{Assumption}
\theoremstyle{remark}

\AtBeginDocument{%
 \abovedisplayskip=4pt plus 2pt minus 4pt
 \abovedisplayshortskip=0pt plus 2pt
 \belowdisplayskip=4pt plus 2pt minus 4pt
 \belowdisplayshortskip=2pt plus 2pt minus 2pt
}
\DeclareMathOperator*{\Var}{Var}
\newcommand{\simiid}{\overset{\textrm{i.i.d.}}{\sim}}

\newcommand{\smallparagraph}[1]{\noindent\textbf{#1}\quad}

\algnewcommand{\ParState}[1]{\State%
    \parbox[t]{\dimexpr\linewidth}{\strut\hangindent=\algorithmicindent \hangafter=1 #1\strut}}

\newcommand{\EE}{\mathbb{E}}
\newcommand{\PP}{\mathbb{P}}
\newcommand{\II}{\mathbf{1}}

\newcommand{\algname}{Greedy Over Random Policy\xspace}
\newcommand{\algac}{GORP\xspace}
\newcommand{\datasetname}{\textsc{Bridge}\xspace}
\newcommand{\statesize}{S}
\newcommand{\state}{s}
\newcommand{\statespace}{\mathcal{S}}
\newcommand{\acsize}{A}
\newcommand{\ac}{a}
\newcommand{\acseq}[1]{\vec{\ac}}
\newcommand{\acspace}{\mathcal{A}}
\newcommand{\eptime}{t}
\newcommand{\horizon}{T}
\newcommand{\discount}{\gamma}
\newcommand{\qdist}{\mathcal{D}}
\newcommand{\efhorizon}[1]{H_{#1}}
\newcommand{\efhorizonbound}[1]{\bar{H}_{#1}}
\newcommand{\covlen}{L}
\newcommand{\epw}{W}
\newcommand{\dynamics}{f}

\newcommand{\reward}{R}
\newcommand{\policy}{\pi}
\newcommand{\policies}{\Pi}
\newcommand{\expolicy}{\policy^\text{expl}}
\newcommand{\randpolicy}{\policy^\text{rand}}
\newcommand{\pret}{J}
\newcommand{\gap}{\Delta}
\newcommand{\samples}{n}
\newcommand{\sampcomplexity}{N}
\newcommand{\numep}[1]{m_{#1}}
\newcommand{\numqvi}{k}
\newcommand{\qvi}{\text{QVI}}
\newcommand{\mdp}{\mathcal{M}}

\title{Bridging Reinforcement Learning Theory \\and Practice with the Effective Horizon}

\author{%
    Cassidy Laidlaw \qquad Stuart Russell \qquad Anca Dragan \\
    Unversity of California, Berkeley \\
    \texttt{\{cassidy\_laidlaw,russell,anca\}@cs.berkeley.edu} \\
}

\begin{document}

\maketitle

\begin{abstract}
Deep reinforcement learning (RL) works impressively in some environments and fails catastrophically in others. Ideally, RL theory should be able to provide an understanding of why this is, i.e. bounds predictive of practical performance. Unfortunately, current theory does not quite have this ability. We compare standard deep RL algorithms to prior sample complexity bounds by introducing a new dataset, \datasetname. It consists of 155 deterministic MDPs from common deep RL benchmarks, along with their corresponding tabular representations, which enables us to exactly compute instance-dependent bounds. We choose to focus on deterministic environments because they share many interesting properties of stochastic environments, but are easier to analyze. Using \datasetname, we find that prior bounds do not correlate well with when deep RL succeeds vs. fails, but discover a surprising property that does. When actions with the highest Q-values under the \emph{random} policy also have the highest Q-values under the \emph{optimal} policy (i.e. when it is optimal to be greedy on the random policy's Q function), deep RL tends to succeed; when they don't, deep RL tends to fail. We generalize this property into a new complexity measure of an MDP that we call the \emph{effective horizon}, which roughly corresponds to how many steps of lookahead search would be needed in that MDP in order to identify the next optimal action, when leaf nodes are evaluated with random rollouts. Using \datasetname, we show that the effective horizon-based bounds are more closely reflective of the empirical performance of PPO and DQN than prior sample complexity bounds across four metrics. We also find that, unlike existing bounds, the effective horizon can predict the effects of using reward shaping or a pre-trained exploration policy.
Our code and data are available at \url{https://github.com/cassidylaidlaw/effective-horizon}.
\end{abstract}

\section{Introduction}
\label{sec:intro}

\begin{wrapfigure}{R}{3in}
    \centering
    \vspace{-36pt}
    \begin{minipage}[b]{0.45\linewidth}
        \begin{tikzpicture}[auto,node distance=8mm,scale=0.7,>=latex,font=\small]
            \tikzstyle{state}=[thick,draw=black,circle,fill=white]
        
            \node[state, label=Current state] at (0, 0) (s1) {};
            \node[state] at (-1, -1) (s2_1) {};
            \draw[->] (s1) -- node[above left] {} (s2_1);
            \node[state] at (1, -1) (s2_2) {};
            \draw[->] (s1) -- node[above right] {} (s2_2);

            \draw[->] plot [smooth] coordinates { (-1.5, -3) (-1.4, -3.2) (-1.5, -3.4) (-1.4, -3.6) (-1.5, -3.8) (-1.4, -4) };
            \draw[->] plot [smooth] coordinates { (-1.5, -3) (-1.7, -3.2) (-1.6, -3.35) (-1.8, -3.5) (-1.7, -3.65) (-1.9, -3.8) };
            \draw[->] plot [smooth] coordinates { (-1.5, -3) (-1.3, -3.2) (-1.3, -3.35) (-1.1, -3.5) (-1.1, -3.65) (-0.9, -3.8) };

            \node[state] at (-1.5, -3) (sT_1) {};
            \draw[->] (s2_1) -- node[left] {} (sT_1);
            \node[state] at (-0.5, -3) (sT_2) {};
            \draw[->] (s2_1) -- node[right] {} (sT_2);
            \node[state] at (0.5, -3) (sT_3) {};
            \draw[->] (s2_2) -- node[left] {} (sT_3);
            \node[state] at (1.5, -3) (sT_4) {};
            \draw[->] (s2_2) -- node[right] {} (sT_4);
            
            \node[fill=white] at (-1, -2) (dots_1) {$\dots$};
            \node[fill=white] at (1, -2) (dots_2) {$\dots$};

            \draw [decorate, decoration = {brace}, thick] (-2, -3) --  (-2, -1)
                node[pos=0.5, left=2pt, rotate=90, anchor=south]{$\numqvi$ timesteps};
                
            \draw [decorate, decoration = {brace}, thick] (-0.9, -4.1) -- (-2, -4.1)
                node[pos=0.5, right=10pt, anchor=west, align=center]{$\numep{}$ random rollouts \\ from each leaf node};
        \end{tikzpicture}
    \end{minipage}
    \hfill
    \begin{minipage}[b]{0.45\linewidth}
        \input{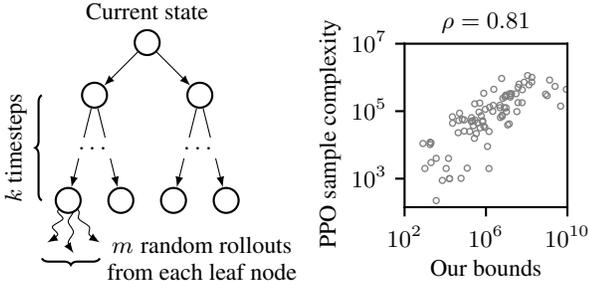}
    \end{minipage}
    \caption{We introduce the effective horizon, a property of MDPs that controls how difficult RL is. Our analysis is motivated by \algname (\algac), a simple Monte Carlo planning algorithm (left) that exhaustively explores action sequences of length $\numqvi$ and then uses $\numep{}$ random rollouts to evaluate each leaf node. The effective horizon combines both $\numqvi$ and $\numep{}$ into a single measure. We prove sample complexity bounds based on the effective horizon that correlate closely with the real performance of PPO, a deep RL algorithm, on our \datasetname dataset of 155 deterministic MDPs (right).}
    \label{fig:overview}
\end{wrapfigure}

Deep reinforcement learning (RL) has produced impressive results in robotics \citep{levine_end--end_2016}, strategic games \citep{silver_mastering_2016}, and control \citep{mnih_human-level_2015}. However, the same deep RL algorithms that achieve superhuman performance in some environments completely fail to learn in others. Sometimes, using techniques like reward shaping or pre-training help RL, and in other cases they don't. Our goal  is to provide a theoretical understanding of why this is---a theoretical analysis that is \emph{predictive} of practical RL performance.

Unfortunately, there is a large gap between the current theory and practice of RL.
Despite RL theorists often focusing on algorithms using strategic exploration (e.g., UCB exploration bonuses; \citet{azar_minimax_2017,jin_is_2018}), the most commonly-used deep RL algorithms, which explore randomly, resist such analysis.
In fact, theory suggests that RL with random exploration is exponentially hard in the worst case \citep{dann_guarantees_2022}, but this is not predictive of practical performance. 
Some theoretical research has explored instance-dependent bounds, identifying properties of MDPs when random exploration should perform better than this worst case \citep{liu_when_2019,malik_sample_2021}. However, it is not clear whether these properties correlate with when RL algorithms work vs. fail---and our results will reveal that they tend not to.

If the current theory literature cannot explain the empirical performance of deep RL, what can? Ideally, a theory of RL should provably show why deep RL succeeds while using random exploration. It should also be able to predict which environments are harder or easier to solve empirically. Finally, it should be able to explain when and why tools like reward shaping or initializing with a pre-trained policy help make RL perform better.

We present a new theoretical complexity measure for MDPs called the \emph{effective horizon} that satisfies all of the above criteria. Intuitively, the effective horizon measures approximately how far ahead an algorithm must exhaustively plan in an environment before evaluating leaf nodes with random rollouts.

In order to assess previous bounds and eventually arrive at such a property, we start by creating a new dataset, \datasetname, of deterministic MDPs from common deep RL benchmarks. A major difficulty with evaluating instance-dependent bounds is that they can't be calculated without tabular representations, so prior work work has typically relied on small toy environments for justification. To get a more realistic picture, we choose 155 MDPs across different benchmarks and compute their tabular representations---some with over 100 million states which must be exhaustively explored and stored. This is a massive engineeering challenge, but it enables connecting theoretical and empirical results at an unprecedented scale. We focus on deterministic MDPs in \datasetname and in this paper because they are simpler to analyze but still have many of the interesting properties of stochastic MDPs, like reward sparsity and credit assignment challenges. Many deep RL benchmarks are (nearly) deterministic, so we believe our analysis is highly relevent to practical RL.

Our journey to the effective horizon began with identifying a surprising property that holds in many of the environments in \datasetname: one can learn to act \emph{optimally} by acting \emph{randomly}.
More specifically, actions with the highest Q-values under the uniformly \emph{random} policy \emph{also} have the highest Q-values under the \emph{optimal} policy. The random policy is about as far as one can get from the optimal policy, so this property may seem unlikely to hold. However, about two-thirds of the environments in \datasetname satisfy the property. This proportion rises to four-fifths among environments that PPO \cite{schulman_proximal_2017}, a popular deep RL algorithm, can solve efficiently (Table \ref{tab:k_eq_1}). Conversely, when this property does not hold, PPO is more likely to fail than succeed---and when it does succeed, so does simply applying a few steps of lookahead on the Q-function of the random policy (Figure \ref{fig:numqvi_dist}).
We found it remarkable that, at least in the environments in \datasetname, modern algorithms seem to boil down to not much more than acting greedily on the random policy Q-values.

The property that it is optimal to act greedily with respect to the random policy's Q-function has important implications for RL theory and practice. Practically, it suggests that very simple algorithms designed to estimate the random policy's Q-function could efficiently find an optimal policy. We introduce such an algorithm, \algname (\algac), which also works in the case where one may need to apply a few steps of value iteration to the random policy's Q-function before acting greedily. Empirically, \algac finds an optimal policy in fewer timesteps than DQN (another deep RL algorithm) in more than half the environments in \datasetname. Theoretically, it is simple to analyze \algac, which consists almost entirely of estimating the random policy's Q-function via a sample average over i.i.d. random rollouts. Since \algac works well empirically and can be easily understood theoretically, we thoroughly analyze it in the hopes of finding sample complexity bounds that can explain the performance of deep RL.

Our analysis of \algname leads to a single metric, the effective horizon, that measures the complexity of model-free RL in an MDP. As shown in Figure \ref{fig:overview}, \algac is an adaptation of a Monte Carlo planning algorithm to the reinforcement learning setup (where the transitions are unknown): it mimics exhaustively planning ahead $\numqvi$ steps and then sampling $\numep{}$ random rollouts from each leaf node. The effective horizon $\efhorizon{}$ combines the depth $\numqvi$ and number of rollouts $\numep{}$. We call it the \emph{effective} horizon because worst-case sample complexity bounds for random exploration are exponential in the horizon $\horizon$, while we prove sample complexity bounds exponential only in the effective horizon $\efhorizon{}$.
For most \datasetname environments, $\efhorizon{} \ll \horizon$, explaining the efficiency of RL in these MDPs.

In the environments in \datasetname, we find that the effective horizon-based sample complexity bounds satisfy all our desiderata above for a theory of RL. They are more predictive of the empirical sample complexities of PPO and DQN than several other bounds from the literature across four metrics, including measures of correlation, tightness, and accuracy (Table \ref{tab:bound_evaluations}). 
Furthermore, the effective horizon can predict the effects of both reward shaping (Table \ref{tab:reward_shaping}) and initializing using a pre-trained policy learned from human data or transferred from a similar environment (Table \ref{tab:exploration_policy}). In contrast, none of the existing bounds we compare to depend on both the reward function and initial policy; thus, they are unable to explain why reward shaping, human data, and transfer learning can help RL.
Although our results focus on deterministic MDPs, we plan to extend our work to stochastic environments in the future and already have some promising results in that direction.

\section{Preliminaries}
\label{sec:preliminaries}

We begin by presenting the reinforcement learning (RL) setting we consider. An RL algorithm acts in a deterministic, tabular, episodic Markov decision process (MDP) with finite horizon. The MDP comprises a set of states $\statespace$, a set of actions $\acspace$, a horizon $\horizon \in \mathbb{N}$ and optional discount factor $\discount \in [0, 1]$, a start state $\state_1$, transition function $\dynamics: \statespace \times \acspace \to \statespace$, and a reward function $\reward: \statespace \times \acspace \to \mathbb{R}$. Throughout the paper we use $\discount = 1$ but all our theory applies equally when $\discount < 1$.

An RL agent interacts with the MDP for a number of episodes, starting at a fixed start state $\state_1$. At each step $\eptime \in [\horizon]$ of an episode (using the notation $[n] = \{1, \hdots, n\}$), the agent observes the state $\state_\eptime$, picks an action $\ac_\eptime$, receives reward $\reward(\state_\eptime, \ac_\eptime)$, and transitions to the next state $\state_{\eptime + 1} = \dynamics(\state_\eptime, \ac_\eptime)$. A policy $\policy$ is a set of functions $\policy_1, \hdots, \policy_{\eptime}: \statespace \to \Delta(\acspace)$, which defines for each state and timestep a distribution $\policy_\eptime(\ac \mid \state)$ over actions. If a policy is deterministic at some state, then with slight abuse of notation we denote $\ac = \policy_\eptime(\state)$ to be the action taken by $\policy_\eptime$ in state $\state$.


We denote a policy's Q-function $Q^\policy_\eptime: \statespace \times \acspace \to \mathbb{R}$ and value function $V^\policy_\eptime: \statespace \to \mathbb{R}$ for each $\eptime \in [\horizon]$.
In this paper, we also use a Q-function which is generalized to \emph{sequences} of actions. We use the shorthand $\ac_{\eptime : \eptime + k}$ to denote the sequence $\ac_\eptime, \hdots, \ac_{\eptime + k}$, and define the action-sequence Q-function as
\begin{align*}
    \textstyle Q^\policy_\eptime(\state_\eptime, \ac_{\eptime : \eptime + k}) & = \EE_\policy \left[  \sum_{\eptime' = \eptime}^\horizon \discount^{\eptime' - \eptime} \: \reward(\state_{\eptime'}, \ac_{\eptime'}) \mid \state_\eptime , \ac_{\eptime : \eptime + k} \right].
\end{align*}

The objective of an RL algorithm is to find an optimal policy $\policy^*$, which maximizes $\textstyle \pret(\policy) = V^\policy_1(\state_1)$, the expected discounted sum of rewards over an episode, also known as the return of the policy $\policy$.

Generally, an RL algorithm can be run for any number of timesteps $\samples$ (i.e., counting one episode as $\horizon$ timesteps), returning a policy $\policy^\samples$. We define the $\emph{sample complexity}$ $\sampcomplexity$ of an RL algorithm as the minimum number of timesteps needed such that the algorithm has at least a 50-50 chance of returning an optimal policy:
\begin{equation*}
  \textstyle \sampcomplexity = \min \left\{ \samples \in \mathbb{N} \mid \PP\left(\pret(\policy^\samples) = \pret^*\right) \geq 1/2 \right\}.
\end{equation*}
Here, the probability is with respect to any randomness in the algorithm itself. One can estimate the sample complexity $\sampcomplexity$ empirically by running an algorithm several times, calculating the number of samples $\samples$ needed to reach the optimal policy during each run, and then taking the median.


The following simple theorem gives upper and lower bounds for the worst-case sample complexity in a deterministic MDP, depending on $\acsize$ and $\horizon$.

\begin{restatable}{theorem}{thmdeterministicwc}
\label{thm:deterministicwc}
There is an RL algorithm which can solve any deterministic MDP with sample complexity $\sampcomplexity \leq \horizon \lceil \acsize^\horizon / 2 \rceil$. Conversely, for any RL algorithm and any values of $\horizon$ and $\acsize$, there must be some deterministic MDP for which its sample complexity $\sampcomplexity \geq \horizon (\lceil \acsize^\horizon / 2 \rceil - 1)$.
\end{restatable}

All proofs are deferred to Appendix \ref{sec:proofs}. In this case, the idea of the proof is quite simple, and will later be useful to motivate our idea of the effective horizon: in an MDP where exactly one sequence of actions leads to a reward, an RL algorithm may have to try almost every sequence of actions to find the optimal policy; there are $\acsize^\horizon$ such sequences. As we develop sample complexity bounds based on the effective horizon in Section \ref{sec:efhorizon}, we can compare them to the worst-case bounds in Theorem \ref{thm:deterministicwc}.

\smallparagraph{Why deterministic MDPs?} We focus on deterministic (as opposed to stochastic) MDPs in this study for several reasons.
First, analyzing deterministic MDPs avoids the need to consider generalization within RL algorithms. In common stochastic MDPs, one often needs neural-network based policies, whereas in a deterministic MDP one can simply learn a sequence of actions. Since neural network generalization is not well understood even in supervised learning, analyzing generalization in RL is an especially difficult task. Second, deterministic MDPs still display many of the interesting properties of stochastic MDPs. For instance, deterministic MDPs have worst case exponential sample complexity when using naive exploration; environments with dense rewards are easier to solve empirically than those with sparse rewards; credit assignment can be challenging; and there is a wide range of how tractable environments are for deep RL, even for environments with similar horizons, state spaces, and action spaces.

Finally, many common RL benchmark environments are deterministic or nearly-deterministic. For instance, the ALE Atari games used to evaluate DQN \citep{mnih_human-level_2015}, Rainbow \citep{hessel_rainbow_2017}, and MuZero \citep{schrittwieser_mastering_2020} are all deterministic after the first state, which is selected randomly from one of only 30 start states. The widely used DeepMind Control Suite \citep{tassa_deepmind_2018} is based on the MuJoCo simulator \citep{todorov_mujoco_2012}, which is also deterministic given the initial state (some of the environments do use a randomized start state). MiniGrid environments \citep{chevalier-boisvert_minimalistic_2018}, which are commonly used for evaluating exploration \citep{seo_state_2021}, environment design \citep{dennis_emergent_2020}, and language understanding \citep{chevalier-boisvert_babyai_2019}, are also deterministic after the initial state. Thus, our investigation of deterministic environments is highly relevant to common deep RL practice.

\section{Related Work}
\label{sec:related_work}

Before delving into our contributions, we briefly summarize existing work in theoretical RL and prior sample complexity bounds. Our novel bounds are contrasted with existing ones in Sections \ref{sec:bound_comparisons} and \ref{sec:experiments}; for a detailed comparison with full definitions and proofs, please see Appendix \ref{sec:more_related_work}.

Recent RL theoretical results largely focus on strategic exploration using techniques like UCB exploration bonuses \citep{kakade_sample_2003,azar_minimax_2017,jiang_contextual_2017,jin_is_2018,jin_provably_2019,du_bilinear_2021,jin_bellman_2021}. Such bounds suggest RL is tractable for smaller or low-dimensional state spaces. In deterministic MDPs, the UCB-based \textsc{R-max} algorithm \citep{brafman_r-max_2002,kakade_sample_2003} has sample complexity bounded by $\statesize \acsize \horizon$.

Some prior work has focused on sample complexity bounds for random exploration. \citet{liu_when_2019} give bounds based on the covering length $\covlen$ of an MDP, which is the number of episodes needed to visit all state-action pairs at least once with probability at least $1/2$ while taking actions at random.
This yields a sample complexity bound of $\horizon \covlen$ for deterministic MDPs. Other work suggests that it may not be necessary to consider rewards all the way to the end of the episode to select an optimal action \citep{kearns_sparse_2002,jiang_structural_2016,malik_sample_2021}. One can define a ``effective planning window'' of $\epw$ timesteps ahead that must be considered, resulting in a sample complexity bound of $\horizon^2 \acsize^\epw$ for deterministic MDPs. Finally, \citet{dann_guarantees_2022} define a ``myopic exploration gap'' that controls the sample complexity of using $\epsilon$-greedy exploration, a form of naive exploration. However, in Appendix \ref{sec:other_bounds}, we demonstrate why their bounds are impractical and often vacuous.

There have been a few prior attempts to bridge the RL theory-practice gap. bsuite \citep{osband_behaviour_2020}, MDP playground \citep{rajan_mdp_2021}, and SEGAR \citep{hjelm_sandbox_2022} are collections of environments that are designed to empirically evaluate deep RL algorithms across various axes of environment difficulty. However, they do not provide theoretical explanations for why environments with varying properties are actually easier or harder. Furthermore, their environments are artificially constructed to have understandable properties. In contrast, we aim to find the mathematical reasons that deep RL succeeds or fails in ``in-the-wild'' environments like Atari and Procgen. \citet{conserva_hardness_2022} calculate two regret bounds and compare the bounds to the empirical performance of RL algorithms. However, they consider tabular RL algorithms in simple artificial environments with less than a thousand states, while we experiment with deep RL algorithms on real benchmark environments with tens of millions of states.

Our \algac algorithm and the effective horizon are inspired by rollout and Monte Carlo planning algorithms, which have a long history \citep{abramson_expected-outcome_1990,brugmann_monte_1993,tesauro_-line_1996,bertsekas_rollout_1997,kearns_sparse_2002,chang_adaptive_2005,kocsis_bandit_2006,coulom_efficient_2007}. These algorithms were used effectively in Backgammon \citep{tesauro_-line_1996}, Go \citep{bouzy_monte-carlo_2004}, and real-time strategy games \citep{chung_monte_2005} before the start of deep RL.
\algac and related Monte Carlo rollout algorithms are sometimes referred to as ``one-step'' or ``multi-step'' lookahead. \citet{bertsekas_rollout_2020,bertsekas_lessons_2022} suggests that one-step lookahead, possibly after a few steps of value iteration, often leads to fast convergence to optimal policies because it is equivalent to a step of Newton's method for finding a fixed-point of the Bellman equation \citep{kleinman_iterative_1968,puterman_convergence_1979}.
Our analysis suggests the success of deep RL is due to similar properties. However, we go beyond previous work by introducing \algac, which approximates a step of policy iteration in model-free RL---a setting where on-line planning approaches are not applicable. Furthermore, unlike previous work in Monte-Carlo planning, we combine our theoretical contributions with extensive empirical analysis to verify that our assumptions hold in common environments.

\begin{wraptable}{R}{2.5in}
    \centering
    \small
    \begin{tabular}{l|rr}
        \toprule
        Acting greedily & \multicolumn{2}{|c}{PPO finds optimal policy} \\
        with respect to & \multicolumn{2}{|c}{in $\leq 5\text{M}$ timesteps?} \\
        $Q^{\randpolicy}$ is optimal? & Yes & No \\
        \midrule
        Yes & \bf 80 MDPs & 24 MDPs \\
        No & 15 MDPs & \bf 36 MDPs \\
        \bottomrule
    \end{tabular}
    \caption{The distribution of the MDPs in our \datasetname dataset according to two criteria: first, whether PPO empirically converges to an optimal policy in 5 million timesteps, and second, whether acting greedily with respect to the Q-function of the random policy is optimal. We find that a surprising number of environments satisfy the latter property, especially when only considering those where PPO succeeds.}
    \label{tab:k_eq_1}
\end{wraptable}

\section{The \datasetname Dataset}
\label{sec:dataset}


In order to assess how well existing bounds predict practical performance, and gain insight about novel properties of MDPs that could be predictive, we constructed \datasetname (Bridging the RL Interdisciplinary Divide with Grounded Environments), a dataset of 155 popular deep RL benchmark environments with full tabular representations. One might assume that the ability to calculate the instance-dependent bounds we just presented in Section \ref{sec:related_work} already exists; however, it turns out that for many real environments even the number of states $\statesize$ is unknown! This is because a significant engineering effort is required to analyze large-scale environments and calculate their properties.

In \datasetname, we tackle this problem by computing tabular representations for all the environments using a program that exhaustively enumerates all states, calculating the reward and transition functions at every state-action pair. We do this for 67 Atari games from the Arcade Learning Enivornment \citep{bellemare_arcade_2013}, 55 levels from the Procgen Benchmark \citep{cobbe_leveraging_2020}, and 33 gridworlds from MiniGrid \citep{chevalier-boisvert_minimalistic_2018} (Figure \ref{fig:dataset_overview}).
The MDPs have state space sizes $\statesize$ ranging across 7 orders of magnitude from tens to tens of millions, 3 to 18 discrete actions, and horizons $\horizon$ ranging from 10 to 200, which are limited in some cases to avoid the state space becoming too large. See Appendix \ref{sec:dataset_details} for the full details of the \datasetname dataset.

\smallparagraph{A surprisingly common property}
To motivate the effective horizon, which is introduced in the next section, we describe a property that we find holds in many of the MDPs in \datasetname. Consider the random policy $\randpolicy$, which assigns equal probability to every action in every state, i.e., $\randpolicy_\eptime(\ac \mid \state) = 1 / \acsize$. We can use dynamic programming on a tabular MDP to calculate the random policy's Q-function $Q^{\randpolicy}$. We denote by $\policies(Q^{\randpolicy})$ the set of policies which act greedily with respect to this Q-function; that is,
\begin{equation*}
    \policies\left(Q^{\randpolicy}\right) = \Big\{\policy \mid \forall \state, \eptime \quad \policy_\eptime(\state) \in \arg \max_{\ac \in \acspace} Q^{\randpolicy}_\eptime(\state, \ac) \Big\}.
\end{equation*}
Perhaps surprisingly, we find that all the policies in $\policies(Q^{\randpolicy})$ are \emph{optimal} in about two-thirds of the MDPs in \datasetname. This proportion is even higher when considering only the environments where PPO empirically succeeds in finding an optimal policy (Table \ref{tab:k_eq_1}). Thus, it seems that this property may be the key to what makes many of these environments tractable for deep RL.


\begin{wrapfigure}{R}{3in}
\vspace{-12pt}
\begin{minipage}{3in}
\begin{algorithm}[H]
    \small
    \caption{The \algname (\algac) algorithm, used to motivate the effective horizon.}
    \begin{algorithmic}[1]
    \Procedure{\algac}{$\numqvi, \numep{}, \expolicy$}
        \For{$i = 1, \hdots, \horizon$}
            \For{$\ac_{i : i + \numqvi - 1} \in \acspace^\numqvi$} \label{line:qirl_action_loop}
                \ParState{
                    sample $\numep{}$ episodes following $\policy_1, \hdots, \policy_{i - 1}$,  \\
                    then actions $\ac_{i : i + \numqvi - 1}$, and finally $\expolicy$.
                }
                \ParState{
                    $\hat{Q}_i(\state_i, \ac_{i : i + \numqvi - 1}) \gets$ \\
                    $\frac{1}{\numep{}} \sum_{j = 1}^{\numep{}} \sum_{\eptime = i}^\horizon \discount^{\eptime - i} \reward(\state_\eptime^j, \ac_\eptime^j)$.
                }
            \EndFor \label{line:qirl_action_loop_end}
            \ParState{
                \label{line:max_action_seq}
                $\policy_i(\state_i) \gets \arg \max_{\ac_i \in \acspace}$ \\
                $\max_{\ac_{i + 1 : i + \numqvi - 1} \in \acspace^{\numqvi - 1}} \hat{Q}_i(\state_i, \ac_i, \ac_{i + 1 : i + \numqvi - 1}).$
            } 
        \EndFor
		\State \Return $\policy$
    \EndProcedure
    \end{algorithmic}
    \label{alg:qirl}
\end{algorithm}
\end{minipage}
\vspace{-12pt}
\end{wrapfigure}

\section{The Effective Horizon}
\label{sec:efhorizon}

We now theoretically analyze why RL should be tractable in environments where, as we observe in \datasetname, it is optimal to act greedily with respect to the random policy's Q-function. This leads to a more general measure of an environment's complexity for model-free RL: the effective horizon.

Our analysis centers around a simple algorithm, \algac (\algname), shown in Algorithm \ref{alg:qirl}. \algac constructs an optimal policy iteratively; each iteration $i$ aims to calculate an optimal policy $\policy_i$ for timestep $\eptime = i$. 
In the case where we set $\numqvi = 1$ and $\expolicy = \randpolicy$, \algac can solve environments which have the property we observe in \datasetname. It does this at each iteration $i$ by simulating $\numep{}$ random rollouts for each action from the state reached at timestep $\eptime = i$. Then, it averages the $\numep{}$ rollouts' returns to obtain a Monte Carlo estimate of $Q^{\randpolicy}$ for each action. Finally, it greedily picks the action with the highest estimated Q-value.

Besides taking advantage of the surprising property we found in \datasetname, \algac has other properties which help us bridge the theory-practice gap in RL.
It explores randomly, like common deep RL algorithms, meaning that it can give us insight into why random exploration works much better than the worst-case given in Theorem \ref{thm:deterministicwc}. Also, unlike other RL algorithms, it has cleanly separated \emph{exploration} and \emph{learning} stages, making it much easier to analyze than algorithms in which exploration and learning are entangled.

Furthermore, \algac is extensible beyond environments satisfying the property we found in \datasetname. First, it can solve MDPs where one may have to apply a few steps of value iteration to the random policy's Q-function before acting randomly. Second, it can use an ``exploration policy'' $\expolicy$ different from the random policy $\randpolicy$. These two generalizations are captured in the following definition. In the definition, we use the notation that a step of Q-value iteration transforms a Q-function $Q$ to $Q' = \qvi(Q)$, where
\begin{equation*}
    Q'_\eptime(\state, \ac) = \reward_\eptime(\state, \ac) + \max_{\ac' \in \acspace} Q_{\eptime + 1}\left(\dynamics(\state, \ac), \ac'\right).
\end{equation*}
\begin{definition}[$\numqvi$-QVI-solvable]
    \label{defn:numqvi}
    Given an exploration policy $\expolicy$ ($\expolicy = \randpolicy$ unless otherwise noted), let $Q^1 = Q^{\expolicy}$ and $Q^{i+1} = \qvi(Q^i)$ for $i = 1, \hdots, \horizon - 1$. We say an MDP is \emph{$\numqvi$-QVI-solvable} for some $\numqvi \in [\horizon]$ if every policy in $\policies(Q^\numqvi)$ is optimal.
\end{definition}

We will see that running \algac with $\numqvi > 1$ will allow it to find an optimal policy in MDPs that are $\numqvi$-QVI-solvable. Although the sample complexity of \algac scales with $\acsize^\numqvi$, we find that nearly all of the environments in \datasetname are $\numqvi$-QVI-solvable for very small values of $\numqvi$ (Figure \ref{fig:numqvi_dist}).

We now use \algac to define the effective horizon of an MDP. Note that the total number of timesteps sampled by \algac with parameters $\numqvi$ and $\numep{}$ is $\horizon^2 \acsize^\numqvi \numep{} = \horizon^2 \acsize^{\numqvi + \log_\acsize \numep{}}$. Thus, analogously to how the horizon $\horizon$ appears in the exponent of the worst-case sample complexity bound $O(\horizon \acsize^\horizon)$, we define the \emph{effective} horizon as the exponent of $\acsize$ in the sample complexity of \algac:

\begin{restatable}[Effective horizon]{definition}{definitionefhorizon}
    \label{defn:efhorizon}
    Given $\numqvi \in [\horizon]$, let $\efhorizon{\numqvi} = \numqvi + \log_\acsize \numep{\numqvi}$,
    where $\numep{\numqvi}$ is the minimum value of $\numep{}$ needed for Algorithm \ref{alg:qirl} with parameter $\numqvi$ to return the optimal policy with probability at least $1/2$, or $\infty$ if no value of $\numep{}$ suffices. The \emph{effective horizon} is $\efhorizon{} = \min_\numqvi \efhorizon{\numqvi}$.
\end{restatable}

By definition, the sample complexity of \algac can be given using the effective horizon:

\begin{restatable}{lemma}{lemmaefhorizonsampcomplexity}
    \label{lemma:efhorizon_sampcomplexity}
    The sample complexity of \algac with optimal choices of $\numqvi$ and $\numep{}$ is $\horizon^2 \acsize^{\efhorizon{}}$.
\end{restatable}

As we noted in the introduction, when $\efhorizon{} \ll \horizon$, as we find is often true in practice, this is far better than the worst-case bound given in Theorem \ref{thm:deterministicwc} which scales with $\acsize^\horizon$. 

Definition \ref{defn:efhorizon} does not give a method for actually calculating the effective horizon. It turns out we can bound the effective horizon using a generalized gap notion like those found throughout the RL theory literature. We denote by $\gap^\numqvi_\eptime$ the gap of the Q-function $Q^\numqvi$ from Definition \ref{defn:numqvi}, where
\begin{align*}
    \gap^\numqvi_\eptime(\state) = \max_{\ac \in \acspace} Q^\numqvi_\eptime(\state, \ac) - \max_{\ac' \not\in \arg \max_{\ac} Q^\numqvi_\eptime(\state, \ac)} Q^\numqvi_\eptime(\state, \ac').
\end{align*}
The following theorem gives bounds on the effective horizon in terms of this gap.
\begin{restatable}{theorem}{thmdeterministicefhorizon}
    \label{thm:deterministicefhorizon}
    Suppose that an MDP is $\numqvi$-QVI-solvable and that all rewards are nonnegative, i.e. $\reward(\state, \ac) \geq 0$ for all $\state, \ac$. Let $\gap^\numqvi$ denote the gap of the Q-function $Q^\numqvi$ as defined in Definition \ref{defn:numqvi}. Then
    \begin{align}
        \efhorizon{\numqvi} \leq \numqvi & + \max_{\eptime \in [\horizon], \state \in \statespace^\text{opt}_i, \ac \in \acspace}
        \log_\acsize \left( \frac{Q^\numqvi_\eptime(\state, \ac) V^*_\eptime(\state)}{\gap_\eptime^\numqvi(\state)^2} \right) \nonumber
        + \log_\acsize 6 \log \left(2 \horizon \acsize^\numqvi\right), \label{eq:deterministicefhorizon}
    \end{align}
    where $\statespace^\text{opt}_i$ is the set of states visited by some optimal policy at timestep $i$ and $V^*_\eptime(\state) = \max_\policy V^\policy_\eptime(\state)$ is the optimal value function.
\end{restatable}

A full proof of Theorem \ref{thm:deterministicefhorizon} is given in Appendix \ref{sec:proofs}. Intuitively, the smaller the gap $\gap^\numqvi_\eptime(\state)$, the more precisely we must estimate the Q-values in \algac in order to pick an optimal action.

The \algac algorithm is very amenable to theoretical analysis because it reduces the problem of finding an optimal policy to the problem of estimating several $\numqvi$-step Q-values, each of which is a simple mean of i.i.d. random variables.
There are endless tail bounds that can be applied to analysis of \algac; we use some of these to obtain even tighter bounds on the effective horizon in Appendix \ref{sec:tighter_bounds}. 

Why should the effective horizon, which is defined in terms of our \algac algorithm, also explain the performance of deep RL algorithms like PPO and DQN which are very different from \algac? In Appendix \ref{sec:additional_efhorizon_theory}, we present two algorithms, PG-\algac and FQI-\algac, which are more similar to PPO and DQN but whose sample complexities can still be bounded with the effective horizon.
We also give additional bounds on the effective horizon and lower bounds on sample complexity.

\begin{figure*}
    \centering
    \begin{subfigure}[b]{0.2\textwidth}
        \captionsetup{font=scriptsize}
        \centering
        \resizebox{\textwidth}{!}{
        \begin{tikzpicture}[auto,node distance=8mm,scale=0.8,>=latex,font=\small]
            \tikzstyle{state}=[thick,draw=black,circle]
        
            \node[state] at (0, 0) (s1) {$\state_1$};
            \node[state] at (-1.1, -1) (s2_1) {$\state_2$};
            \draw[->] (s1) -- node[above left] {$\reward = 0$} (s2_1);
            \node at (0, -1) (s2_ellipsis) {$\dots$};
            \draw[->] (s1) -- node[right] {$0$} (s2_ellipsis);
            \node[state] at (1.1, -1) (s2_2) {$\state_2$};
            \draw[->] (s1) -- node[above right] {$0$} (s2_2);

            \node at (-1.6, -2) (s3_1) {$\dots$};
            \draw[->] (s2_1) -- node[above left] {$\reward = 0$} (s3_1);
            \node at (-0.6, -2) (s3_2) {$\dots$};
            \draw[->] (s2_1) -- node[above right] {$0$} (s3_2);
            \node at (0.6, -2) (s3_3) {$\dots$};
            \draw[->] (s2_2) -- node[above left] {$0$} (s3_3);
            \node at (1.6, -2) (s3_4) {$\dots$};
            \draw[->] (s2_2) -- node[above right] {$0$} (s3_4);

            \node[state] at (-1.6, -3) (sT_1) {$\state_\horizon$};
            \draw[->] (s3_1) -- node[left] {$\reward = 1$} (sT_1);
            \node[state] at (-0.6, -3) (sT_2) {$\state_\horizon$};
            \draw[->] (s3_2) -- node[right] {$0$} (sT_2);
            \node[state] at (0.6, -3) (sT_3) {$\state_\horizon$};
            \draw[->] (s3_3) -- node[left] {$0$} (sT_3);
            \node[state] at (1.6, -3) (sT_4) {$\state_\horizon$};
            \draw[->] (s3_4) -- node[right] {$0$} (sT_4);
        \end{tikzpicture}
        }
        \caption{Sparse rewards: when only one sequence of actions gives a reward of 1 and all others give 0, the effective horizon $\efhorizon{} = \tilde{O}(\horizon)$.}
        \label{fig:sparse_reward_example}
    \end{subfigure}
    \hfill
    \begin{subfigure}[b]{0.2\textwidth}
        \captionsetup{font=scriptsize}
        \centering
        \resizebox{\textwidth}{!}{
        \begin{tikzpicture}[auto,node distance=6mm,scale=0.8,>=latex,font=\small]
            \tikzstyle{state}=[thick,draw=black,circle]
        
            \node[state] at (0, 0) (s1) {$\state_1$};
            \node[state] at (-1.1, -1) (s2_1) {$\state_2$};
            \draw[->] (s1) -- node[above left] {$\reward = 1$} (s2_1);
            \node at (0, -1) (s2_ellipsis) {$\dots$};
            \draw[->] (s1) -- node[right] {$0$} (s2_ellipsis);
            \node[state] at (1.1, -1) (s2_2) {$\state_2$};
            \draw[->] (s1) -- node[above right] {$0$} (s2_2);

            \node at (-1.6, -2) (s3_1) {$\dots$};
            \draw[->] (s2_1) -- node[above left] {$\reward = 1$} (s3_1);
            \node at (-0.6, -2) (s3_2) {$\dots$};
            \draw[->] (s2_1) -- node[above right] {$0$} (s3_2);
            \node at (0.6, -2) (s3_3) {$\dots$};
            \draw[->] (s2_2) -- node[above left] {$0$} (s3_3);
            \node at (1.6, -2) (s3_4) {$\dots$};
            \draw[->] (s2_2) -- node[above right] {$0$} (s3_4);

            \node[state] at (-1.6, -3) (sT_1) {$\state_\horizon$};
            \draw[->] (s3_1) -- node[left] {$\reward = 1$} (sT_1);
            \node[state] at (-0.6, -3) (sT_2) {$\state_\horizon$};
            \draw[->] (s3_2) -- node[right] {$0$} (sT_2);
            \node[state] at (0.6, -3) (sT_3) {$\state_\horizon$};
            \draw[->] (s3_3) -- node[left] {$0$} (sT_3);
            \node[state] at (1.6, -3) (sT_4) {$\state_\horizon$};
            \draw[->] (s3_4) -- node[right] {$0$} (sT_4);
        \end{tikzpicture}
        }
        \caption{Dense rewards: when every optimal action gives a reward of 1 and suboptimal actions give no reward, the effective horizon $\efhorizon{} = \tilde{O}(1)$.}
        \label{fig:dense_reward_example}
    \end{subfigure}
    \hfill
    \begin{subfigure}[b]{0.2\textwidth}
        \captionsetup{font=scriptsize}
        \centering
        \resizebox{\textwidth}{!}{
        \begin{tikzpicture}[auto,node distance=6mm,scale=0.8,>=latex,font=\small]
            \tikzstyle{state}=[thick,draw=black,circle]
        
            \node[state] at (0.5, 0) (s1) {$\state_1$};
            \node[state] at (-0.3, -1) (s2_1) {$\state_2$};
            \draw[->] (s1) -- node[above left] {$\reward = 0$} (s2_1);
            \node[state] at (1.8, -1) (s2_2) {$\state_2$};
            \draw[->] (s1) -- node[above right] {$0$} (s2_2);
            
            \node at (-1.1, -2) (s3_1) {$\dots$};
            \draw[->] (s2_1) -- node[above left] {$\reward = 0$} (s3_1);
            \node at (0.6, -2) (s3_3) {$\dots$};
            \draw[->] (s2_1) -- node[above right] {$0$} (s3_3);
            \node at (1.8, -2) (s3_4) {$\dots$};
            \draw[->] (s2_2) -- node[right] {$0$} (s3_4);

            \node[state] at (-1.6, -3) (sT_1) {$\state_\horizon$};
            \draw[->] (s3_1) -- node[left] {$\reward = \horizon$} (sT_1);
            \node[state] at (-0.6, -3) (sT_2) {$\state_\horizon$};
            \draw[->] (s3_1) -- node[right] {$\horizon - 1$} (sT_2);
            \node[state] at (0.6, -3) (sT_3) {$\state_\horizon$};
            \draw[->] (s3_3) -- node[right] {$1$} (sT_3);
            \node[state] at (1.8, -3) (sT_4) {$\state_\horizon$};
            \draw[->] (s3_4) -- node[right] {$0$} (sT_4);
        \end{tikzpicture}
        }
        \caption{Delayed rewards: when the rewards in (b) are all delayed to the end of the episode, the effective horizon remains $\tilde{O}(1)$.}
        \label{fig:delayed_reward_example}
    \end{subfigure}
    \hfill
    \begin{subfigure}[b]{0.3\textwidth}
        \captionsetup{font=scriptsize}
        \centering
        \resizebox{\textwidth}{!}{
        \begin{minipage}[b]{0.45\textwidth}
            \includegraphics[width=\textwidth]{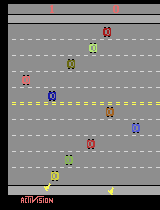}
        \end{minipage}
        \hfill
        \begin{minipage}[b]{0.45\textwidth}
            \scriptsize
            \vspace{-12pt}
            \begin{align*}
                Q^1_{10}(\state, \textsc{Noop}) & = .118 \\
                Q^1_{10}(\state, \textsc{Up}) & = .136 \\
                Q^1_{10}(\state, \textsc{Down}) & = .116 \\
            \end{align*}
            \vspace{-24pt}
            \begin{align*}
                Q^*_{10}(\state, \textsc{Noop}) & = 4 \\
                Q^*_{10}(\state, \textsc{Up}) & = 5 \\
                Q^*_{10}(\state, \textsc{Down}) & = 4
            \end{align*}
        \end{minipage}
        }
        \caption{For the first 50 timesteps of the Atari game Freeway, we can bound $\efhorizon{} \leq 10.2$, which is much lower than the horizon $\horizon = 50$.}
        \label{fig:atari_example}
    \end{subfigure}
    \caption{Examples of calculating the effective horizon $\efhorizon{}$ using Theorem \ref{thm:deterministicefhorizon}; see Section \ref{sec:efhorizon_examples} for the details.}
    \label{fig:efhorizon_examples}
\end{figure*}

\subsection{Examples of the effective horizon}
\label{sec:efhorizon_examples}

To gain some intuition for the bound in Theorem \ref{thm:deterministicefhorizon}, consider the examples in Figure \ref{fig:efhorizon_examples}. MDP (a) has extremely sparse rewards, with a reward of 1 only given for a single optimal action sequence. However, note that this MDP is still 1-QVI-solvable by Definition \ref{defn:numqvi}. The maximum of the bound in Theorem \ref{thm:deterministicefhorizon} is at $\eptime = 1$ with the optimal action, where $Q^1_1(\state, \ac) = 1 / A^{\horizon - 1}$, $V^*_1(\state) = 1$, and $\gap^1_1(\state) = 1 / A^{\horizon - 1}$. Ignoring logarithmic factors and constants gives $\efhorizon{} \lesssim 1 + \log_A A^{\horizon - 1} = \horizon$. That is, in the case of MDP (a), the effective horizon is no better than the horizon.

Next, consider MDP (b), which has dense rewards of 1 for every optimal action. Again, this MDP is 1-QVI-solvable. The maximum in the bound is obtained at $\eptime = 1$ and the optimal action with $Q^1_1(\state, \ac) \leq 2$, $V^*(\state) = \horizon$, and the gap $\gap^1_1(\state, \ac) \geq 1$. Again ignoring logarithmic factors gives in this case $\efhorizon{} \lesssim 1 + \log_\acsize \horizon = \tilde{O}(1)$. In this case, the effective horizon is much shorter than the horizon, and barely depends on it! This again reflects our intuition that in this case, finding the optimal policy via RL should be much easier.

MDP (c) is similar to MDP (b) except that all rewards are delayed to the end of the episode.
In this case, the $Q$ function is the same as in MDP (b) so the effective horizon remains $\tilde{O}(1)$. This may seem counterintuitive since one needs to consider rewards $\horizon$ timesteps ahead to act optimally. However, the way \algac uses random rollouts to evaluate leaf nodes means that it can implicitly consider rewards quite far in the future even without needing to exhaustively plan that many timesteps ahead.

Finally, consider MDP (d), the first 50 timesteps of the Atari game Freeway, which is included in the \datasetname dataset. This MDP is also 1-QVI-solvable and the maximum in the bound is obtained in the state shown in Figure \ref{fig:atari_example} at timestep $\eptime = 10$. 
Plugging in the Q values shown in the figure gives $\efhorizon{} \leq 10.2$, which is far lower than the horizon $\horizon = 50$.
The low effective horizon reflects how this MDP is much easier than the worst case in practice. Both PPO and DQN are able to solve it with a sample complexity of less than 1.5 million timesteps, while the worst case bound would suggest a sample complexity greater than $50 \times 3^{50} / 2 \approx 10^{25}$ timesteps!

\smallparagraph{Comparison to other bounds}
\label{sec:bound_comparisons}
Intuitively, why might the effective horizon give better sample complexity bounds than previous works presented in Section \ref{sec:related_work}? The MDP in Figure \ref{fig:dense_reward_example} presents a problem for the covering length and UCB-based bounds, both of which are $\Omega(\acsize^\horizon)$. The exponential sample complexity arises because these bounds depend on visiting every state in the MDP during training. In contrast, \algac doesn't need to visit every state to find an optimal policy. The effective horizon of $\tilde{O}(1)$ for MDP (b) reflects this, showing that our effective horizon-based bounds can actually be much smaller than the state space size, which is on the order of $\acsize^\horizon$ for MDP (b).

The effective planning window (EPW) does manage to capture the same intuition as the effective horizon in the MDP in Figure \ref{fig:dense_reward_example}: in this case, $\epw = 1$. However, the analysis based on the EPW is unsatisfactory because it entirely ignores rewards beyond the planning window. Thus, in MDP (c) the EPW $\epw = \horizon$, making EPW-based bounds no better than the worst case.
In contrast, the effective horizon-based bound remains the same between MDPs (b) and (c), showing that it can account for the ability of RL algorithms to use rewards beyond the window where exhaustive planning is possible.

\section{Experiments}
\label{sec:experiments}

We now show that sample complexity bounds based on the effective horizon predict the empirical performance of deep RL algorithms far better than other bounds in the literature.
For each MDP in the \datasetname dataset, we run deep RL algorithms to determine their empirical sample complexity. We also use the tabular representations of the MDPs to calculate the effective horizon and other sample complexity bounds for comparison.

\smallparagraph{Deep RL algorithms} We run both PPO and DQN for five million timesteps for each MDP in \datasetname, and record the empirical sample complexity (see Appendix \ref{sec:experiment_details} for hyperparameters and experiment details). PPO converges to the optimal policy in 95 of the 155 MDPs, and DQN does in 117 of 155. At least one of the two finds the optimal policy in 119 MDPs.


\smallparagraph{Sample complexity bounds} We also compute sample complexity bounds for each MDP in \datasetname. These include the worst-case bound of $\horizon \acsize^\horizon$ from Theorem \ref{thm:deterministicwc}, the effective-horizon-based bound of $\horizon^2 \acsize^{\efhorizon{}}$ from Lemma \ref{lemma:efhorizon_sampcomplexity}, as well as three other bounds from the literature, introduced in Section \ref{sec:related_work} and proved in Appendix \ref{sec:more_related_work}: the UCB-based bound $\statesize \acsize \horizon$, the covering-length-based bound $\horizon \covlen$, and the effective planning window (EPW)-based bound of $\horizon^2 \acsize^\epw$.

\smallparagraph{Evaluation metrics} To determine which sample complexity bounds best reflect the empirical performance of PPO and DQN, we compute a few summary metrics for each bound. First, we measure the \emph{Spearman (rank) correlation} between the sample complexity bounds and the empirical sample complexity over environments where the algorithm converged to the optimal policy.
The correlation (higher is better) is a useful for measuring how well the bounds can rank the relative difficulty of RL in different MDPs.

Second, we compute the \emph{median ratio} between the sample complexity bound and the empirical sample complexity for environments where the algorithm converged.
The ratio between the bound $\sampcomplexity_\text{bound}$ and empirical value $\sampcomplexity_\text{emp}$ is calculated as $\max \{ \sampcomplexity_\text{bound} / \sampcomplexity_\text{emp}, \sampcomplexity_\text{emp} / \sampcomplexity_\text{bound} \}$.
For instance, a median ratio of 10 indicates that half the sample complexity bounds were within a factor of 10 of the empirical sample complexity. 
Lower values indicate a better bound; this metric is useful for determining whether the sample complexity bounds are vacuous or tight.

Finally, we consider the binary classification task of predicting whether the algorithm will converge at all within five million steps using the sample complexity bounds. That is, we consider simply thresholding each sample complexity bound and predicting that only environments with bounds below the threshold will converge. We compute the \emph{area under the ROC curve (AUROC)} for this prediction task as well as the \emph{accuracy} with the optimal threshold. Higher AUROC and accuracy both indicate a better bound.

\begin{table*}[t]
    \centering
    \resizebox{\textwidth}{!}{
    \begin{tabular}{l|rrrr|rrrr}
        \toprule
        & \multicolumn{4}{|c|}{\bf PPO} & \multicolumn{4}{|c}{\bf DQN} \\
        \bf Bound & Correl. & Median ratio & AUROC & Acc. & Correl. & Median ratio & AUROC & Acc. \\
        \midrule
        Worst-case ($\horizon \lceil \acsize^\horizon / 2 \rceil$) & 0.24 & $7.2 \times 10^{10}$ & 0.57 & 0.63 & 0.15 & $5.5 \times 10^{10}$ & 0.67 & 0.76 \\
        Covering length ($\horizon \covlen$) & 0.35 & $6.3 \times 10^{6}$ & 0.78 & 0.72 & 0.27 & $3.9 \times 10^{6}$ & 0.86 & 0.85 \\
        EPW ($\horizon^2 \acsize^\epw$) & 0.69 & $1.1 \times 10^{5}$ & 0.78 & 0.75 & 0.58 & $8.0 \times 10^{4}$ & 0.88 & 0.85 \\
        UCB ($\statesize \acsize \horizon$) & 0.26 & \bf 20 & 0.68 & 0.67 & 0.31 & \bf 31 & 0.67 & 0.77 \\
        Effective horizon ($\horizon^2 \acsize^{\efhorizon{}{}}$) & \bf 0.81 & 31 & \bf 0.92 & \bf 0.86 & \bf 0.74 & 67 & \bf 0.92 & \bf 0.86 \\
        \midrule
        \emph{Other deep RL algorithm} & 0.77 & 2.3 & 0.84 & 0.85 & 0.77 & 2.3 & 0.86 & 0.99 \\
        \emph{\algac empirical} & 0.79 & 7.3 & 0.77 & 0.82 & 0.65 & 11 & 0.80 & 0.94 \\
        \bottomrule
    \end{tabular}
    }
    \vspace{-6pt}
    \caption{Effective horizon-based sample complexity bounds are the most predictive of the real performance of PPO and DQN according to the four metrics we describe in Section \ref{sec:experiments}. The effective horizon bounds are about as good at predicting the sample complexity of PPO and DQN as one algorithm's sample complexity is at predicting the other's.}
    \label{tab:bound_evaluations}
    \vspace{-6pt}
\end{table*}

\smallparagraph{Results}
\label{sec:results}
The results of our experiments are shown in Table \ref{tab:bound_evaluations}. The effective horizon-based bounds have higher correlation with the empirical sample complexity than the other bounds for both PPO and DQN.
While the EPW-based bounds are also reasonably correlated, they are significantly off in absolute terms: the typical bound based on the EPW is 3-4 orders of magnitude off, while the effective horizon yields bounds that are typically within 1-2 orders of magnitude.
The UCB-based bounds are somewhat closer to the empirical sample complexity, but are not well correlated; this makes sense since the UCB bounds depend on strategic exploration, while PPO and DQN use random exploration.
Finally, the effective horizon bounds are able to more accurately predict whether PPO or DQN will find an optimal policy, as evidenced by the AUROC and accuracy metrics.

As an additional baseline, we also calculate the four evaluation metrics when using the empirical sample complexity of PPO to predict the empirical sample complexity of DQN, or vice-versa, and when using the empirical sample complexity of \algac to predict PPO or DQN's performance (bottom two rows of Table \ref{tab:bound_evaluations}). While these are not provable bounds, they provide another point of comparison for each metric.
The effective horizon-based bounds correlate about as well with PPO and DQN's sample complexities as they do with each other's. The empirical performance of \algac is typically even closer to that of PPO and DQN than the effective horizon-based bounds.

\begin{table}[t]
    \begin{subtable}{0.55\textwidth}
        \centering
        \small
        \begin{tabular}{l|rr|rr}
            \toprule
            & \multicolumn{2}{|c|}{\bf PPO} & \multicolumn{2}{|c}{\bf DQN} \\
            \bf Bound & Correl. & Ratio & Correl. & Ratio \\
            \midrule
            EPW & 0.20 & 2.1 & \bf 0.70 & 12 \\
            Effective horizon & \bf 0.48 & 2.4 & 0.35 & 12 \\
            Other bounds & 0.00 & \bf 1.3 & 0.00 & \bf 1.9 \\
            \bottomrule
        \end{tabular}
        \caption{Reward shaping. \\}
        \vspace{12pt}
        \label{tab:reward_shaping}
    \end{subtable}
    \hfill
    \begin{subtable}{0.41\textwidth}
        \centering
        \small
        \begin{tabular}{l|rr}
            \toprule
            & \multicolumn{2}{|c}{\bf PPO} \\
            \bf Bound & Correl. & Ratio \\
            \midrule
            Covering length & -0.36 & $2.5 \times 10^{4}$ \\
            Effective horizon & \bf 0.57 & 2.7 \\
            Other bounds & 0.00 & \bf 2.2 \\
            \bottomrule
        \end{tabular}
        \caption{Initializing with a policy trained on human data or transferred from similar environments.}
        \label{tab:exploration_policy}
    \end{subtable}
    \caption{The effective horizon explains the effects of reward shaping and initializing with a pretrained policy by accurately predicting their effects on the empirical sample complexity of PPO and DQN. Correlation and median ratio are measured between the predicted change in sample complexity and the empirical change. See Section \ref{sec:intervention_experiments} for further discussion.}
    \vspace{-12pt}
\end{table}

\smallparagraph{Reward shaping, human data, and transfer learning}
\label{sec:intervention_experiments}
In order for RL theory to be useful practically, it should help practitioners make decisions about which techniques to apply in order to improve their algorithms' performance. We show how the effective horizon can be used to explain the effect of three such techniques: reward shaping, using human data, and transfer learning.

Potential-based reward shaping \cite{ng_policy_1999} is a classic technique which can speed up the convergence of RL algorithms.
It is generally used to transform a sparse-reward MDP like the one in Figure \ref{fig:sparse_reward_example} to a dense-reward MDP like in Figure \ref{fig:dense_reward_example} without changing the optimal policy.
If the effective horizon is a good measure of the difficulty of RL in an environment, then it should be able to predict whether (and by how much) the sample complexity of practical algorithms changes when reward shaping is applied. We develop 77 reward-shaped versions of the original 33 Minigrid environments and run PPO and DQN. Results in Table \ref{tab:reward_shaping} show the effective horizon accurately captures the change in sample complexity from using reward shaping. We use similar metrics to those in Table \ref{tab:bound_evaluations}: the correlation between the predicted and empirical ratio of the shaped sample complexity to the unshaped sample complexity, and the median ratio between the predicted change and the actual change.

Out of the five bounds we consider, three---worst case, covering length, and UCB---don't even depend on the reward function. The EPW does depend on the reward function and captures some of the effect of reward shaping for DQN, but does worse at predicting the effect on PPO. In comparison, the effective horizon does well for both algorithms, showing that it can accurately capture how reward shaping affects RL performance.

Another tool used to speed up RL is initializing with a pre-trained policy, which is used practically to make RL work on otherwise intractable tasks. Can the effective horizon also predict whether initializing RL with a pre-trained policy will help?
We initialize PPO with pre-trained policies for 82 of the MDPs in \datasetname, then calculate new sample complexity bounds based on using the pre-trained policies as an exploration policy $\expolicy$. Table \ref{tab:exploration_policy} shows that the effective horizon accurately predicts the change in sample complexity due to using a pre-trained policy. Again, three bounds---worst case, EPW, and UCB---do not depend on the exploration policy at all, while the covering length gives wildly inaccurate predictions for its effect. In contrast, the effective horizon is accurate at predicting the changes in sample complexities when using pre-trained policies.

\begin{figure}[t]
    \input{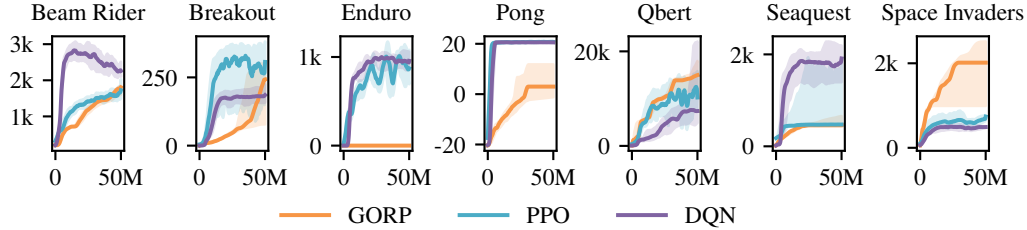}
    \caption{Learning curves for PPO, DQN, and GORP on full-horizon Atari games. We use 5 random seeds for all algorithms. The solid line shows the median return throughout training while the shaded region shows the range of returns over random seeds.}
    \label{fig:learning_curves_full_atari}
\end{figure}

\smallparagraph{Long-horizon environments}
\label{sec:long_horizon}
We also perform experiments on full-length Atari games to evaluate the predictive power of the effective horizon in longer-horizon environments. It is intractable to construct tabular representations of these environments and thus we cannot compute instance-dependent sample complexity bounds. However, it is still possible to compare the empirical performance of PPO, DQN, and \algac. If the performance of \algac is close to that of PPO and DQN, then this suggests that the effective horizon, which is defined in terms of \algac, can explain RL performance in these environments as well. Figure \ref{fig:learning_curves_full_atari} compares the learning curves of PPO, DQN, and \algac in deterministic versions of the seven Atari games from \citet{mnih_playing_2013} (see Appendix \ref{sec:full_atari} for details). \algac performs better than both PPO and DQN in two games and better than at least one deep RL algorithm in an additional three games. This provides evidence that the effective horizon is also predictive of RL performance in long-horizon environments.

\section{Discussion}
\label{sec:discussion}

Overall, our results suggest the effective horizon is a key measure of the difficulty of solving an MDP via reinforcement learning.
The intuition behind the effective horizon presented in Section \ref{sec:efhorizon} and the empirical evidence in Section \ref{sec:experiments} both support its importance for better understanding RL.

\smallparagraph{Limitations}
\label{sec:limitations}
While we have presented a thorough theoretical and empirical justification of the effective horizon, there are still some limitations to our analysis. First, we focus on deterministic MDPs with discrete action spaces, leaving the extension to stochastic environments and those with continuous action spaces an open question. Furthermore, the effective horizon is not easily calculable without full access to the MDP's tabular representation. Despite this, it serves as a useful perspective for understanding RL's effectiveness and potential improvement areas. An additional limitation is that the effective horizon cannot capture the performance effects of generalization---the ability to use actions that work well at some states for other similar states. For an example where the effective horizon fails to predict generalization, see Appendix \ref{sec:generalization_example}. However, the effective horizon is still quite predictive of deep RL performance even without modeling generalization.

\smallparagraph{Implications and future work}
\label{sec:implications}
We hope that this paper helps to bring the theoretical and empirical RL communities closer together in pursuit of shared understanding. Theorists can extend our analysis of the effective horizon to new algorithms or explore related properties, using our \datasetname dataset to ground their investigations by testing assumptions in real environments. Empirical RL researchers can use the effective horizon as a foundation for new algorithms. For instance, \citet{brandfonbrener_offline_2021} present an offline RL algorithm similar to \algac with $\numqvi = 1$; our theoretical understanding of \algac might provide insights for improving it or developing related algorithms.

\begin{ack}
We would like to thank Jacob Steinhardt, Kush Bhatia, Lawrence Chan, Paria Rashidinejad, Ruiqi Zhong, and Yaodong Yu for feedback on drafts as well as Jiantao Jiao and Banghua Zhu for helpful discussions.

This work was partially supported by the Open Philanthropy Project, the Office of Naval Researcher's Young Investigator Program (YIP), and Weill Neurohub. Cassidy Laidlaw is supported by an Open Philanthropy AI Fellowship and a National Defense Science and Engineering Graduate (NDSEG) Fellowship. Stuart Russell is supported by an AI 2050 Senior Fellowship from Schmidt Futures.
\end{ack}

\bibliographystyle{unsrtnat}
\bibliography{paper}

\newpage
\appendix

{\Large \bf Appendix}

\section{Proofs of main results}
\label{sec:proofs}

\subsection{Proof of Theorem \ref{thm:deterministicwc}}

\thmdeterministicwc*

\begin{proof}
Consider the following RL algorithm:
\begin{algorithmic}
    \Procedure{ExhaustiveSearch}{}
        \State $\mathcal{T} \gets \text{Shuffle}(\acspace^\horizon)$
        \State $\mathcal{J}$ is an array of size $\lceil \acsize^\horizon / 2 \rceil$
        \For{$i = 1, \hdots, \lceil \acsize^\horizon / 2 \rceil$}
            \State run one episode, taking the actions in $\mathcal{T}[i]$
            \State $\mathcal{J}[i] \gets \sum_{\eptime=1}^\horizon \gamma^\eptime \reward(\state_\eptime, \ac_\eptime)$
        \EndFor
        \State $i^* \gets \arg \max_i \mathcal{J}[i]$
		\State \Return the policy which takes the actions in $\mathcal{T}[i^*]$
    \EndProcedure
\end{algorithmic}
Since the MDP is deterministic, an RL algorithm only needs to find an optimal sequence of actions. Clearly, there is at least a $1/2$ chance that some optimal sequence of actions is in the first $\lceil \acsize^\horizon / 2 \rceil$ elements of $\mathcal{T}$. If this is the case, then $\textsc{ExhaustiveSearch}$ will return an optimal policy corresponding to that optimal sequence. Since the number of environment timesteps taken by $\textsc{ExhaustiveSearch}$ is equal to $\horizon \lceil \acsize^\horizon / 2 \rceil$, we have that $\sampcomplexity \leq \horizon \lceil \acsize^\horizon / 2 \rceil$.

For the converse, fix $\acsize$ and $\horizon$ along with any RL algorithm. Consider a set of states indexed by sequences of actions of length $0$ to $\horizon$:
\begin{equation*}
    \statespace = \{ \state_{\ac_{1:\ell}} \mid \ell \in 0, \hdots, \horizon, \ac_{1:\ell} \in \acspace^\ell \}.
\end{equation*}
Then, define a transition function
\begin{equation*}
    \dynamics(\state_{\ac_{1:\ell}}, \ac) = \state_{\ac_{1:\ell}, \ac}.
\end{equation*}
Now consider $\acsize^\horizon$ different MDPs which share the state space $\statespace$ and transition function $\dynamics$, differing only in their reward functions:
\begin{equation*}
    \mathbb{M} = \{\mdp_{\ac_{1:\horizon}} \mid \ac_{1:\horizon} \in \acspace^\horizon \}
    \quad \text{where the MDP } \mdp_{\ac_{1:\horizon}} \text{ has } \reward(\state_{\ac_{1:\ell}}, \ac) = \begin{cases}
        1 \quad & \ac_{1:\ell}, \ac = \ac_{1:\horizon} \\
        0 \quad & \text{otherwise.}
    \end{cases}
\end{equation*}
That is, each MDP has a single optimal sequence of actions that gives reward 1 on the final timestep; all other rewards are 0.

Let the RL algorithm in question take in some source of randomness $z$ and output a policy $\policy^\samples_\eptime(\ac \mid \state; z, \mdp)$ after $\samples$ timesteps in MDP $\mdp$. Now suppose by way of contradiction the sample complexity of the algorithm is less than $\horizon ( \lceil \acsize^\horizon / 2 \rceil - 1 )$ in all MDPs in $\mathbb{M}$. By our definition of sample complexity, this means that
\begin{equation}
    \label{eq:exhaustive_converse_sampcomplexity}
    \forall \mdp \in \mathbb{M} \qquad \PP_z \left( \pret\left(\policy^{\horizon (\lceil \acsize^\horizon / 2 \rceil - 2)}_\eptime(\cdot ; z, \mdp)\right) = 1 \right) \geq 1/2.
\end{equation}
Clearly a policy can only be optimal in these MDPs if it is deterministic, so let $\tau(z, \mdp)$ be the sequence of actions that $\policy^{\horizon (\lceil \acsize^\horizon / 2 \rceil - 2)}_\eptime(\ac \mid \state ; z, \mdp)$ takes if it is optimal, and let it be any suboptimal sequence of actions if the policy is suboptimal. We can rewrite (\ref{eq:exhaustive_converse_sampcomplexity}) as
\begin{equation*}
    \forall \mdp_{\ac_{1:\horizon}} \in \mathbb{M} \qquad \PP_z \left( \tau(z, \mdp_{\ac_{1:\horizon}}) = \ac_{1:\horizon} \right) \geq 1/2.
\end{equation*}
Letting $\text{Unif}(\acspace^\horizon)$ define a uniform distribution over action sequences of length $\horizon$, this implies
\begin{equation*}
    \PP_{\ac_{1:\horizon} \sim \text{Unif}(\acspace^\horizon), z} \Big( \tau(z, \mdp_{\ac_{1:\horizon}}) = \ac_{1:\horizon} \Big) \geq 1/2
\end{equation*}
which means that there must be some particular $z$ such that
\begin{align}
    \PP_{\ac_{1:\horizon} \sim \text{Unif}(\acspace^\horizon)} \Big( \tau(z, \mdp_{\ac_{1:\horizon}}) = \ac_{1:\horizon} \Big) & \geq 1/2 \nonumber \\
    \left| \left\{ \ac_{1:\horizon} \in \acspace^\horizon \mid \tau(z, \mdp_{\ac_{1:\horizon}}) = \ac_{1:\horizon} \right\} \right|  & \geq \acsize^\horizon / 2 \label{eq:exhaustive_num_optimal_traj}
\end{align}
Given that the RL algorithm is now deterministic due to the fixed $z$, we will prove that the LHS of (\ref{eq:exhaustive_num_optimal_traj}) must be less than or equal to $\lceil \acsize^\horizon / 2 \rceil - 1$. This can be shown via induction. For the first episode, the algorithm must take the same actions in every MDP since all MDPs give zero reward until the final action (so there is no way to distinguish them). After the first episode, only one MDP can be distinguished from the others: the one corresponding to the action sequence taken in the first episode, which has reward 1 instead of 0. Thus in the remaining $\acsize^\horizon - 1$ MDPs the algorithm must take the same actions in the second episode. Continuing this argument shows that after episode $\lceil \acsize^\horizon / 2 \rceil - 2$, the algorithm must still be unable to distinguish between $\acsize^\horizon - \left( \lceil \acsize^\horizon / 2 \rceil - 2 \right) = \lfloor \acsize^\horizon / 2 \rfloor + 2$ of the MDPs, and so $\tau(z, \mdp)$ must be the same for all MDPs in this set. Since all of these MDPs have different optimal action sequences, $\tau(z, \mdp)$ can only be optimal in one of them. Thus $\tau(z, \mdp)$ must be suboptimal in at least $\lfloor \acsize^\horizon / 2 \rfloor + 1$ MDPs, which means the LHS of (\ref{eq:exhaustive_num_optimal_traj}) must be at most $\acsize^\horizon - (\lfloor \acsize^\horizon / 2 \rfloor + 1) = \lceil \acsize^\horizon / 2 \rceil - 1$.

Combining this with (\ref{eq:exhaustive_num_optimal_traj}) gives $\lceil \acsize^\horizon / 2 \rceil - 1 \geq \acsize^\horizon / 2$, which is a contradiction. Thus, the sample complexity of the RL algorithm must be at least $\horizon ( \lceil \acsize^\horizon / 2 \rceil - 1 )$.
\end{proof}

\subsection{Proof that $\numqvi = \horizon$ in the worst case}

\begin{lemma}
    \label{lemma:numqvi_wc}
    Let $Q^1 = Q^{\randpolicy}$, $Q^{i+1}$ = $\qvi(Q^i)$ for $i = 2, \hdots, \horizon$, and $\policies(Q^i)$ be defined as in Section \ref{sec:dataset}. Then for any horizon $\horizon$ and number of actions $\acsize \geq 2$ there is an MDP such that no policy in $\policies(Q^i)$ is optimal for $i < \horizon$.
\end{lemma}
\begin{proof}
As in the proof of Theorem \ref{thm:deterministicwc}, define an MDP where every action sequence leads to a different state. Pick an arbitrary action sequence $\ac_{1:\horizon}$, and let the reward of taking the final action in that sequence be $1$:
\begin{equation*}
    \reward(\state_{\ac_{1:\horizon-1}}, \ac_\horizon) = 1.
\end{equation*}
Now take some action $\ac_1' \neq \ac_1$, and let the reward for taking that action at the beginning of an episode be $3/4$:
\begin{equation*}
    \reward(\state_1, \ac_1') = 3/4.
\end{equation*}
Let the rewards for all other state-action pairs be 0. We will show by induction that
\begin{equation}
    \label{eq:numqvi_wc_inductive_hypothesis}
    Q^i_\eptime(\state_{\ac_{1:\eptime - 1}}, \ac_\eptime) = \frac{1}{\acsize^{\max \{ \horizon - i - \eptime + 1, 0 \}}}
    \qquad \text{and} \qquad
    Q^i_1(\state_1, \ac_1') = 3/4.
\end{equation}
The left half of (\ref{eq:numqvi_wc_inductive_hypothesis}) is clearly true for $i = 1$, since the random policy will take all action sequences following the $\eptime-1$-th optimal action with probability $1 / \acsize^{\horizon - \eptime}$, and exactly one of those gives reward 1. The right half is also clear since following $a_1'$ gives immediate reward of 3/4 and then no reward afterwards.

For the inductive step, we begin with the left half of (\ref{eq:numqvi_wc_inductive_hypothesis}); assume it holds for some $i$. By the definition of Q-value iteration, we have for $\eptime < \horizon$
\begin{align*}
    & Q^{i + 1}_\eptime(\state_{\ac_{1:\eptime - 1}}, \ac_\eptime) \\
    & \quad \overset{\text{(i)}}{=} \reward(\state_{\ac_{1:\eptime - 1}}, \ac_\eptime)
    + \max_\ac Q^i_{\eptime + 1}(\state_{\ac_{1:\eptime}}, \ac) \\
    & \quad = 0 + \frac{1}{\acsize^{\max\{\horizon - i - (\eptime + 1) + 1, 0\}}} \\
    & \quad = \frac{1}{\acsize^{\max\{\horizon - (i + 1) - \eptime + 1, 0\}}}
\end{align*}
and for $\eptime = \horizon$,
\begin{align*}
    Q^{i + 1}_\horizon(\state_{\ac_{1:\horizon - 1}}, \ac_\horizon) = \reward(\state_{\ac_{1:\horizon - 1}}, \ac_\horizon) = 1
    = \frac{1}{\acsize^0} = \frac{1}{\acsize^{\max\{\horizon - (i + 1) - \eptime + 1, 0\}}}.
\end{align*}
(i) is because only one action at timestep $\eptime + 1$ can have a Q-value greater than 0, since only one action sequence leads to reward after taking $\ac_1$. For the right half of (\ref{eq:numqvi_wc_inductive_hypothesis}), we have
\begin{align*}
    Q^{i+1}_1(\state_1, \ac_1') = \reward(\state_1, \ac_1') + \max_\ac Q^i_2(\state_{\ac_1'}, \ac)  = 3/4 + 0 = 3/4.
\end{align*}
which is due to no reward being possible after taking $\ac_1'$ at the first timestep.

Given that (\ref{eq:numqvi_wc_inductive_hypothesis}) holds for $i = 1, \hdots, \horizon$, it is easy to see that for $i < \horizon$, any $\policy \in \policies(Q^i)$ will take $\ac_1'$ on the first timestep, since
\begin{equation*}
    Q^i_1(\state_1, \ac_1) = \frac{1}{\acsize^{\max \{ \horizon - i, 0 \}}} = \frac{1}{\acsize^{\horizon - i}} \leq \frac{1}{\acsize} \leq \frac{1}{2} \leq \frac{3}{4} = Q^i_1(\state_1, \ac_1').
\end{equation*}
This means that $\policy$ will be suboptimal, since $\ac_1$ is the only optimal action at $\eptime = 1$.
\end{proof}

\subsection{Proof of Lemma \ref{lemma:efhorizon_sampcomplexity}}

\lemmaefhorizonsampcomplexity*

\begin{proof}
Recall the definition of effective horizon:
\definitionefhorizon*
Let $\numqvi \in \arg \min_{\numqvi} \efhorizon{\numqvi}$. Then by Definition \ref{defn:efhorizon}, \algac (Algorithm \ref{alg:qirl}) with parameters $\numqvi, \numep{\numqvi}$ will converge to an optimal policy with probability at least $1/2$. Clearly, Algorithm \ref{alg:qirl} interacts with the environment for $\horizon$ iterations, each of which require evaluating $\acsize^\numqvi$ action sequences with $\numep{\numqvi}$ episodes of $\horizon$ timesteps each, for a total of
\begin{equation*}
    \horizon^2 \acsize^\numqvi \numep{\numqvi} = \horizon^2 \acsize^{\efhorizon{\numqvi}} = \horizon^2 \acsize^{\efhorizon{}}
\end{equation*}
timesteps. Thus the sample complexity of \algac satisfies $\sampcomplexity_\text{\algac} \leq \horizon^2 \acsize^{\efhorizon{}}$.

Now, suppose by way of contradiction that $\sampcomplexity_\text{\algac} < \horizon^2 \acsize^{\efhorizon{}}$. Then this must mean that there are parameters $\numqvi', \numep{}'$ such running \algac with these parameters converges to an optimal policy with probability at least $1/2$, and
\begin{equation*}
    \numqvi' + \log_\acsize \numep{}' < \numqvi + \log_\acsize \numep{\numqvi}.
\end{equation*}
By Definition \ref{defn:efhorizon}, this means that $\numep{\numqvi'} \leq \numep{}'$, and thus
\begin{equation*}
    \efhorizon{}
    = \min_\numqvi \efhorizon{}
    \leq \numqvi' + \log_\acsize \numep{\numqvi'}
    \leq \numqvi' + \log_\acsize \numep'
    < \numqvi + \log_\acsize \numep{\numqvi}
    = \efhorizon{}
\end{equation*}
which is clearly a contradiction. Thus, it must be that $\sampcomplexity_\text{\algac} = \horizon^2 \acsize^{\efhorizon{}}$.
\end{proof}

\subsection{Proof of Theorem \ref{thm:deterministicefhorizon}}

\thmdeterministicefhorizon*

\begin{proof}
Let
\begin{equation*}
    \numep{} = \log \left(2 \horizon \acsize^\numqvi\right) \max_{\eptime \in [\horizon], \state \in \statespace^\text{opt}_\eptime, \ac \in \acspace} \frac{6 Q^\numqvi_\eptime(\state, \ac) V^*_\eptime(\state)}{\gap_\eptime^\numqvi(\state)^2}.
\end{equation*}
We will show that \algac (Algorithm \ref{alg:qirl}) converges to the optimal policy with probability at least $1/2$ given parameters $\numqvi$ and $\numep{}$. By Definition \ref{defn:efhorizon}, this means the effective horizon must be at most $\numqvi + \log_\acsize \numep{}$, which gives the bound in the theorem.
More precisely, we will show that \algac converges to a policy in $\policies(Q^\numqvi)$ with probability at least $1/2$, which must be optimal because of the assumption that the MDP is $\numqvi$-QVI-solvable.

First, we will show the following relationship between the $\numqvi$-action $Q^1$ values and $Q^\numqvi$:
\begin{equation}
    \label{eq:k_step_q_to_q_k}
    Q^\numqvi_i(\state_i, \ac_i)
    = \max_{\ac_{i + 1 : i + \numqvi - 1} \in \acspace^{\numqvi - 1}} Q^1_i(\state_i, \ac_i, \ac_{i + 1 : i + \numqvi - 1}).
\end{equation}
We prove that (\ref{eq:k_step_q_to_q_k}) holds inductively. For $\numqvi = 1$, (\ref{eq:k_step_q_to_q_k}) is obviously true. Supposing it holds for $\numqvi$, then
\begin{align*}
    Q^{\numqvi + 1}_i(\state_i, \ac_i)
    & = \qvi(Q^\numqvi_i)(\state_i, \ac_i) \\
    & = \reward(\state_i, \ac_i) + \max_{\ac_{i + 1} \in \acspace} Q^\numqvi_{i + 1}(\dynamics(\state_i, \ac_i), \ac_{i + 1}) \\
    & \overset{\text{(i)}}{=} \reward(\state_i, \ac_i) + \max_{\ac_{i + 1 : i + \numqvi} \in \acspace^\numqvi} Q^1_{i + 1}(\dynamics(\state_i, \ac_i), \ac_{i + 1 : i + \numqvi}) \\
    & = \max_{\ac_{i + 1 : i + \numqvi} \in \acspace^\numqvi} Q^1_i(\state_i, \ac_i, \ac_{i + 1 : i + \numqvi}),
\end{align*}
which shows (\ref{eq:k_step_q_to_q_k}) holds for $\numqvi + 1$. (i) holds due to the inductive hypothesis.

Recall that in Algorithm \ref{alg:qirl}, we use $\hat{Q}_i(\state_i, \ac_{i : i + \numqvi - 1})$ to denote the estimated Q-value of the $\numqvi$-action sequence $\ac_{i : i + \numqvi - 1}$. Analogously to (\ref{eq:k_step_q_to_q_k}), define
\begin{equation*}
    \hat{Q}^k_i(\state_i, \ac_i)
    = \max_{\ac_{i + 1 : i + \numqvi - 1} \in \acspace^{\numqvi - 1}} \hat{Q}_i(\state_i, \ac_i, \ac_{i + 1 : i + \numqvi - 1})
\end{equation*}
to be the maximum estimated Q-value of any action sequence that starts with $\ac_i$. We can rewrite line \ref{line:max_action_seq} of Algorithm \ref{alg:qirl} as
\begin{equation*}
    \policy_i(\state_i) \gets \arg \max_{\ac_i \in \acspace}
    \hat{Q}^\numqvi_i(\state_i, \ac_i).
\end{equation*}
That is, the action that \algac selects for timestep $i$ is chosen from those with the highest values of $\hat{Q}^k_i(\state_i, \ac_i)$. Suppose we can show that
\begin{equation}
    \label{eq:qirl_finds_optimal_action}
    \PP \left( \arg \max_{\ac_i \in \acspace}
    \hat{Q}^\numqvi_i(\state_i, \ac_i) \subseteq \arg \max_{\ac_i \in \acspace}
    Q^\numqvi_i(\state_i, \ac_i) \right) \geq 1 - \frac{1}{2 \horizon}
\end{equation}
holds for each $i \in [\horizon]$. Then by a union bound,
\begin{equation}
    \label{eq:qirl_success_prob_union_bound}
    \PP \left( \forall i \in [\horizon] \quad \policy_i(\state_i) \in \arg \max_{\ac_i \in \acspace}
    Q^\numqvi_i(\state_i, \ac_i) \right) \geq \frac{1}{2}.
\end{equation}
This implies that $\policy \in \policies(Q^k)$, which is the desired result.

It remains to show that (\ref{eq:qirl_finds_optimal_action}) holds. We will actually prove the bound assuming that 
$\state_i \in \statespace^\text{opt}_i$.
This is still sufficient to imply (\ref{eq:qirl_success_prob_union_bound}) since one can inductively assume that previous actions are optimal. We can write (\ref{eq:qirl_finds_optimal_action}) equivalently as
\begin{equation}
    \label{eq:qirl_finds_suboptimal_action}
    \PP \left( \exists \ac_i \in  \arg \max_{\ac_i \in \acspace}
    \hat{Q}^\numqvi_i(\state_i, \ac_i) \; : \; \ac_i \notin \arg \max_{\ac_i \in \acspace}
    Q^\numqvi_i(\state_i, \ac_i) \right) \leq \frac{1}{2 \horizon}.
\end{equation}
Let $\ac^*_{i : i + \numqvi - 1} \in \arg \max_{\ac_{i : i + \numqvi - 1} \in \acspace^\numqvi} Q^1_i(\state_i, \ac_{i : i + \numqvi - 1})$ be chosen arbitrarily. By (\ref{eq:k_step_q_to_q_k}), this implies that $\ac^*_i \in \arg \max_{\ac_i \in \acspace} Q^\numqvi_i(\state_i, \ac_i)$. Then
\begin{align}
    & \PP \left( \exists \ac_i \in  \arg \max_{\ac_i \in \acspace}
    \hat{Q}^\numqvi_i(\state_i, \ac_i) \; : \; \ac_i \notin \arg \max_{\ac_i \in \acspace}
    Q^\numqvi_i(\state_i, \ac_i) \right) \nonumber \\
    & \quad \leq \PP \left( \exists \ac_i \notin \arg \max_{\ac_i \in \acspace}
    Q^\numqvi_i(\state_i, \ac_i) \; : \; \hat{Q}^\numqvi_i(\state_i, \ac_i) \geq \hat{Q}^\numqvi_i(\state_i, \ac_i^*) \right) \nonumber \\
    & \quad \leq \PP \left( \exists \ac_i \notin \arg \max_{\ac_i \in \acspace} Q^\numqvi_i(\state_i, \ac_i), \ac_{i + 1 : i + \numqvi - 1} \in \acspace^{\numqvi - 1}
     \; : \; \hat{Q}_i(\state_i, \ac_i, \ac_{i + 1 : \numqvi - 1}) \geq \hat{Q}_i(\state_i, \ac_{i : i + \numqvi - 1}^*) \right) \nonumber \\
    & \quad \leq \sum_{\ac_i \notin \arg \max_{\ac_i \in \acspace} Q^\numqvi_i(\state_i, \ac_i), \ac_{i + 1 : i + \numqvi - 1} \in \acspace^{\numqvi - 1}}
    \PP \left( \hat{Q}_i(\state_i, \ac_i, \ac_{i + 1 : \numqvi - 1}) \geq \hat{Q}_i(\state_i, \ac_{i : i + \numqvi - 1}^*) \right). \label{eq:qirl_finds_suboptimal_action_seq}
\end{align}
Consider a single term of the sum in (\ref{eq:qirl_finds_suboptimal_action_seq}). By the definition of the gap and the fact that $\ac_i \notin \arg \max_{\ac_i \in \acspace} Q^\numqvi_i(\state_i, \ac_i)$, we know that
\begin{equation*}
    Q^\numqvi_i(\state_i, \ac_i) \leq Q^\numqvi_i(\state_i, \ac^*_i) - \gap^\numqvi_i(\state_i).
\end{equation*}
Combining this with (\ref{eq:k_step_q_to_q_k}) implies that
\begin{align}
    Q^1_i(\state_i, \ac_{i : i + \numqvi - 1}) & \leq Q^1_i(\state_i, \ac^*_{i : i + \numqvi - 1}) - \gap^\numqvi_i(\state_i) \nonumber \\
    Q^1_i(\state_i, \ac^*_{i : i + \numqvi - 1}) - Q^1_i(\state_i, \ac_{i : i + \numqvi - 1}) & \geq \gap^\numqvi_i(\state_i). \label{eq:z_mean_bound}
\end{align}
Now, consider the random variables $\hat{Q}_i(\state_i, \ac_i, \ac_{i + 1 : \numqvi - 1})$ and $\hat{Q}_i(\state_i, \ac_{i : i + \numqvi - 1}^*)$. Let $X_j = \sum_{\eptime = i}^\horizon \gamma^{\eptime - i} \reward(\state^j_\eptime, \ac^j_\eptime)$ be the discounted reward from timestep $i$ in the $j$th episode used to estimate $\hat{Q}_i(\state_i, \ac_i, \ac_{i + 1: i + \numqvi - 1})$; let $Y_j$ be the equivalent for estimating $\hat{Q}_i(\state_i \ac_{i : i + \numqvi - 1}^*)$. Then we can write
\begin{equation*}
    \hat{Q}_i(\state_i, \ac_i, \ac_{i + 1 : i + \numqvi - 1}) = \frac{1}{\numep{}} \sum_{j = 1}^{\numep{}} X_j
    \qquad
    \hat{Q}_i(\state_i, \ac_{i : i + \numqvi - 1}^*) = \frac{1}{\numep{}} \sum_{j = 1}^{\numep{}} Y_j.
\end{equation*}
By the definition of the optimal value function $V^*$ and the assumption that all rewards are nonnegative, we have that $X_j, Y_j \in \left[ 0, V^*_i(\state_i) \right]$. We also know that the expectations of the $X_j$ and $Y_j$ are bounded:
\begin{align*}
    \EE[X_j] & = Q^1_i(\state_i, \ac_{i : i + \numqvi - 1}) \leq Q^\numqvi_i(\state_i, \ac_i) \leq \max_{\ac_i} Q^\numqvi_i(\state_i, \ac_i). \\
    \EE[Y_j] & = Q^1_i(\state_i, \ac^*_{i : i + \numqvi - 1}) = Q^\numqvi_i(\state_i, \ac_i^*) \leq \max_{\ac_i} Q^\numqvi_i(\state_i, \ac_i).
\end{align*}
The variance of a random variable with mean $\mu$ bounded on an interval $[\alpha, \beta]$ is at most $(\beta - \mu)(\mu - \alpha)$. This means we can bound the variance of $X_j$ and $Y_j$ as well:
\begin{equation*}
    \Var(X_j) \leq \max_{\ac_i} Q^\numqvi_i(\state_i, \ac_i) V^*_i(\state_i)
    \qquad
    \Var(Y_j) \leq \max_{\ac_i} Q^\numqvi_i(\state_i, \ac_i) V^*_i(\state_i).
\end{equation*}
Now define
\begin{equation*}
    Z = \hat{Q}^1_i(\state_i, \ac^*_{i : i + \numqvi - 1}) - \hat{Q}^1_i(\state_i,  \ac_i, \ac_{i + 1: i + \numqvi - 1}) = \frac{1}{\numep{}} \sum_{j = 1}^{\numep{}} Y_j - X_j = \frac{1}{\numep{}} \sum_{j = 1}^{\numep{}} Z_j,
\end{equation*}
where $Z_j = Y_j - X_j$.
Since $X_j$ and $Y_j$ are independent,
\begin{equation*}
    \Var(Z_j) = \Var(X_j) + \Var(Y_j) \leq 2 \max_{\ac_i} Q^\numqvi_i(\state_i, \ac_i) V^*_i(\state_i).
\end{equation*}
Also define centered versions $\bar{Z}_j = Z_j - \EE[Z_j]$ and $\bar{Z} = Z - \EE[Z]$. By (\ref{eq:z_mean_bound}), $\EE[Z] \geq \gap_i^\numqvi(\state_i)$. Furthermore, since $X_j, Y_j \in [0, V^*_i(\state_i)]$, we also know that $|Z_j| \leq  V^*_i(\state_i)$ and $|\bar{Z}_j| \leq V^*_i(\state_i) + \EE[Z] \leq 2 V^*_i(\state_i)$.

We can now finally apply Bernstein's inequality to bound the probability of one term in (\ref{eq:qirl_finds_suboptimal_action_seq}):
\begin{align}
    & \PP \left( \hat{Q}_i(\state_i, \ac_i, \ac_{i + 1 : \numqvi - 1}) \geq \hat{Q}_i(\state_i, \ac_{i : i + \numqvi - 1}^*) \right) \nonumber \\
    & \quad = \PP \left( Z \leq 0 \right) \nonumber \\
    & \quad = \PP \left( \bar{Z} \leq -\EE[Z] \right) \nonumber \\
    & \quad \leq \PP \left( \bar{Z} \leq -\gap_i^\numqvi(\state_i) \right) \nonumber \\
    & \quad \overset{\text{(i)}}{\leq} \exp \left\{ \frac{-\frac{1}{2} \numep{} \gap_i^\numqvi(\state_i)^2}{\Var(Z_j) + \frac{2}{3} V^*_i(\state_i) \gap_i^\numqvi(\state_i)} \right\} \nonumber \\
    & \quad \overset{\text{(ii)}}{\leq} \exp \left\{ \frac{-\frac{1}{2} \numep{} \gap_i^\numqvi(\state_i)^2}{2 \max_{\ac_i} Q^\numqvi_i(\state_i, \ac_i) V^*_i(\state_i) + \frac{2}{3} V^*_i(\state_i) \gap_i^\numqvi(\state_i)} \right\} \nonumber \\
    & \quad \overset{\text{(iii)}}{\leq} \exp \left\{ \frac{- \numep{} \gap_i^\numqvi(\state_i)^2}{6 \max_{\ac_i} Q^\numqvi_i(\state_i, \ac_i) V^*_i(\state_i)} \right\} \nonumber \\
    & \quad \overset{\text{(iv)}}{\leq} \exp \left( -\log (2 \horizon \acsize^\numqvi) \right) \nonumber \\
    & \quad = \frac{1}{2 \horizon \acsize^\numqvi}. \label{eq:bernstein_for_action_seq}
\end{align}
Here, (i) is a direct application of Bernstein's inequality to the sum $\bar{Z} = \frac{1}{\numep{}} \sum_{j = 1}^{\numep{}} \bar{Z}_j$. (ii) uses the bound on $\Var(Z_j) = \Var(\bar{Z}_j)$ and (iii) uses the fact that $\gap^\numqvi_i(\state_i) \leq \max_{\ac_i} Q^\numqvi_i(\state_i, \ac_i)$ by definition of the gap $\gap^\numqvi_i$. Finally, (iv) uses the definition of $\numep{}$; since $\state_i \in \statespace^\text{opt}_i$ by assumption,
\begin{align*}
    \numep{} \geq \log \left(2 \horizon \acsize^\numqvi\right) \max_{\ac_i \in \acspace} \frac{6 Q^\numqvi_\eptime(\state_i, \ac_i) V^*_\eptime(\state_i)}{\gap_\eptime^\numqvi(\state_i)^2}.
\end{align*}

Applying (\ref{eq:bernstein_for_action_seq}) to (\ref{eq:qirl_finds_suboptimal_action_seq}) gives
\begin{align*}
    & \PP \left( \exists \ac_i \in  \arg \max_{\ac_i \in \acspace}
    \hat{Q}^\numqvi_i(\state_i, \ac_i) \; : \; \ac_i \notin \arg \max_{\ac_i \in \acspace}
    Q^\numqvi_i(\state_i, \ac_i) \right) \\
    & \quad \leq \sum_{\ac_i \notin \arg \max_{\ac_i \in \acspace} Q^\numqvi_i(\state_i, \ac_i), \ac_{i + 1 : i + \numqvi - 1} \in \acspace^{\numqvi - 1}}
    \frac{1}{2 \horizon \acsize^\numqvi} \\
    & \quad \leq \frac{1}{2 \horizon},
\end{align*}
which is the only thing left that is needed to complete the proof.
\end{proof}

\section{Additional theoretical results concerning the effective horizon}
\label{sec:additional_efhorizon_theory}

In this appendix, we present some additional theoretical results concerning the effective horizon. First, we explore two algorithms---one in the style of policy gradient and one similar to fitted Q-iteration---whose sample complexities can also be bounded by a quantity related to the effective horizon. Then we show conditions under which the effective horizon is small, as well as information-theoretic lower bounds for the sample complexity of RL in terms of the effective horizon.

\subsection{PG-\algac and FQI-\algac}
\label{sec:variants}

The two algorithms we introduce in this section, PG-\algac and FQI-\algac, can be viewed as a bridge between \algac, which we use to define the effective horizon, and PPO and DQN, the deep RL algorithms whose performance we predict using the effective horizon in Section \ref{sec:experiments}. They help to explain why the effective horizon is not only useful for understanding the performance of \algac, but also other RL algorithms.

We will actually give sample complexity bounds on PG-\algac and FQI-\algac in terms of the bound in Theorem \ref{thm:deterministicefhorizon}, rather than the effective horizon itself. Supposing that the MDP is $\numqvi$-QVI-solvable, define
\begin{equation*}
    \efhorizonbound{\numqvi} = \numqvi \nonumber + \max_{\eptime \in [\horizon], \state \in \statespace^\text{opt}_i, \ac \in \acspace}
        \log_\acsize \left( \frac{Q^\numqvi_\eptime(\state, \ac) V^*_\eptime(\state)}{\gap_\eptime^\numqvi(\state)^2} \right) + \log_\acsize 6 \log \left(2 \horizon \acsize^\numqvi\right).
\end{equation*}
Our bounds will also depend on a quantity measuring how far the exploration policy $\expolicy$ is from the uniformly random policy $\randpolicy$:
\begin{equation*}
    \left \| \frac{\randpolicy}{\expolicy} \right \|_\infty = \max_{(\eptime, \state, \ac) \in [\horizon] \times \statespace \times \acspace} \frac{1}{\acsize \: \expolicy_\eptime(\ac \mid \state)}.
\end{equation*}
$\left \| \frac{\randpolicy}{\expolicy} \right \|_\infty = 1$ in the case when $\expolicy = \randpolicy$, and increases as the smallest probabilities $\expolicy$ assigns to actions becomes smaller.

We now introduce the first algorithm, PG-\algac.

\begin{algorithm}[H]
    \caption{The PG-\algac algorithm.}
    \begin{algorithmic}[1]
    \Procedure{PG-\algac}{$\numep{}$}
        \State $\policy \gets \expolicy$.
        \For{$i = 1, \hdots, \horizon$}
            \State Sample $\numep{}$ episodes following $\policy$. \label{line:pg_sample}
            \For{$\ac \in \acspace$}
                \State $\hat{\nabla}_i (\ac \mid \state_i) \gets \frac{1}{\numep{}} \sum_{j = 1}^{\numep{}} \frac{\II_{\ac^j_i = \ac}}{\policy_i(\ac \mid \state_i)} \sum_{\eptime = i}^\horizon \discount^{\eptime - i} \reward(\state_\eptime^j, \ac_\eptime^j)$.
            \EndFor
            \State $\policy_i(\state_i) \gets \arg \max_{\ac \in \acspace} \hat{\nabla}_i (\ac \mid \state_i)$. \label{line:pg_opt}
        \EndFor
		\State \Return $\policy$
    \EndProcedure
    \end{algorithmic}
    \label{alg:pg_qirl}
\end{algorithm}

PG-\algac resembles the REINFORCE algorithm \citep{williams_simple_1992}, which gave rise to other policy gradient-based algorithms like PPO. At each iteration, Algorithm \ref{alg:pg_qirl} first samples a number of episodes following its current policy (line \ref{line:pg_sample}). Then, it computes a an approximate gradient over the policy parameters---in this case, just the action probabilities $\policy_i(\ac \mid \state_i)$---via the so-called ``policy gradient theorem,'' which states
\begin{align*}
    \nabla_{\policy_i(\cdot \mid \state_i)} \pret(\policy)
    & \approx \frac{1}{\numep{}} \sum_{j = 1}^{\numep{}} \nabla_{\policy_i(\cdot \mid \state_i)} \log \policy_i(\ac^j_i \mid \state_i) \sum_{\eptime = i}^\horizon \discount^{\eptime - i} \reward(\state_\eptime^j, \ac_\eptime^j) \\
    & = \frac{1}{\numep{}} \sum_{j = 1}^{\numep{}} \frac{\nabla_{\policy_i(\cdot \mid \state_i)} \policy_i(\ac^j_i \mid \state_i)}{\policy_i(\ac^j_i \mid \state_i)} \sum_{\eptime = i}^\horizon \discount^{\eptime - i} \reward(\state_\eptime^j, \ac_\eptime^j) \\
    & = \hat{\nabla}_i (\cdot \mid \state_i).
\end{align*}
Then, in line \ref{line:pg_opt}, Algorithm \ref{alg:pg_qirl} applies optimization to $\policy$ based on its estimate of the gradient. In this case, it optimizes until $\policy_i$ assigns all probability to only one action $\policy_i(\state_i)$ at $\state_i$.

\begin{theorem}[Sample complexity of PG-\algac]
    \label{thm:pg_sampcomplexity}
    Suppose that an MDP is 1-QVI-solvable and that all rewards are nonnegative. Then the sample complexity of PG-\algac is at most
    \begin{equation*}
        2 \horizon^2 \acsize^{\efhorizonbound{1}} \left \| \frac{\randpolicy}{\expolicy} \right \|_\infty.
    \end{equation*}
\end{theorem}
\begin{proof}
Let
\begin{equation*}
    \numep{} = \acsize^{\efhorizonbound{1}} \left \| \frac{\randpolicy}{\expolicy} \right \|_\infty
    = 12 \acsize \log \left(2 \horizon \acsize \right) \max_{\eptime \in [\horizon], \state \in \statespace^\text{opt}_i, \ac \in \acspace}
        \left( \frac{Q^\numqvi_\eptime(\state, \ac) V^*_\eptime(\state)}{\gap_\eptime^1(\state)^2} \right) \left\| \frac{\randpolicy}{\expolicy} \right \|_\infty.
\end{equation*}
We will show that Algorithm \ref{alg:pg_qirl} converges with probability at least $1/2$ with this choice of parameter, giving the sample complexity bound in the theorem since the algorithm clearly samples $\horizon^2 \numep{}$ total timesteps from the environment.

Similarly to the proof of Theorem \ref{thm:deterministicefhorizon}, we will prove this by showing that with probability at least $1 - 1/(2 \horizon)$, at each iteration $\policy_i(\state_i) \in \arg \max_{\ac \in \acspace} Q^{\expolicy}_i(\state_i, \ac)$. This gives $\policy \in \policies(Q^1)$, which by 1-QVI-solvability means $\policy$ must be optimal.

Consider the $i$th iteration of Algorithm \ref{alg:pg_qirl}. Define for each $\ac \in \acspace$ and $j \in [\numep{}]$ the random variable
\begin{equation*}
    X_j(\ac) = \frac{\II_{\ac^j_i = \ac}}{\policy_i(\ac \mid \state_i)} \sum_{\eptime = i}^\horizon \discount^{\eptime - i} \reward(\state_\eptime^j, \ac_\eptime^j).
\end{equation*}
First, can see that
\begin{equation*}
    0 \leq X_j(\ac)
    \leq \frac{V^*_i(\state_i)}{\policy_i(\ac \mid \state_i)}
    = \frac{V^*_i(\state_i)}{\expolicy_i(\ac \mid \state_i)}
    \leq \acsize \: V^*_i(\state_i) \left\| \frac{\randpolicy}{\expolicy} \right\|_\infty,
\end{equation*}
since $\policy_i = \expolicy_i$ until line \ref{line:pg_opt}.

Second, we have
\begin{equation*}
    \EE\left[ X_j(\ac) \right] = \frac{1}{\policy_i(\ac \mid \state_i)} \PP\left(\ac^j_i = \ac\right) \EE\left [ \sum_{\eptime = i}^\horizon \discount^{\eptime - i} \reward(\state_\eptime^j, \ac_\eptime^j) \mid \ac^j_i = \ac \right] = Q^{\expolicy}_i(\state_i, \ac).
\end{equation*}
Finally, using same the reasoning as in the proof of Theorem \ref{alg:pg_qirl}, we can bound the variance of $X_j(\ac)$:
\begin{equation*}
    \Var \left(X_j\right) \leq \acsize \: V^*_i(\state_i) \left\| \frac{\randpolicy}{\expolicy} \right\|_\infty \EE \left[ X_j \right]
    = \acsize \: Q^{\expolicy}_i(\state_i, \ac) V^*_i(\state_i) \left\| \frac{\randpolicy}{\expolicy} \right\|_\infty.
\end{equation*}

We now apply Bernstein's inequality to
\begin{equation*}
    \hat{\nabla}_i(\ac \mid \state_i) = \frac{1}{\numep{}} \sum_{j = 1}^{\numep{}} X_j(\ac)
\end{equation*}
for each $\ac \in \acspace$. If $\ac \in \arg \max_{\ac \in \acspace} Q^{\expolicy}_i(\state_i, \ac)$, we apply a lower tail bound:
\begin{align*}
    & \PP\left( \hat{\nabla}_i(\ac \mid \state_i) \leq Q^{\expolicy}_i(\state_i, \ac) - \frac{1}{2} \gap^1_i(\state_i) \right) \\
    & \quad \leq \exp \left\{ - \frac{ \numep{} \left( \gap^1_i(\state_i) \right)^2 / 8}{\left(Q^{\expolicy}_i(\state_i, \ac) + \frac{1}{3} \gap^1_i(\state_i) \right) V^*_i(\state_i) \left\| \frac{\randpolicy}{\expolicy} \right\|_\infty } \right\} \\
    & \quad \leq \exp \left\{ - \frac{ \numep{} \left( \gap^1_i(\state_i) \right)^2}{12 Q^{\expolicy}_i(\state_i, \ac) V^*_i(\state_i) \left\| \frac{\randpolicy}{\expolicy} \right\|_\infty } \right\} \\
    & \quad \leq \frac{1}{2 \horizon \acsize}.
\end{align*}
If $\ac \notin \arg \max_{\ac \in \acspace} Q^{\expolicy}_i(\state_i, \ac)$, we apply an identical upper tail bound:
\begin{align*}
    \PP\left( \hat{\nabla}_i(\ac \mid \state_i) \geq Q^{\expolicy}_i(\state_i, \ac) + \frac{1}{2} \gap^1_i(\state_i) \right) \leq \frac{1}{2 \horizon \acsize}.
\end{align*}
These tail bounds hold simultaneously for all actions with probability at least $1 - \frac{1}{2 \horizon}$. Furthermore, assuming they hold and using the definition of the gap $\gap^1_i(\state_i)$, it must be that
\begin{equation*}
    \arg \max_{\ac \in \acspace} \hat{\nabla}_i (\ac \mid \state_i) \subseteq \arg \max_{\ac \in \acspace} Q^{\expolicy}_i(\state_i, \ac),
\end{equation*}
which is enough to show that $\policy_i(\state_i) \in \arg \max_{\ac \in \acspace} Q^{\expolicy}_i(\state_i, \ac)$ with probability at least $1 - \frac{1}{2 \horizon}$ and thus prove the theorem.
\end{proof}

Now that we have seen that PG-\algac enjoys similar sample complexity bounds to \algac in the common case that an MDP is 1-QVI-solvable, we introduce FQI-\algac. FQI-\algac derives its name from fitted Q-iteration (FQI), which was originally proposed by \citet{ernst_tree-based_2005}. DQN was inspired by neural FQI \citep{riedmiller_neural_2005}, so FQI-\algac provides a natural connection to DQN.

\begin{algorithm}[H]
    \caption{The FQI-\algac algorithm.}
    \begin{algorithmic}[1]
    \Procedure{FQI-\algac}{$\numqvi, \numep{}$}
        \For{$i = 1, \hdots, \horizon$}
            \State Sample $\acsize^{\numqvi} \numep{}$ episodes, following $\policy$ for timesteps $1$ to $i - 1$ and then $\expolicy$. \label{line:fqi_sample}
            \State $\hat{Q}^1_{i + \numqvi - 1} \gets \arg \min_{\hat{Q}^1_{i + \numqvi - 1}} \frac{1}{\acsize^\numqvi \numep{}} \sum_{j = 1}^{\acsize^\numqvi \numep{}} \left( \hat{Q}^1_{i + \numqvi - 1}(\state^j_{i + \numqvi - 1}, \ac^j_{i + \numqvi - 1}) - \sum_{\eptime = i + \numqvi - 1}^\horizon \reward(\state_\eptime^j, \ac_\eptime^j) \right)^2$. \label{line:fqi_fit_1}
            \For{$\eptime = i + \numqvi - 2, \hdots, i$}
                \State $\hat{Q}^{i + \numqvi - \eptime}_\eptime \gets \arg \min_{\hat{Q}^{i + \numqvi - \eptime}_\eptime} \frac{1}{\acsize^\numqvi \numep{}} \sum_{j = 1}^{\acsize^\numqvi \numep{}} \left( \hat{Q}^{i + \numqvi - \eptime}_\eptime(\state^j_\eptime, \ac^j_\eptime) - \reward(\state^j_\eptime, \ac^j_\eptime) - \max_{\ac' \in \acspace} \hat{Q}^{i + \numqvi - \eptime - 1}_{\eptime + 1}(\state^j_{\eptime + 1}, \ac') \right)^2$. \label{line:fqi_fit_2}
            \EndFor
            \State $\policy_i(\state_i) \gets \arg \max_{\ac \in \acspace} \hat{Q}^\numqvi_i(\state_i, \ac)$. \label{line:fqi_opt}
        \EndFor
		\State \Return $\policy$
    \EndProcedure
    \end{algorithmic}
    \label{alg:fqi_qirl}
\end{algorithm}

FQI-\algac iteratively constructs a series of Q-functions at each iteration by minimizing a mean-squared temporal difference error loss, similar to FQI and DQN.

\begin{theorem}[Sample complexity of FQI-\algac]
    \label{thm:fqi_sampcomplexity}
    Suppose that an MDP is $\numqvi$-QVI-solvable and that all rewards are nonnegative. Then the sample complexity of FQI-\algac is at most
    \begin{equation*}
        \horizon^2 \left \| \frac{\randpolicy}{\expolicy} \right \|_\infty^\numqvi \max \left\{ 4 \acsize^{\efhorizonbound{\numqvi}}, 10 \log(4 \horizon \acsize^\numqvi) \right\}.
    \end{equation*}
\end{theorem}
\begin{proof}
Let
\begin{equation*}
    \numep{} = \left \| \frac{\randpolicy}{\expolicy} \right \|_\infty^\numqvi \max \left\{ 24 \log \left(2 \horizon \acsize \right) \max_{\eptime \in [\horizon], \state \in \statespace^\text{opt}_i, \ac \in \acspace} \left( \frac{Q^\numqvi_\eptime(\state, \ac) V^*_\eptime(\state)}{\gap_\eptime^1(\state)^2} \right), \frac{10 \log(4 \horizon \acsize^\numqvi)}{\acsize^\numqvi} \right\}.
\end{equation*}
We will show that FQI-\algac with parameters $\numqvi$ and $\numep{}$ will return an optimal policy with probability at least $1/2$. Since FQI-\algac samples a number of timesteps from the environment equal to $\horizon^2 \acsize^\numqvi \numep{}$, this will prove the bound in the theorem.

Consider the $i$th iteration of Algorithm \ref{alg:pg_qirl}. We will show that with probability at least $1 - 1/(2 \horizon)$, for every $\state_{i + \numqvi - 1} \in \statespace$ reachable at timestep $i + \numqvi - 1$ starting from $\state_i$, and for every action $\ac_{i + \numqvi - 1}$
\begin{equation}
    \label{eq:fqi_concentration}
    \left| \hat{Q}^1_{i + \numqvi - 1}(\state^j_{i + \numqvi - 1}, \ac^j_{i + \numqvi - 1}) - Q^1_{i + \numqvi - 1}(\state^j_{i + \numqvi - 1}, \ac^j_{i + \numqvi - 1}) \right| < \frac{\gap_i^1(\state_i)}{2}.
\end{equation}
To prove (\ref{eq:fqi_concentration}), it is first helpful to write an explicit formula fitted Q-value, assuming that the loss in line \ref{line:fqi_fit_1} of Algorithm \ref{alg:fqi_qirl} is minimized:
\begin{equation*}
    \hat{Q}^1_{i + \numqvi - 1}(\state, \ac) =
    \frac{\sum_{j = 1}^{\acsize^\numqvi \numep{}} \II_{\state^j_{i + \numqvi - 1} = \state \wedge \ac^j_{i + \numqvi - 1} = \ac} \sum_{\eptime = i + \numqvi - 1}^\horizon \reward(\state_\eptime^j, \ac_\eptime^j)}{\sum_{j = 1}^{\acsize^\numqvi \numep{}} \II_{\state^j_{i + \numqvi - 1} = \state \wedge \ac^j_{i + \numqvi - 1} = \ac}}.
\end{equation*}
That is, the Q-value is a simple average of several reward-to-go values, each of which has expectation $Q^1_{i + \numqvi - 1}(\state, \ac)$. 
The probability of reaching some reachable state-action pair $(\state, \ac)$ at timestep $i + \numqvi - 1$ must be at least $\acsize^{-\numqvi} \left\| \frac{\randpolicy}{\expolicy} \right\|_\infty^{-\numqvi}$. Thus, we can bound the sample size below via a concentration inequality for binomial random variables:
\begin{equation*}
    \PP \left( \sum_{j = 1}^{\acsize^\numqvi \numep{}} \II_{\state^j_{i + \numqvi - 1} = \state \wedge \ac^j_{i + \numqvi - 1} = \ac} < 12 \log \left(2 \horizon \acsize \right) \max_{\ac_i \in \acspace} \frac{Q^\numqvi_i(\state_i, \ac_i) V^*_i(\state_i)}{\gap_i^1(\state_i)^2} \right)
    \leq \exp \{ \frac{-3 (10 \log (4 \horizon \acsize^\numqvi))}{28} \} \leq \frac{1}{4 \horizon \acsize^\numqvi}.
\end{equation*}
If the sample size is at least $12 \log \left(2 \horizon \acsize \right) \max_{\ac_i \in \acspace} \frac{Q^\numqvi_i(\state_i, \ac_i) V^*_i(\state_i)}{\gap_i^1(\state_i)^2}$, then Bernstein's inequality (as applied in the proofs of Theorems \ref{thm:deterministicefhorizon} and \ref{thm:pg_sampcomplexity}) gives
\begin{align*}
    & \PP \left( \left| \hat{Q}^1_{i + \numqvi - 1}(\state^j_{i + \numqvi - 1}, \ac^j_{i + \numqvi - 1}) - Q^1_{i + \numqvi - 1}(\state^j_{i + \numqvi - 1}, \ac^j_{i + \numqvi - 1}) \right| \geq \frac{\gap_i^1(\state_i)}{2} \right) \leq \frac{1}{4 \horizon \acsize^\numqvi}.
\end{align*}
Thus, taking a union bound, we have that the probability (\ref{eq:fqi_concentration}) does \emph{not} hold for some reachable state-action pair at timestep $i + \numqvi - 1$ must be at most $1/(2 \horizon)$, since there can be at most $\acsize^\numqvi$ such pairs.

We will now show by induction that given that (\ref{eq:fqi_concentration}) holds for all reachable state-action pairs,
\begin{equation}
    \label{eq:fqi_concentration_2}
    \left| \hat{Q}^{i + \numqvi - \eptime}_\eptime(\state^j_{i + \numqvi - 1}, \ac^j_{i + \numqvi - 1}) - \hat{Q}^{i + \numqvi - \eptime}_\eptime{i + \numqvi - 1}(\state^j_{i + \numqvi - 1}, \ac^j_{i + \numqvi - 1}) \right| < \frac{\gap_i^1(\state_i)}{2}.
\end{equation}
holds for $\eptime = i + \numqvi - 1, \hdots, i$ for all reachable state-action pairs at $\eptime$. The base case of $\eptime = i + \numqvi - 1$ is already taken care of, so we only need to show the inductive step.

Assume (\ref{eq:fqi_concentration_2}) holds for all reachable state-action pairs at $\eptime + 1$. We can write an explicit formula for the fitted Q-values at timestep $\eptime$, given that the loss function on line \ref{line:fqi_fit_2} of Algorithm \ref{alg:fqi_qirl} is minimized:
\begin{align*}
    \hat{Q}^{i + \numqvi - \eptime}_\eptime(\state, \ac) & = 
    \frac{\sum_{j = 1}^{\acsize^\numqvi \numep{}} \II_{\state^j_\eptime = \state \wedge \ac^j_\eptime = \ac} \left( \reward(\state^j_\eptime, \ac^j_\eptime) + \max_{\ac' \in \acspace} \hat{Q}^{i + \numqvi - \eptime - 1}_{\eptime + 1}(\state^j_{\eptime + 1}, \ac') \right)}{\sum_{j = 1}^{\acsize^\numqvi \numep{}} \II_{\state^j_\eptime = \state \wedge \ac^j_\eptime = \ac}} \\
    & = \reward(\state, \ac) + \max_{\ac' \in \acspace} \hat{Q}^{i + \numqvi - \eptime - 1}_{\eptime + 1}(\dynamics(\state, \ac), \ac').
\end{align*}
Given that (\ref{eq:fqi_concentration_2}) holds for each pair $(\dynamics(\state, \ac), \ac')$, it is now easy to see that (\ref{eq:fqi_concentration_2}) must hold for $\eptime$ as well.

By induction (\ref{eq:fqi_concentration_2}) must hold for $\eptime = i$ with probability at least $1 - 1 / (2 \horizon)$. Given that it holds, and by definition of the gap, this implies that $\policy_i(\state_i) \in \arg\max_{\ac \in \acspace} Q^{\numqvi}_i(\state_i, \ac)$. Thus with probability at least $1/2$, $\policy \in \policies(Q^\numqvi)$. By the assumption that the MDP is $\numqvi$-QVI solvable, $\policy$ must be optimal with probability at least $1/2$.
\end{proof}

\subsection{Goal MDPs}

Now, we will prove bounds on the effective horizon in one particular class of MDPs: goal MDPs.
\begin{definition}[Goal MDP]
    \label{definition:goal_mdp}
    An MDP is considered a \emph{goal MDP} if there is some set of goal states $\statespace_\text{goal}$ which are absorbing, i.e., $\dynamics(\state, \ac) = \state$ for every $\state \in \statespace_\text{goal}$, and furthermore the reward function is of the form
    \begin{equation*}
        \reward(\state, \ac) = \begin{cases}
            1 & \quad \state \notin \statespace_\text{goal} \wedge \dynamics(\state, \ac) \in \statespace_\text{goal} \\
            0 & \quad \text{otherwise.}
        \end{cases}
    \end{equation*}
\end{definition}
That is, in a goal MDP reward is only received for reaching some set of goal states; the total episode reward is 1 if a goal state is reached and 0 otherwise. As an example, all of the Minigrid environments in \datasetname are goal MDPs. We can show the following bound on the effective horizon in goal MDPs.
\begin{theorem}[The effective horizon in goal MDPs]
    \label{thm:goal_efhorizon}
    Suppose that $\expolicy(\ac \mid \state) > 0$ for all $\state \in \statespace$ and $\ac \in \acspace$. Then any goal MDP is 1-QVI-solvable. Furthermore, suppose that there is some $p > 0$ such that, for all timesteps $\eptime \in [\horizon]$ and all state-action pairs $\state_\eptime, \ac_\eptime$ at that timestep from which a goal state can be reached,
    \begin{equation*}
        \PP_{\expolicy}\left(\state_\horizon \in \statespace_\text{goal} \mid \state_\eptime, \ac_\eptime\right) \geq p.
    \end{equation*}
    Then the effective horizon can be bounded as
    \begin{equation}
        \label{eq:goal_efhorizon}
        \efhorizon{} \leq 1 + \log_\acsize \frac{\log (2 \horizon)}{p}.
    \end{equation}
\end{theorem}
Before we see the proof, note that Theorem \ref{thm:goal_efhorizon} agrees with our intuition that it should be harder to find an optimal policy for a goal MDP when it is less likely that the exploration policy reaches the goal, i.e., when $p$ is smaller.

For instance, consider the MDP in Figure \ref{fig:sparse_reward_example}. In this MDP, the minimum probability of reaching the goal with the random exploration policy after taking some action is exponentially small: $p = 1 / \acsize^{\horizon - 1}$. Applying Theorem \ref{thm:goal_efhorizon} gives a bound of $\efhorizon{} \leq \horizon + \log_\acsize \log(2 \horizon)$.

\begin{figure}[t]
    \centering
    \includegraphics[width=1.5in]{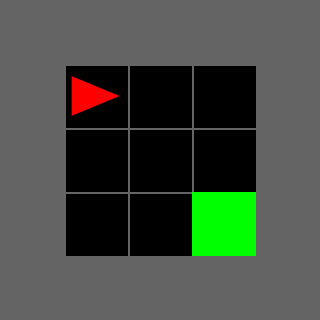}
    \caption{Empty-5x5, one of the Minigrid MDPs from \datasetname and an example of a goal MDP (Definition \ref{definition:goal_mdp}). The agent (red triangle) can turn left, turn right, or go forward, and its goal is to reach the green square, which gives a reward of 1.}
    \label{fig:empty_5x5}
\end{figure}

In contrast, consider the Minigrid gridworld in Figure \ref{fig:empty_5x5} from \datasetname. Here we can bound $p \approx 0.00137$, which gives $\efhorizon{} \leq 1 + \log_\acsize (1 / p) + \log_\acsize \log(2 \horizon) = 1 + \log_3 729 + \log_3 \log(200) \leq 7 + 1.52 = 8.52 \ll \horizon = 100$. The techniques in Appendix \ref{sec:tighter_bounds} give a much tighter bound of $\efhorizon{} \leq 1.64$.

\begin{proof}
We can assume that $V^*_1(\state_1) = 1$, since otherwise every trajectory in the MDP gives reward 0 and thus the effective horizon is trivially bounded. First, we will show that
\begin{equation}
    \label{eq:q_positive_means_optimal}
    Q^{\expolicy}_\eptime(\state, \ac) > 0 \qquad \Leftrightarrow \qquad Q^*_\eptime(\state, \ac) = 1.
\end{equation}
This is enough to imply the MDP is 1-QVI-solvable, since any policy in $\policies(Q^{\expolicy})$ must take only actions with $Q^{\expolicy}_\eptime(\state, \ac) > 0$, which must be optimal according to (\ref{eq:q_positive_means_optimal}).

To show that (\ref{eq:q_positive_means_optimal}) holds, first consider the $\Rightarrow$ implication. Assume $Q^{\expolicy}_\eptime(\state, \ac) > 0$ and by way of contradiction suppose that $Q^*_\eptime(\state, \ac) \neq 1$, which means that $Q^*_\eptime(\state, \ac) = 0$. Clearly this cannot happen since this would imply $Q^{\expolicy}_\eptime(\state, \ac) \leq 0$. Now, consider the $\Leftarrow$ direction. If $Q^*_\eptime(\state, \ac) = 1$, then there must be some sequence of actions starting with $\ac$ which leads from $\state$ to a goal state. By assumption, $\expolicy$ assigns positive probability to each action in this sequence. Thus $Q^{\expolicy}_\eptime(\state, \ac) = \PP_{\expolicy}(\state_\horizon \in \statespace_\text{goal} \mid \state, \ac) > 0$.

Next, we will prove the bound on $\efhorizon{}$ from (\ref{eq:goal_efhorizon}). From (\ref{eq:q_positive_means_optimal}), we can see that Algorithm \ref{alg:qirl} (\algac) will return an optimal policy for the MDP as long as at each iteration $i$ it picks some action $\ac_i$ with $Q^{\expolicy}_i(\state_i, \ac_i) > 0$. In turn, this will happen as long as $\hat{Q}^1_i(\state_i, \ac_i) > 0$ for some such $\ac_i$, since $\hat{Q}^1_i(\state_i, \ac)$ must be 0 for any suboptimal $\ac$.

Thus, we need only show that
\begin{equation}
    \label{eq:goal_iter_fail_prob}
    \PP \left( \exists \ac_i \in \acspace \quad \hat{Q}^1_i(\state_i, \ac_i) > 0 \right) \geq 1 - \frac{1}{2 \horizon}
\end{equation}
holds for each $i$ when $\numep{} \geq \log(2 \horizon) / p$. This will allow us to conclude via a union bound that Algorithm \ref{alg:qirl} will find an optimal policy with probability at least $1/2$ in this case, which gives the desired bound on the effective horizon.

To show (\ref{eq:goal_iter_fail_prob}), consider iteration $i$ of Algorithm \ref{alg:qirl} and let $\ac_i \in \acspace$ such that $Q^{\randpolicy}_i(\state_i, \ac_i) > 0$. We can assume that such an action exists as long as Algorithm \ref{alg:qirl} has succeeded in iterations previous to $i$. Let $X_j = \sum_{\eptime = i}^\horizon \gamma^{\eptime - i} \reward(\state^j_\eptime, \ac^j_\eptime)$ be the reward-to-go from the $j$th episode sampled to evaluate $\hat{Q}^1_i(\state_i, \ac_i)$. By the definition of a goal MDP, each $X_j(\ac_i) \in \{0, 1\}$. Furthermore, since $\hat{Q}^1_i(\state_i, \ac_i) = \frac{1}{\numep{}} \sum_{j = 1}^{\numep{}} X_j$, then $\hat{Q}^1_i(\state_i, \ac) > 0$ as long as some $X_j = 1$. This implies
\begin{align*}
    & \PP \left( \exists \ac_i \in \acspace \quad \hat{Q}^1_i(\state_i, \ac_i) > 0 \right) \\
    & \quad \geq \PP \left( \hat{Q}^1_i(\state_i, \ac_i) > 0 \right) \\
    & \quad = \PP \Big( \exists  j \in [\numep{}] \quad X_j = 1 \Big) \\
    & \quad = 1 - \PP \Big( \forall  j \in [\numep{}] \quad X_j = 0 \Big) \\
    & \quad = 1 - \PP \Big( X_j = 0 \Big)^{\numep{}} \\
    & \quad \overset{\text{(i)}}{\geq} 1 - (1 - p)^{\numep{}} \\
    & \quad \geq 1 - \exp ( - \numep{} p) \\
    & \quad \geq 1 - \frac{1}{2 \horizon},
\end{align*}
the bound previously proposed in (\ref{eq:goal_iter_fail_prob}) which we argued gives the desired bound. (i) uses the assumption that $Q^{\expolicy}_i(\state_i, \ac_i) = \PP( X_j = 1) \geq p$.
\end{proof}

\subsection{Lower bounds}

Next, we will show that there are MDPs in which exponential dependence on the effective horizon is unavoidable. That is, in some cases there is an information-theoretic lower bound on the sample complexity of RL proportional to $\acsize^{\efhorizon{}}$.

\begin{theorem}
    \label{thm:deterministic_efhorizon_lower_bound}
    Fix $\horizon \geq 1$, $\acsize \geq 2$, and $\efhorizon{} \in [\horizon]$. Then for any RL algorithm, there is an MDP with $\acsize$ actions, horizon $\horizon$, and effective horizon at most $\efhorizon{}$ for which the algorithm's sample complexity is at least
    \begin{equation*}
        \horizon \lfloor \horizon / \efhorizon{} \rfloor \left( \lceil \acsize^{\efhorizon{}} / 2 \rceil - 1 \right)
        = \Omega( \horizon^2 \acsize^{\efhorizon{}} / \efhorizon{} ).
    \end{equation*}
\end{theorem}

Note that this matches the upper bound on the sample complexity of \algac given in Lemma \ref{lemma:efhorizon_sampcomplexity} up to a factor of roughly $2 \efhorizon{}$. When $\efhorizon{} = \horizon$, it exactly agrees with the lower bound given in Theorem \ref{thm:deterministicwc}.

\begin{proof}
The proof uses MDPs with the same state space and transition function as in Theorem \ref{thm:deterministicwc}. Define $\acsize^\horizon$ such MDPs which differ only in their reward functions:
\begin{equation*}
    \mathbb{M} = \{\mdp_{\ac_{1:\horizon}} \mid \ac_{1:\horizon} \in \acspace^\horizon \}
    \quad \text{where the MDP } \mdp_{\ac_{1:\horizon}} \text{ has } \reward(\state_{\ac_{1:\ell}'}, \ac') = \begin{cases}
        1 \quad & \ac_{1:\ell}', \ac' = \ac_{1:\ell + 1} \text{ and } \ell \equiv 0 \pmod{\efhorizon{}} \\
        0 \quad & \text{otherwise.}
    \end{cases}
\end{equation*}
That is, each MDP has a single optimal sequence of actions that gives reward 1 every $\efhorizon{}$ timesteps.

By the same argument as in the proof of Theorem \ref{thm:deterministicwc}, for any RL algorithm there must be some MDP in $\mathbb{M}$ such that after interacting with the environment for less than $\lceil \acsize^{\efhorizon{}} / 2 \rceil - 1$ episodes, the algorithm cannot with probability at least $1/2$ identify the optimal actions for timesteps $\eptime = 1$ to $\eptime = \efhorizon{}$. We can repeat this line of reasoning for timesteps $\eptime = \efhorizon{} + 1$ to $\eptime = 2 \efhorizon{}$, and so on for a total of $\lfloor \horizon / \efhorizon{} \rfloor$ steps, to show that with less than $\lfloor \horizon / \efhorizon{} \rfloor \left( \lceil \acsize^{\efhorizon{}} / 2 \rceil - 1 \right)$ episodes, there must be some MDP in $\mathbb{M}$ whose optimal action sequence cannot be identified with probability greater than $(1 / 2)^{\lfloor \horizon / \efhorizon{} \rfloor} \leq 1/2$. Thus, the sample complexity of the RL algorithm on this MDP must be at least
\begin{equation*}
    \horizon \lfloor \horizon / \efhorizon{} \rfloor \left( \lceil \acsize^{\efhorizon{}} / 2 \rceil - 1 \right),
\end{equation*}
which is the desired bound.

It only remains to be shown that the effective horizon of the MDPs in $\mathbb{M}$ is actually $\efhorizon{}$. To see why, consider running Algorithm \ref{alg:qirl} with $\numqvi = \efhorizon{}$ and $\numep{} = 1$. That is, at each iteration $i$, \algac will try all $\efhorizon{}$-length action sequences followed by actions from $\expolicy$. Then, it will pick the action sequence with the highest empirical reward-to-go. From the definition of the MDPs in $\mathbb{M}$, all action sequences starting with a suboptimal action must have empirical reward-to-go of 0. Furthermore, at least one $\efhorizon{}$-length action sequence starting with an optimal action must get reward-to-go of at least 1. Thus, \algac will with probability 1 choose an optimal action at each timestep. This means that the effective horizon must be at most $\efhorizon{} + \log_\acsize 1 = \efhorizon{}$.
\end{proof}

\section{Tighter bounds on the effective horizon}
\label{sec:tighter_bounds}

\newcommand{\failevent}{\mathcal{E}}

In Theorem \ref{thm:deterministicefhorizon}, we obtained bounds on the effective horizon and thus on the sample complexity of \algac. However, we find that the bounds given by Theorem \ref{thm:deterministicefhorizon} are often very loose compared to the empirical performance of \algac due to two factors. First, Theorem \ref{thm:deterministicefhorizon} requires considering the worst case of $Q^\numqvi_\eptime(\state, \ac) V^*_\eptime(\state) / \gap^\numqvi_\eptime(\state)^2$ over all optimal states. However, in many MDPs in our dataset, there are optimal states with extremely small gaps that in practice are almost never reached by \algac. When Theorem \ref{thm:deterministicefhorizon} is applied, these states make the sample complexity bounds very large despite \algac working well empirically. Second, Theorem \ref{thm:deterministicefhorizon} uses asymptotically tight techniques for bounding the sample complexity that can be quite loose for small sample sizes. Below, we describe the algorithm we use to provably bound the sample complexity of \algac (and thus the effective horizon) that gives much tighter results.

Consider the \algac algorithm as given in Algorithm \ref{alg:qirl}. Let $\ac_\eptime$ denote the random variable corresponding to the action ultimately chosen by the algorithm for timestep $\eptime$. Let $\state_\eptime$ denote the state reached by actions $\ac_1, \hdots, \ac_{\eptime - 1}$. Denote by $\PP_{\numep{}}$ the probability measure given by running the algorithm with parameter $\numep{}$. We would like to bound the probability that the algorithm does not achieve the optimal return in the MDP. Let this event by denoted as
\begin{align*}
    \failevent & \coloneqq \sum_{\eptime = 1}^\horizon \reward_\eptime(\state_\eptime, \ac_\eptime) < V^*_1(\state_1) \\
    & \Leftrightarrow \exists \eptime \quad \ac_\eptime \notin \acspace^*_\eptime(\state_\eptime),
\end{align*}
where $\acspace^*_\eptime(\state)$ denotes the set of optimal actions in state $\state$ at timestep $\eptime$, i.e.
\begin{equation*}
    \acspace^*_\eptime(\state) = \arg \max_{\ac \in \acspace} Q^*_\eptime(\state).
\end{equation*}
It is straightforward to see from Algorithm \ref{alg:qirl} that it requires $\horizon^2 \acsize^\numqvi \numep{}$ timesteps of interaction the environment. Thus the sample complexity of the algorithm is
\begin{equation*}
    \horizon^2 \acsize^\numqvi \numep{}
    \quad \text{where} \quad
    \numep{} = \min \{ \numep{} \in \mathbb{N} \mid \PP_{\numep{}}(\failevent) < 1/2 \}.
\end{equation*}
Clearly, we can upper bound the sample complexity using any $\numep{}$ such that the probability of failure is bounded as $\PP_{\numep{}}(\failevent) < 1/2$. To do so, for each value of $\numqvi$, we perform a binary search over values of $\numep{}$ from $1$ to $10^{100}$. For each possible $\numep{}$, we calculate an upper bound on $\PP_{\numep{}}(\failevent)$. If the upper bound is below $1/2$, we then search below $\numep{}$; if it is greater, we search above $\numep{}$. When the search has converged to a relative precision of $1/100$, we output $\horizon^2 \acsize^\numqvi \numep{}$ as the sample complexity and $\efhorizon{\numqvi} = \numqvi + \log_\acsize \numep{}$ as the effective horizon for that particular value of $\numqvi$.

\subsection{Upper bounding $\PP_{\numep{}}(\failevent)$}

We upper bound the failure probability $\PP_{\numep{}}(\failevent)$ recursively. Let $\mathcal{O}_\eptime$ denote the event that all actions taken before $\eptime$ have been optimal, i.e.,
\begin{equation*}
    \mathcal{O}_\eptime \coloneqq \forall \eptime' < \eptime \quad \ac_\eptime \in \acspace^*_\eptime(\state_\eptime).
\end{equation*}
To begin the recursion, note that at the final timestep,
\begin{equation*}
    \PP(\failevent \mid \mathcal{O}_\horizon, \state_\horizon, \ac_\horizon)
    = \II \{ \ac_\horizon \notin \acspace^*_\horizon(\state_\horizon) \}.
\end{equation*}
We will use two recursion rules: one from next states to state-action pairs and one from state-action pairs to states. The first rule is
\begin{equation}
    \label{eq:fail_prob_recursion}
    \PP(\failevent \mid \mathcal{O}_\eptime, \state_\eptime)
    = \sum_{\ac_\eptime \in \acspace} \PP(\failevent \mid \mathcal{O}_\eptime, \state_\eptime, \ac_\eptime) \PP(\ac_\eptime \mid \state_\eptime)
\end{equation}
and the second (for $\eptime < \horizon$) is
\begin{equation}
    \label{eq:fail_prob_recursion_2}
    \PP(\failevent \mid \mathcal{O}_\eptime, \state_\eptime, \ac_\eptime)
    = \begin{cases}
        1 & \quad \ac_\eptime \notin \acspace^*_\eptime(\state_\eptime) \\
        \PP(\failevent \mid \mathcal{O}_{\eptime + 1}, \state_{\eptime + 1}) & \quad \ac_\eptime \in \acspace^*_\eptime(\state_\eptime).
    \end{cases}
\end{equation}
We apply these results recursively from $\eptime = \horizon, \hdots, 1$ to finally obtain $\PP(\failevent \mid \mathcal{O}_1, \state_1) = \PP(\failevent)$.

The remaining difficulty is calculating $\PP(\ac_\eptime \mid \state_\eptime)$. Recall from Algorithm \ref{alg:qirl} that $\ac_\eptime$ is chosen as the first action of a $\numqvi$-action sequence
\begin{equation*}
    \ac_{\eptime:\eptime + \numqvi - 1} \in \arg \max_{\ac_{\eptime:\eptime + \numqvi - 1} \in \acspace^\numqvi} \hat{Q}_\eptime(\state_\eptime, \ac_{\eptime:\eptime + \numqvi - 1}),
\end{equation*}
where $\hat{Q}_\eptime(\state_\eptime, \ac_{\eptime:\eptime + \numqvi - 1})$ is the empirical mean return-to-go from $\numep{}$ episodes starting in state $\state_\eptime$ and taking actions $\ac_\eptime, \hdots, \ac_{\eptime + \numqvi - 1}$ followed by actions sampled from the exploration policy. To simplify notation, let $\acseq{\numqvi}_\eptime$ denote $\ac_{\eptime:\eptime + \numqvi - 1}$. We use various inequalities, described in detail below, to bound the probability that a particular $\numqvi$-action sequence is chosen:
\begin{align*}
    \underline{p}(\acseq{\numqvi}_\eptime)
    & \leq
    \PP_{\numep{}}\left( \hat{Q}_\eptime(\state_\eptime, \acseq{\numqvi}_\eptime) > \max_{\acseq{\numqvi}_\eptime' \in \acspace^\numqvi \setminus \{ \acseq{\numqvi}_\eptime \}} \hat{Q} ( \state_\eptime, \acseq{\numqvi}_\eptime' ) \right) \\
    & \leq \PP_{\numep{}}\left( \hat{Q}_\eptime(\state_\eptime, \acseq{\numqvi}_\eptime) \geq \max_{\acseq{\numqvi}_\eptime' \in \acspace^\numqvi \setminus \{ \acseq{\numqvi}_\eptime \}} \hat{Q} ( \state_\eptime, \acseq{\numqvi}_\eptime' ) \right)
    \leq \overline{p}(\acseq{\numqvi}_\eptime).
\end{align*}
Letting $p(\acseq{\numqvi}_\eptime)$ denote the actual probability an action sequence is chosen, we can rewrite (\ref{eq:fail_prob_recursion}) to
\begin{align}
    \label{eq:fail_prob_lp_objective}
    \PP(\failevent \mid \mathcal{O}_\eptime, \state_\eptime)
    = \sum_{\ac_\eptime \in \acspace} \PP(\failevent \mid \mathcal{O}_\eptime, \state_\eptime, \ac_\eptime) \sum_{\ac_{\eptime + 1:\eptime + \numqvi - 1} \in \acspace^{\numqvi - 1}} p(\acseq{\numqvi}_\eptime).
\end{align}
Given the bounds on $p(\acseq{\numqvi}_\eptime)$ (i.e., $\underline{p}$ and $\overline{p}$), we formulate a linear program with the bounds as constraints, plus the constraint that $\sum_{\acseq{\numqvi}_\eptime \in \acspace^\numqvi} p(\acseq{\numqvi}_\eptime) = 1$, with the objective of maximizing (\ref{eq:fail_prob_lp_objective}). Solving this gives an upper bound on $\PP(\failevent \mid \mathcal{O}_\eptime, \state_\eptime)$. We can then propagate this bound recursively using (\ref{eq:fail_prob_recursion_2}) to bound $\PP(\failevent)$.

\subsection*{Calculating $\underline{p}$ and $\overline{p}$}
We use up to four methods to bound $p(\acseq{\numqvi}_\eptime)$, and pick the one which gives the lowest conditional failure probability after solving the linear program described above. As previously, let $\qdist_\eptime(\state_\eptime, \acseq{\numqvi}_\eptime)$ denote the distribution of the reward-to-go starting in $\state_\eptime$ and taking actions $\ac_\eptime, \hdots, \ac_{\eptime + \numqvi - 1}$ followed by actions sampled from the exploration policy. Thus we can write
\begin{equation*}
    \hat{Q}_\eptime(\state_\eptime, \acseq{\numqvi}_\eptime)
    = \frac{1}{\numep{}} \sum_{i = 1}^{\numep{}} X_i \qquad \text{where} \quad X_1, \hdots, X_{\numep{}} \simiid \qdist_\eptime(\state_\eptime, \acseq{\numqvi}_\eptime).
\end{equation*}
Most of the following methods use the following decomposition:
\begin{align}
    & \PP_{\numep{}}\left( \hat{Q}_\eptime(\state_\eptime, \acseq{\numqvi}_\eptime) > \max_{\acseq{\numqvi}_\eptime' \in \acspace^\numqvi \setminus \{ \acseq{\numqvi}_\eptime \}} \hat{Q} ( \state_\eptime, \acseq{\numqvi}_\eptime' ) \right) \nonumber \\
    & \overset{\text{(i)}}{=} \int \prod_{\acseq{\numqvi}_\eptime' \in \acspace^\numqvi \setminus \{ \acseq{\numqvi}_\eptime \}} \PP_{\numep{}}\left( \hat{Q} ( \state_\eptime, \acseq{\numqvi}_\eptime' ) < \hat{Q}_\eptime(\state_\eptime, \acseq{\numqvi}_\eptime) \right) \; d \PP_{\numep{}} \left( \hat{Q}_\eptime(\state_\eptime, \acseq{\numqvi}_\eptime) \right) \nonumber \\
    & \overset{\text{(ii)}}{\geq} \sum_{i = 1}^N \PP_{\numep{}} \left( q_{i - 1} < \hat{Q}_\eptime(\state_\eptime, \acseq{\numqvi}_\eptime) \leq q_i \right) \prod_{\acseq{\numqvi}_\eptime' \in \acspace^\numqvi \setminus \{ \acseq{\numqvi}_\eptime \}} \PP_{\numep{}}\left( \hat{Q} ( \state_\eptime, \acseq{\numqvi}_\eptime' ) \leq q_{i - 1} \right) \label{eq:riemann_sum_bound}
\end{align}
for some sequence $-\infty = q_0 \leq q_1 \leq \hdots \leq q_N = \infty$. Here, (i) uses the fact that the random variables $\hat{Q}_\eptime(\state_\eptime, \acseq{\numqvi}_\eptime)$ across all action sequences $\acseq{\numqvi}_\eptime \in \acspace^\numqvi$ are independent. (ii) is a lower bound on the integral via a Riemann sum.

Alternatively, suppose we know that the CDF of $\hat{Q}_\eptime(\state_\eptime, \acseq{\numqvi}_\eptime)$ is bounded by 
\begin{equation*}
    \PP_{\numep{}} \left( \hat{Q}_\eptime(\state_\eptime, \acseq{\numqvi}_\eptime) \leq x \right)
    \geq F_{\underline{Z}}(x)
\end{equation*}
where $F_{\underline{Z}}(x)$ is the continuous CDF of some random variable $\underline{Z}$. Then
\begin{align}
    & \PP_{\numep{}}\left( \hat{Q}_\eptime(\state_\eptime, \acseq{\numqvi}_\eptime) > \max_{\acseq{\numqvi}_\eptime' \in \acspace^\numqvi \setminus \{ \acseq{\numqvi}_\eptime \}} \hat{Q} ( \state_\eptime, \acseq{\numqvi}_\eptime' ) \right) \nonumber \\
    & \geq \PP_{\numep{}}\left( \underline{Z} > \max_{\acseq{\numqvi}_\eptime' \in \acspace^\numqvi \setminus \{ \acseq{\numqvi}_\eptime \}} \hat{Q} ( \state_\eptime, \acseq{\numqvi}_\eptime' ) \right) \nonumber \\
    & = \int \prod_{\acseq{\numqvi}_\eptime' \in \acspace^\numqvi \setminus \{ \acseq{\numqvi}_\eptime \}} \PP_{\numep{}}\left( \hat{Q} ( \state_\eptime, \acseq{\numqvi}_\eptime' ) < \hat{Q}_\eptime(\state_\eptime, \acseq{\numqvi}_\eptime) \right) \; d \PP_{\numep{}} \left( \underline{Z} \right) \nonumber \\
    & \geq \frac{1}{N} \sum_{i = 1}^N \prod_{\acseq{\numqvi}_\eptime' \in \acspace^\numqvi \setminus \{ \acseq{\numqvi}_\eptime \}} \PP_{\numep{}}\left( \hat{Q} ( \state_\eptime, \acseq{\numqvi}_\eptime' ) \leq F^{-1}_{\underline{Z}} \left(\frac{i - 1}{N}\right) \right) \label{eq:inverse_cdf_bound}
\end{align}
by a similar argument. We also have equivalent bounds in the other direction:
\begin{align}
    \PP_{\numep{}}\left( \hat{Q}_\eptime(\state_\eptime, \acseq{\numqvi}_\eptime) \geq \max_{\acseq{\numqvi}_\eptime' \in \acspace^\numqvi \setminus \{ \acseq{\numqvi}_\eptime \}} \hat{Q} ( \state_\eptime, \acseq{\numqvi}_\eptime' ) \right)
    & \leq \sum_{i = 1}^N \PP_{\numep{}} \left( q_{i - 1} < \hat{Q}_\eptime(\state_\eptime, \acseq{\numqvi}_\eptime) \leq q_i \right) \prod_{\acseq{\numqvi}_\eptime' \in \acspace^\numqvi \setminus \{ \acseq{\numqvi}_\eptime \}} \PP_{\numep{}}\left( \hat{Q} ( \state_\eptime, \acseq{\numqvi}_\eptime' ) \leq q_i \right) \label{eq:riemann_sum_bound_2} \\
    \PP_{\numep{}}\left( \hat{Q}_\eptime(\state_\eptime, \acseq{\numqvi}_\eptime) \geq \max_{\acseq{\numqvi}_\eptime' \in \acspace^\numqvi \setminus \{ \acseq{\numqvi}_\eptime \}} \hat{Q} ( \state_\eptime, \acseq{\numqvi}_\eptime' ) \right)
    & \leq \frac{1}{N} \sum_{i = 1}^N \prod_{\acseq{\numqvi}_\eptime' \in \acspace^\numqvi \setminus \{ \acseq{\numqvi}_\eptime \}} \PP_{\numep{}}\left( \hat{Q} ( \state_\eptime, \acseq{\numqvi}_\eptime' ) \leq F^{-1}_{\overline{Z}} \left(\frac{i}{N}\right) \right) \label{eq:inverse_cdf_bound_2}
\end{align}
where the CDF of $\overline{Z}$ is greater than or equal to that of $\hat{Q}_\eptime(\state_\eptime, \acseq{\numqvi}_\eptime)$. We use $N = 100$ when using these bounds.

\smallparagraph{Binomial bounds} In the case where for all $\acseq{\numqvi}_\eptime \in \acspace^\numqvi$, the distribution $\qdist_\eptime(\state_\eptime, \acseq{\numqvi}_\eptime)$ has mass on only $0$ and some other value $C$, we have
\begin{equation*}
    \hat{Q}_\eptime(\state_\eptime, \acseq{\numqvi}_\eptime) \sim \frac{C}{\numep{}} \; \text{Binom}\left(\numep{}, \frac{1}{C} Q_\eptime(\state_\eptime, \acseq{\numqvi}_\eptime)\right).
\end{equation*}
This case occurs in many environments that are goal-based, i.e. where the agent gets reward only for reaching some goal and then the episode ends. We find that it significantly improves the sample complexity bounds in those environments. Without loss of generality, we may assume $C = 1$. We then apply (\ref{eq:riemann_sum_bound}) and (\ref{eq:riemann_sum_bound_2}) to obtain $\underline{p}(\acseq{\numqvi}_\eptime)$ and $\overline{p}(\acseq{\numqvi}_\eptime)$. We let $q_0, \hdots, q_N$ be set such that
\begin{equation*}
    \PP_{\numep{}} \left( q_{i - 1} < \hat{Q}_\eptime(\state_\eptime, \acseq{\numqvi}_\eptime) \leq q_i \right) \approx \frac{1}{N}
\end{equation*}
using either the exact inverse CDF of the binomial distribution for small $\numep{}$ or a normal approximation for large $\numep{}$. Then, we can calculate all terms in the bounds (\ref{eq:riemann_sum_bound}) and (\ref{eq:riemann_sum_bound_2}) using the binomial CDF. Since the CDF is more expensive to calculate for larger $\numep{}$, we only use the binomial-based bounds when $\numep{} \leq 10^6$ and $\numqvi = 1$.

\smallparagraph{Berry-Esseen bounds} For this type of bound, we calculate the variance $\sigma^2 = \Var( \qdist_\eptime(\state_\eptime, \acseq{\numqvi}_\eptime) )$ and third absolute moment
\begin{equation*}
    \rho = \EE_{X \sim \qdist_\eptime(\state_\eptime, \acseq{\numqvi}_\eptime)} [ |X|^3 ]
\end{equation*}
for each $\acseq{\numqvi}_\eptime \in \acspace^\numqvi$. Then by the Berry-Esseen theorem \citep{shevtsova_absolute_2011}, we have that
\begin{equation*}
    \left| \PP_{\numep{}}( \hat{Q}_\eptime(\state_\eptime, \acseq{\numqvi}_\eptime) \leq u )
    - \PP_{X \sim \mathcal{N}(Q_\eptime(\state_\eptime, \acseq{\numqvi}_\eptime), \sigma^2 / \numep{}) } (X \leq u) \right|
    \leq \frac{ \min \{ 0.3328 (\rho / \sigma^3 + 0.429), 0.33554 (\rho / \sigma^3 + 0.415) \} }{\sqrt{\numep{}} }.
\end{equation*}
The resulting upper and lower bounds on the CDFs of $\hat{Q}_\eptime(\state_\eptime, \acseq{\numqvi}_\eptime)$ for all $\acseq{\numqvi}_\eptime \in \acspace^\numqvi$ can be used in (\ref{eq:inverse_cdf_bound}) and (\ref{eq:inverse_cdf_bound_2}) to calculate $\underline{p}(\acseq{\numqvi}_\eptime)$ and $\overline{p}(\acseq{\numqvi}_\eptime)$. Since this bound requires order $N$ evaluations of the normal CDF and inverse CDF, which is somewhat expensive, we only use it when $\acsize^\numqvi \leq 100$.

\smallparagraph{Bernstein bounds} Similarly to the Berry-Esseen bounds, Bernstein's inequality can be used to bound the CDF of $\hat{Q}_\eptime(\state_\eptime, \acseq{\numqvi}_\eptime)$, and is sometimes superior to Berry-Esseen for large $\numep{}$ due to giving tail bounds that decay exponentially rather than quadratically. In particular, suppose $\qdist_\eptime(\state_\eptime, \acseq{\numqvi}_\eptime)$ is supported on the interval $[\alpha, \beta]$; we can compute these bounds via value iteration. Then
\begin{equation*}
    \Var(\qdist_\eptime(\state_\eptime, \acseq{\numqvi}_\eptime)) \leq V = \left(\beta - Q_\eptime(\state_\eptime, \acseq{\numqvi}_\eptime)\right) \left( Q_\eptime(\state_\eptime, \acseq{\numqvi}_\eptime) - \alpha\right).
\end{equation*}
Bernstein's inequality gives the following bounds on the CDF of $\hat{Q}_\eptime(\state_\eptime, \acseq{\numqvi}_\eptime)$:
\begin{align*}
    \PP_{\numep{}}\left(\hat{Q}_\eptime(\state_\eptime, \acseq{\numqvi}_\eptime) \leq Q_\eptime(\state_\eptime, \acseq{\numqvi}_\eptime) + u\right)
    & \geq \begin{cases}
        1 & \quad u \leq 0 \\
        1 - \exp \left\{ -\frac{\numep{} u^2 / 2}{V + (\beta - \alpha) u / 3} \right\} & \quad \text{otherwise}
    \end{cases} \\
    \PP_{\numep{}}\left(\hat{Q}_\eptime(\state_\eptime, \acseq{\numqvi}_\eptime) \leq Q_\eptime(\state_\eptime, \acseq{\numqvi}_\eptime) + u\right)
    & \leq \begin{cases}
        1 & \quad u \geq 0 \\
        \exp \left\{ -\frac{\numep{} u^2 / 2}{V + (\beta - \alpha) u / 3} \right\} & \quad \text{otherwise}.
    \end{cases}
\end{align*}
Similarly to the Berry-Esseen bounds, we use these in (\ref{eq:inverse_cdf_bound}) and (\ref{eq:inverse_cdf_bound_2}) to calculate $\underline{p}(\acseq{\numqvi}_\eptime)$ and $\overline{p}(\acseq{\numqvi}_\eptime)$ when $\acsize^\numqvi \leq 100$.

\smallparagraph{Bennett bounds} The final method we use to calculate $\underline{p}(\acseq{\numqvi}_\eptime)$ and $\overline{p}(\acseq{\numqvi}_\eptime)$ is computationally cheaper than the others, so we can use it no matter the size of $\acsize^\numqvi$. As in the Bernstein bounds, we calculate the interval support and bound the variance of each $\qdist_\eptime(\state_\eptime, \acseq{\numqvi}_\eptime)$. We then let $u$ be the arithematic mean of the highest action sequence Q value and the second-highest, i.e.
\begin{align*}
    u = \frac{1}{2} \left( \max_{\acseq{\numqvi}_\eptime \in \acspace^\numqvi} Q_\eptime(\state_\eptime, \acseq{\numqvi}_\eptime)
    + \max_{\acseq{\numqvi}_\eptime' \notin \arg\max_{\acseq{\numqvi}_\eptime \in \acspace^\numqvi} Q_\eptime(\state_\eptime, \acseq{\numqvi}_\eptime)} Q_\eptime(\state_\eptime, \acseq{\numqvi}_\eptime') \right).
\end{align*}
Then, we for each action sequence with less-than-highest Q-values, i.e. for each $\acseq{\numqvi}_\eptime \notin \arg\max_{\acseq{\numqvi}_\eptime \in \acspace^\numqvi} Q_\eptime(\state_\eptime, \acseq{\numqvi}_\eptime)$, we calculate the upper bound
\begin{align}
    & \PP_{\numep{}}\left( \hat{Q}_\eptime(\state_\eptime, \acseq{\numqvi}_\eptime) \geq \max_{\acseq{\numqvi}_\eptime' \in \acspace^\numqvi \setminus \{ \acseq{\numqvi}_\eptime \}} \hat{Q} ( \state_\eptime, \acseq{\numqvi}_\eptime' ) \right) \nonumber \\
    & \leq \PP_{\numep{}}\left( \hat{Q}_\eptime(\state_\eptime, \acseq{\numqvi}_\eptime) \geq \max_{\acseq{\numqvi}_\eptime' \in \arg\max_{\acseq{\numqvi}_\eptime \in \acspace^\numqvi} Q_\eptime(\state_\eptime, \acseq{\numqvi}_\eptime)} \hat{Q} ( \state_\eptime, \acseq{\numqvi}_\eptime' ) \right) \nonumber \\
    & = 1 - \PP_{\numep{}}\left( \exists \acseq{\numqvi}_\eptime' \in \arg\max_{\acseq{\numqvi}_\eptime' \in \acspace^\numqvi} Q_\eptime(\state_\eptime, \acseq{\numqvi}_\eptime') \qquad \hat{Q}_\eptime(\state_\eptime, \acseq{\numqvi}_\eptime) < \hat{Q} ( \state_\eptime, \acseq{\numqvi}_\eptime' ) \right) \nonumber \\
    & \leq 1 - \PP_{\numep{}}\left(  \hat{Q}_\eptime(\state_\eptime, \acseq{\numqvi}_\eptime) < u \qquad \wedge \qquad \forall \acseq{\numqvi}_\eptime' \in \arg\max_{\acseq{\numqvi}_\eptime' \in \acspace^\numqvi} Q_\eptime(\state_\eptime, \acseq{\numqvi}_\eptime') \quad \hat{Q} ( \state_\eptime, \acseq{\numqvi}_\eptime' ) > u \right) \nonumber \\
    & = 1 - \PP_{\numep{}}\left(  \hat{Q}_\eptime(\state_\eptime, \acseq{\numqvi}_\eptime) < u\right) \prod_{\acseq{\numqvi}_\eptime' \in \arg\max_{\acseq{\numqvi}_\eptime' \in \acspace^\numqvi} \hat{Q} ( \state_\eptime, \acseq{\numqvi}_\eptime' )} \PP_{\numep{}}\left(\hat{Q} ( \state_\eptime, \acseq{\numqvi}_\eptime' ) > u\right) \nonumber \\
    & = 1 - \left( 1 - \PP_{\numep{}}\left(  \hat{Q}_\eptime(\state_\eptime, \acseq{\numqvi}_\eptime) \geq u\right) \right) \prod_{\acseq{\numqvi}_\eptime' \in \arg\max_{\acseq{\numqvi}_\eptime' \in \acspace^\numqvi} \hat{Q} ( \state_\eptime, \acseq{\numqvi}_\eptime' )} \left(1 - \PP_{\numep{}}\left(\hat{Q} ( \state_\eptime, \acseq{\numqvi}_\eptime' ) \leq u\right)\right). \label{eq:bennett_bound}
\end{align}
Each of the tail bounds in (\ref{eq:bennett_bound}) can be upper bounded using Bennett's inequality to obtain $\overline{p}(\acseq{\numqvi}_\eptime)$. We let $\overline{p}(\acseq{\numqvi}_\eptime) = 1$ if $\acseq{\numqvi}_\eptime \in \arg\max_{\acseq{\numqvi}_\eptime \in \acspace^\numqvi} Q_\eptime(\state_\eptime, \acseq{\numqvi}_\eptime)$ and we set  $\underline{p}(\acseq{\numqvi}_\eptime) = 0$ for all $\acseq{\numqvi}_\eptime \in \acsize^\numqvi$.

\section{Previously proposed sample complexity bounds}
\label{sec:more_related_work}

In this appendix, we give proofs of sample complexity bounds based on properties previously proposed in the RL theory literature. We also compare these bounds to our effective horizon-based bounds in examples that showcase their failure modes.

\subsection{Upper confidence bounds (UCB) and strategic exploration}
\label{sec:ucb}

A central problem in RL is exploration: how to efficiently reach enough states in an MDP in order to identify the optimal policy. One common way of approaching exploration is with upper-confidence bounds (UCB), which originated in the bandit literature. Algorithms using UCBs generally choose actions based on the current best estimate of that action's value plus an exploration ``bonus'' that incentivizes exploration of little-seen states. Examples in the RL literature include \citet{kakade_sample_2003,azar_minimax_2017,jiang_contextual_2017,jin_is_2018,jin_provably_2019,du_bilinear_2021,jin_bellman_2021}. Generally, these UCB algorithms achieve minimax sample complexity in terms of some measure of the ``size'' of the state space---either the number of states $\statespace$ \citep{azar_minimax_2017}, or quantities like the Bellman-Eluder dimension \citep{jin_bellman_2021}.

A very simple UCB-type algorithm, \textsc{R-max} \citep{brafman_r-max_2002,kakade_sample_2003}, achieves a sample complexity bounded by $\statesize \acsize \horizon$ in deterministic, tabular MDPs. It depends on knowing the maximum reward at any state-action pair in the MDP, $\reward_\text{max} = \max_{(\state, \ac) \in \statespace \times \acspace} \reward(\state, \ac)$. It also requires access to a computational oracle that can calculate an optimal policy for any transition function $\dynamics$ and reward function $\reward$, for instance via value iteration.
\begin{algorithmic}[1]
    \Procedure{R-max}{}
        \State initialize $\hat{\dynamics}(\state, \ac) \gets \state$ for all $(\state, \ac) \in \statespace \times \acspace$
        \State initialize $\hat{\reward}(\state, \ac) \gets \reward_\text{max}$ for all $(\state, \ac) \in \statespace \times \acspace$
        \For{$j = 1, \hdots, \statesize \acsize$}
            \For{$\eptime = 1, \hdots, \horizon$}
                \State take an action $\ac$ in the current state $\state$ according to the optimal policy for $\hat{\dynamics}$ and $\hat{\reward}$
                \State $\hat{\reward}(\state, \ac) \gets$ the observed reward
                \State $\hat{\dynamics}(\state, \ac) \gets$ the observed next state $\state'$
            \EndFor
        \EndFor
        \State \Return an optimal policy for $\hat{\dynamics}$ and $\hat{\reward}$
    \EndProcedure
\end{algorithmic}
This is the version of \textsc{R-max} for deterministic MDPs; there is a more complex version for stochastic MDPs. The exploration bonuses in \textsc{R-max} are simply the initialization of $\hat{\reward}$ to the maximum possible reward. This ensures that an optimal value function computed from $\hat{\dynamics}$ and $\hat{\reward}$ is always an upper bound on the true optimal value function. The following result shows that \textsc{R-max} finds the optimal policy, and thus proves that its sample complexity is at most $\statesize \acsize \horizon$.
\begin{theorem}
    \label{thm:ucb}
    \textsc{R-max} returns an optimal policy.
\end{theorem}
\begin{proof}
Let $\hat{V}^*$ and $\hat{Q}^*$ be the optimal value function and Q-function under $\hat{\dynamics}$ and $\hat{\reward}$. We will begin by showing that at any point in the algorithm, $V^*_\eptime(\state) \leq \hat{V}^*_\eptime(\state)$ and $Q^*_\eptime(\state, \ac) \leq \hat{Q}^*_\eptime(\state, \ac)$ for all $(\eptime, \state, \ac) \in [\horizon] \times \statespace \times \acspace$.

The proof is via induction on $\eptime$ from $\horizon$ to $1$. To begin, clearly $\hat{Q}^*_\horizon(\state, \ac) \in \{Q^*_\horizon(\state, \ac), \reward_\text{max}\}$, so the bound holds for $\hat{Q}^*_\horizon$. Now, suppose that at some $\eptime \in [\horizon]$ and all $(\state, \ac) \in \statespace \times \acspace$, $Q^*_\eptime(\state, \ac) \leq \hat{Q}^*_\eptime(\state, \ac)$. Then
\begin{equation*}
    \hat{V}^*_\eptime(\state)
    = \max_{\ac \in \acspace} \hat{Q}^*_\eptime(\state, \ac)
    \geq \max_{\ac \in \acspace} Q^*_\eptime(\state, \ac)
    =  V^*_\eptime(\state),
\end{equation*}
so the bound holds for $\hat{V}^*_\eptime$ as well. Finally, say that at some $\eptime \in [\horizon - 1]$ and for all $\state \in \statespace$, $V^*_{\eptime + 1}(\state) \leq \hat{V}^*_{\eptime + 1}(\state)$. To show this implies $Q^*_\eptime(\state, \ac) \leq \hat{Q}^*_\eptime(\state, \ac)$ for any $(\state, \ac) \in \statespace \times \acspace$, consider two cases. First, if $(\state, \ac)$ has been seen by the algorithm, then
\begin{equation*}
    \hat{Q}^*_\eptime(\state, \ac)
    = \hat{\reward}(\state, \ac) + \hat{V}^*_{\eptime + 1}(\hat{\dynamics}(\state, \ac)
    = \reward(\state, \ac) + \hat{V}^*_{\eptime + 1}(\dynamics(\state, \ac)
    \geq \reward(\state, \ac) + V^*_{\eptime + 1}(\dynamics(\state, \ac)
    = Q^*_\eptime(\state, \ac).
\end{equation*}
Otherwise, if $(\state, \ac)$ has not been seen, then $\hat{\reward}(\state, \ac) = \reward_\text{max}$ and $\hat{\dynamics}(\state, \ac) = \state$. In this case,
\begin{equation*}
    \hat{Q}^*_\eptime(\state, \ac) = (\horizon - \eptime + 1) \reward_\text{max} \geq Q^*_\eptime(\state, \ac).
\end{equation*}
By induction we see that $V^*_\eptime(\state) \leq \hat{V}^*_\eptime(\state)$ and $Q^*_\eptime(\state, \ac) \leq \hat{Q}^*_\eptime(\state, \ac)$ for all $(\eptime, \state, \ac) \in [\horizon] \times \statespace \times \acspace$.

Next, we will prove that any optimal policy $\policy$ for $\hat{\dynamics}$ and $\hat{\reward}$ must either (a) be optimal for $\dynamics$ and $\reward$ or (b) reach a previously unseen state-action pair. In particular, we will show that if $V^\policy_1(\state_1) < \hat{V}^*_1(\state_1)$, then $\policy$ must reach a previously unseen state-action pair. Otherwise, $V^\policy_1(\state_1) \geq \hat{V}^*_1(\state_1) \geq V^*_1(\state_1)$, showing that $\policy$ is optimal for $\dynamics$ and $\reward$.

We will again work inductively starting from the last timestep. First, suppose that $Q^\policy_\horizon(\state, \ac) < \hat{Q}^*_\horizon(\state, \ac)$ for some $(\state, \ac) \in \statespace \times \acspace$. This is equivalent to $\reward(\state, \ac) < \hat{\reward}(\state, \ac)$, which clearly means that $(\state, \ac)$ cannot have been explored.

Now, suppose that at some timestep $\eptime \in [\horizon]$, we know that for any $(\state, \ac) \in \statespace \times \acspace$, $Q^\policy_\eptime(\state, \ac) < \hat{Q}^*_\eptime(\state, \ac)$ implies that $\policy$ must explore some new state-action pair at or after timestep $\eptime$ starting in $(\state, \ac)$. If $V^\policy_\eptime(\state) < \hat{V}^*_\eptime(\state)$ for some $\state \in \statespace$, then
\begin{align*}
    Q^\policy_\eptime(\state, \policy_\eptime(\state)) < \max_{\ac \in \acspace} \hat{Q}^*_\eptime(\state, \ac) = \hat{Q}^*_\eptime(\state, \policy_\eptime(\state)).
\end{align*}
By assumption this means $\policy$ must explore some new state-action pair at or after timestep $\eptime$, since it takes an action $\ac$ satisfying $Q^\policy_\eptime(\state, \ac) < \hat{Q}^*_\eptime(\state, \ac)$.

Finally, suppose that for some $\eptime \in [\horizon]$, we know that for any $\state \in \statespace$, $V^\policy_{\eptime + 1}(\state) < \hat{V}^*_{\eptime 
 + 1}(\state)$ implies that $\policy$ must explore some new state-action pair at or after timestep $\eptime + 1$ starting in $\state$. Suppose for some $(\state, \ac) \in \statespace \times \acspace$ that $Q^\policy_\eptime(\state, \ac) < \hat{Q}^*_\eptime(\state, \ac)$. Then
\begin{align*}
    \reward(\state, \ac) + V^\policy_{\eptime + 1}(\dynamics(\state, \ac)) < \hat{\reward}(\state, \ac) + V^\policy_{\eptime + 1}(\hat{\dynamics}(\state, \ac))
\end{align*}
which implies that either $\reward(\state, \ac) < \hat{\reward}(\state, \ac)$ or $V^\policy_{\eptime + 1}(\dynamics(\state, \ac)) < V^\policy_{\eptime + 1}(\hat{\dynamics}(\state, \ac))$. In the first case, $(\state, \ac)$ must be unexplored. In the second case, either $(\state, \ac)$ is unexplored, or
\begin{equation*}
    V^\policy_{\eptime + 1}(\dynamics(\state, \ac)) < V^\policy_{\eptime + 1}(\dynamics(\state, \ac)).
\end{equation*}
In any of these cases, $\policy$ must explore a new state-action pair either in this timestep or in the future starting from $(\state, \ac)$.

Inductively, this shows that $V^\policy_1(\state_1) < \hat{V}^*_1(\state_1)$ implies that $\policy$ must reach a previously unseen state-action pair. We will show that this property implies that \textsc{R-max} must return an optimal policy.

In particular, note that after the $j$th loop iteration in $\textsc{R-max}$, it must either have an optimal policy or have explored at least $j$ of the state-action pairs in the MDP.
This is a simple consequence of the above property: at each iteration, either the policy used by $\textsc{R-max}$ must be optimal or it must explore at least one additional state-action pair.
This means that after all the $\statesize \acsize$ loop iterations, $\textsc{R-max}$ will either have an optimal policy or have explored \emph{all} the state-action pairs, in which case it will also have an optimal policy.
\end{proof}

\subsection{Covering length}
\label{sec:covering_length}

The \emph{covering length} of an MDP was originally proposed by \citet{even-dar_learning_2003} and later used by \citet{liu_when_2019} to prove sample complexity bounds on RL algorithms which use random exploration. \citet{liu_when_2019} show various bounds on the covering length using graph-theoretic notions. While they focus on discounted infinite-horizon MDPs, we use a version of covering length adapted to finite-horizon MDPs, similar to that used by \citet{dann_guarantees_2022}.

\begin{definition}[Covering length]
    The \emph{covering length} $\covlen$ of an MDP under an exploration policy $\expolicy$ is the number of episodes needed until all state-action pairs have been visited with probability at least $1/2$.
\end{definition}

One can easily show sample complexity bounds based on the covering length.

\begin{theorem}[Covering length sample complexity bound]
    There is an RL algorithm which can solve any MDP with sample complexity $\horizon \covlen$ given an exploration policy $\expolicy$, where $\covlen$ is the covering length of $\expolicy$.
\end{theorem}

\begin{proof}
Consider the following RL algorithm:
\begin{algorithmic}[1]
    \Procedure{CoveringLengthRL}{$\expolicy, \covlen$}
        \State collect a dataset of $\covlen$ episodes, sampling actions according to $\expolicy$
        \State record $\hat{\reward}(\state, \ac)$ and $\hat{\dynamics}(\state, \ac)$ for all state-action pairs seen in the dataset
        \State define $\hat{\reward}(\state, \ac)$ and $\hat{\dynamics}(\state, \ac)$ arbitrarily for state-action pairs not seen in the dataset
        \State run value iteration using $\hat{\reward}$ and $\hat{\dynamics}$ to obtain a policy $\policy$
		\State \Return $\policy$
    \EndProcedure
\end{algorithmic}
By the definition of covering length, with probability at least $1/2$ the algorithm should produce $\hat{\reward}(\state, \ac) = \reward(\state, \ac)$ and $\hat{\dynamics}(\state, \ac) = \dynamics(\state, \ac)$ for all $(\state, \ac) \in \statespace \times \acspace$. In this case, $\policy$ will be an optimal policy. Thus, \textsc{CoveringLengthRL} returns an optimal policy with probability at least $1/2$ while interacting with the environment for $\horizon \covlen$ timesteps. This means the sample complexity of \textsc{CoveringLengthRL} is at most $\horizon \covlen$.
\end{proof}
To bound the covering length for MDPs in the \datasetname dataset, we make use of the following result.
\begin{lemma}[Bounds on the covering length]
    \label{lemma:covlen_bounds}
    Define the occupancy measure $\mu$ of $\expolicy$ as
    \begin{equation*}
        \mu_\eptime(\state, \ac) = \PP_{\expolicy}(\state_\eptime = \state \wedge \ac_\eptime = \ac).
    \end{equation*}
    Suppose that for every state-action pair $(\state, \ac)$, there is some timestep $\eptime$ when $\mu_\eptime(\state, \ac) > 0$. Then
    \begin{equation*}
        \frac{\log(2)}{2 \min_{(\state, \ac) \in \statespace \times \acspace} \; \sum_{\eptime \in [\horizon]} \mu_\eptime(\state, \ac)}  \leq \covlen \leq \left\lceil \frac{\log(2 \statesize \acsize)}{\min_{(\state, \ac) \in \statespace \times \acspace} \; \max_{\eptime \in [\horizon]} \mu_\eptime(\state, \ac)} \right\rceil.
    \end{equation*}
\end{lemma}
We calculate $\mu_\text{min} = \min_{(\state, \ac) \in \statespace \times \acspace} \; \max_{\eptime \in [\horizon]} \mu_\eptime(\state, \ac)$ for each MDP in \datasetname and use the upper bound from Lemma \ref{lemma:covlen_bounds} to obtain a sample complexity bound of $\horizon \log(2 \statesize \acsize \horizon) / \mu_\text{min}$. Since $\sum_{\eptime \in [\horizon]} \mu_\eptime(\state, \ac) \leq \horizon \max_{\eptime \in [\horizon]} \mu_\eptime(\state, \ac)$ the upper and lower bounds in Lemma \ref{lemma:covlen_bounds} agree up to a factor of $\horizon \log(2 \statesize \acsize) / \log(2)$, so this is reasonably tight. In fact, in 122 of the 155 MDPs in \datasetname, $\min_{(\state, \ac) \in \statespace \times \acspace} \; \sum_{\eptime \in [\horizon]} \mu_\eptime(\state, \ac) = \min_{(\state, \ac) \in \statespace \times \acspace} \; \max_{\eptime \in [\horizon]} \mu_\eptime(\state, \ac)$, making the upper and lower bounds tight up to only logarithmic factors.
\begin{proof}
    Let $\PP_{\numep{}}$ be a probability measure corresponding to sampling $\numep{}$ episodes following $\expolicy$. Let $\mathcal{E}^j_\eptime(\state, \ac)$ denote the event that the $j$th episode has $(\state_\eptime, \ac_\eptime) = (\state, \ac)$. Let $\mathcal{E}_\eptime(\state, \ac)$ be the event that $\mathcal{E}^j_\eptime(\state, \ac)$ occurs in at least one one of those episodes, i.e.
    \begin{equation*}
        \mathcal{E}_\eptime(\state, \ac) \coloneqq \bigvee_{j = 1}^{\numep{}} \mathcal{E}^j_\eptime(\state, \ac).
    \end{equation*}
    Finally, let $\mathcal{C}$ be the event defined by
    \begin{equation*}
        \mathcal{C} \coloneqq \bigwedge_{(\state, \ac) \in \statespace \times \acspace} \; \bigvee_{\eptime \in [\horizon]} \mathcal{E}_\eptime(\state, \ac).
    \end{equation*}
    That is, $\mathcal{C}$ is when every state-action pair has been seen at some timestep in at least one episode. We can thus equivalently define $\covlen = \min \{ \numep{} \mid \PP_{\numep{}} ( \mathcal{C} ) \geq 1/2 \}$.

    We will now start by showing the upper bound of $\covlen \leq \lceil \log(2 \statesize \acsize) / p \rceil$. Let $\numep{} = \lceil \log(2 \statesize \acsize) / p \rceil$. We can write
    \begin{align*}
        \PP_{\numep{}} ( \neg \mathcal{C} )
        & = \PP_{\numep{}} \left( \exists (\state, \ac) \in \statespace \times \acspace \;:\; \forall \eptime \in [\horizon], j \in [\numep{}] \quad \neg \mathcal{E}^j_\eptime(\state, \ac) \right) \\
        & \leq \sum_{(\state, \ac) \in \statespace \times \acspace} \PP_{\numep{}} \left( \forall \eptime \in [\horizon], j \in [\numep{}] \quad \neg \mathcal{E}^j_\eptime(\state, \ac) \right) \\
        & \leq \sum_{(\state, \ac) \in \statespace \times \acspace} \min_{\eptime \in [\horizon]} \; \PP_{\numep{}} \left( \forall j \in [\numep{}] \quad \neg \mathcal{E}^j_\eptime(\state, \ac) \right) \\
        & = \sum_{(\state, \ac) \in \statespace \times \acspace}  \Big( 1 - \max_{\eptime \in [\horizon]} \; \PP_{\numep{}} (  \mathcal{E}^1_\eptime(\state, \ac) ) \Big)^{\numep{}} \\
        & \leq \sum_{(\state, \ac) \in \statespace \times \acspace}  \Big( 1 - \min_{(\state', \ac') \in \statespace \times \acspace} \; \max_{\eptime \in [\horizon]} \; \PP_{\numep{}} (  \mathcal{E}^1_\eptime(\state', \ac') ) \Big)^{\numep{}} \\
        & \leq \sum_{(\state, \ac) \in \statespace \times \acspace}  \Big( 1 - \min_{(\state', \ac') \in \statespace \times \acspace} \; \max_{\eptime \in [\horizon]} \; \mu_\eptime(\state', \ac') \Big)^{\numep{}} \\
        & \leq \sum_{(\state, \ac) \in \statespace \times \acspace} \exp \left(- \numep{} \min_{(\state', \ac') \in \statespace \times \acspace} \; \max_{\eptime \in [\horizon]} \; \mu_\eptime(\state', \ac') \right) \\
        & \leq \sum_{(\state, \ac) \in \statespace \times \acspace} \frac{1}{2 \statesize \acsize} \\
        & = \frac{1}{2},
    \end{align*}
    which proves the upper bound.

    To show the lower bound, take any $(\state, \ac) \in \arg \min_{(\state, \ac) \in \statespace \times \acspace} \; \sum_{\eptime \in [\horizon]} \mu_\eptime(\state, \ac)$. First, suppose that $\sum_{\eptime \in [\horizon]} \mu_\eptime(\state, \ac) \geq \log(2)$. Then the lower bound in the lemma must be at most $1/2$, which is clearly true. Thus, we may subsequently assume $\sum_{\eptime \in [\horizon]} \mu_\eptime(\state, \ac) < \log(2)$.
    
    Suppose $\numep{} < C / \sum_{\eptime \in [\horizon]}. \mu_\eptime(\state, \ac)$ Then
    \begin{align*}
        \PP_{\numep{}} ( \neg \mathcal{C} )
        & = \PP_{\numep{}} \left( \exists (\state', \ac') \in \statespace \times \acspace \;:\; \forall \eptime \in [\horizon], j \in [\numep{}] \quad \neg \mathcal{E}^j_\eptime(\state', \ac') \right) \\
        & \geq \PP_{\numep{}} \left( \forall \eptime \in [\horizon], j \in [\numep{}] \quad \neg \mathcal{E}^j_\eptime(\state, \ac) \right) \\
        & = \Big( 1 - \PP_{\numep{}} \left( \exists \eptime \in [\horizon] \quad \mathcal{E}^1_\eptime(\state, \ac) \right) \Big)^{\numep{}} \\
        & \geq \left( 1 - \sum_{\eptime \in [\horizon]} \PP_{\numep{}} \left(\mathcal{E}^1_\eptime(\state, \ac) \right) \right)^{\numep{}} \\
        & = \left( 1 - \sum_{\eptime \in [\horizon]} \mu_\eptime(\state, \ac) \right)^{\numep{}} \\
        & \overset{\text{(i)}}{\geq} \exp \left( - 2 \numep{} \sum_{\eptime \in [\horizon]} \mu_\eptime(\state, \ac) \right) \\
        & > \frac{1}{2},
    \end{align*}
    where (i) uses the fact that $1 - x \geq \exp (-2x)$ for $x \in [0, \log(2)]$. This shows that with probability greater than $1/2$, not all state-action pairs will be seen in $\numep{}$ episodes, and thus establishes that $\covlen > \numep{}$, which is the desired bound.
\end{proof}

\subsection{Effective planning window (EPW)}
\label{sec:epw}

Perhaps the closest existing concepts to our effective horizon are various notions of ``effective planning window.'' This generally refers to tree-based planning algorithms which only consider action sequences of some length $\epw$ from the current state, rather than considering action sequences all the way until the end of the MDP. For instance, \citet{kearns_sparse_2002} show that in discounted MDPs, one need only plan to some $\epsilon$-horizon to obtain an $\epsilon$-optimal policy. \citet{jiang_dependence_2015} build on this and show that one may want to use a different discount factor for planning than the one that is used for evaluation. \citet{malik_sample_2021} also introduce a notion of effective planning window based on the number of timesteps one must look ahead in an MDP to avoid terminal states.

We do not directly apply any of these previous results to our setting. Since we are concerned primarily with finite-horizon undiscounted MDPs, it does not make much sense to apply a discount factor as in \citet{kearns_sparse_2002} and \citet{jiang_dependence_2015}. We find that the assumptions in \citet{malik_sample_2021} are quite unusual and do not really hold in any of the environments in \datasetname. In particular, the analysis in \citet{malik_sample_2021} requires that a trajectory through an MDP is either optimal or ends early in a terminal state.

Instead of directly using these results, we define a notion of effective planning window based on the length of action sequences one must consider in an MDP while ignoring any rewards after the sequence.
\begin{definition}[Effective planning window]
    \label{defn:epw}
    Define $Q^1_\eptime(\state, \ac) = \reward(\state, \ac)$ for all $(\eptime, \state, \ac) \in [\horizon] \times \statespace \times \acspace$ and let $Q^i = \qvi(Q^{i - 1})$ for $i = 2, \hdots, \horizon$.
    The effective planning window of an MDP is the minimum $\epw \in [\horizon]$ such that all policies in $\policies(Q^\epw)$ are optimal.
\end{definition}
Note that the effective planning window bears significant similarity to the $k$-QVI-solvability property from Definition \ref{defn:numqvi}. However, $Q^1$ is defined as equal to the reward function for the EPW, while in Definition \ref{defn:numqvi} it is equal to $Q^{\expolicy}$.

The EPW also results in sample complexity bounds of $\horizon^2 \acsize^\epw$ very similar to those of $\horizon^2 \acsize^{\efhorizon{}}$ for the effective horizon. However, we find empirically that $\efhorizon{} < \epw$ in 82\% of the MDPs in \datasetname, making the effective horizon-based bounds generally tighter.

\begin{theorem}
    \label{thm:epw_sampcomplexity}
    For any MDP with effective planning window $\epw$, there is an RL algorithm whose sample complexity is at most $\horizon^2 \acsize^\epw$.
\end{theorem}
\begin{proof}
We will use the following algorithm:
\begin{algorithmic}[1]
    \Procedure{PlanOverWindow}{$\epw$}
        \For{$i = 1, \hdots, \horizon$}
            \For{$\ac_{i : i + \epw - 1} \in \acspace^\numqvi$}
                \State{
                    Sample an episode following $\policy_1, \hdots, \policy_{i - 1}$,
                    then actions $\ac_{i : i + \epw - 1}$, and then arbitrary actions.
                }
                \State{
                    $\hat{\reward}_i(\state_i, \ac_{i : i + \epw - 1}) \gets$
                    $\sum_{\eptime = i}^{i + \epw - 1} \reward(\state_\eptime, \ac_\eptime)$.
                }
            \EndFor
            \State{
                \label{line:epw_max_action_seq}
                $\policy_i(\state_i) \gets \arg \max_{\ac_i \in \acspace}$
                $\max_{\ac_{i + 1 : i + \epw - 1} \in \acspace^{\numqvi - 1}} \hat{\reward}_i(\state_i, \ac_{i : i + \epw - 1}).$
            } 
        \EndFor
		\State \Return $\policy$
    \EndProcedure
\end{algorithmic}
Again, this algorithm is quite similar to \algac (Algorithm \ref{alg:qirl}) except that it only samples a single episode per action sequence and it ignores rewards beyond the planning window. Clearly, \textsc{PlanOverWindow} will take $\horizon^2 \acsize^\epw$ steps in the environment. Thus, to bound the sample complexity, we only need to show it returns an optimal policy with probability at least $1/2$.

To prove this, we will show that
\begin{equation}
    \label{eq:q_w_eq_max}
    \max_{\ac_{i + 1 : i + \epw - 1} \in \acspace^{\numqvi - 1}} \hat{\reward}_i(\state_i, \ac_{i : i + \epw - 1}) = Q^\epw_i(\state_i, \ac_i).
\end{equation}
Based on line \ref{line:epw_max_action_seq} of the algorithm, this is enough to show that $\policy \in \policies(Q^\epw)$, and thus that $\policy$ must be optimal by Definition \ref{defn:epw}.

To prove (\ref{eq:q_w_eq_max}), we will first show by induction that $Q^j_\eptime(\state_\eptime, \ac_\eptime) = \max_{\ac_{\eptime + 1 : \eptime + j - 1} \in \acspace^{j - 1}} \sum_{\eptime' = \eptime}^{\eptime + j - 1} \reward(\state_{\eptime'}, \ac_{\eptime'})$, where $\state_{\eptime' + 1} = \dynamics(\state_{\eptime'}, \ac_{\eptime'})$ for $\eptime' = i, \hdots, i + \epw - 2$. The base case when $j = 1$ is by definition: $Q^1_\eptime(\state_\eptime, \ac_\eptime) = \reward(\state_\eptime, \ac_\eptime)$. For the inductive step, assume the formula holds for $j$ and note that 
\begin{align*}
    Q^{j + 1}_\eptime(\state_\eptime, \ac_\eptime)
    & = \qvi(Q^j_\eptime)(\state_\eptime, \ac_\eptime) \\
    & = \reward(\state_\eptime, \ac_\eptime) + \max_{\ac_{\eptime + 1} \in \acspace} Q^j_{\eptime + 1}(\dynamics(\state_\eptime, \ac_\eptime), \ac_{\eptime + 1}) \\
    & = \reward(\state_\eptime, \ac_\eptime) + \max_{\ac_{\eptime + 1} \in \acspace} \max_{\ac_{\eptime + 2 : \eptime + j} \in \acspace^{j - 1}} \sum_{\eptime' = \eptime + 1}^{\eptime + j} \reward(\state_{\eptime'}, \ac_{\eptime'}) \\
    & = \max_{\ac_{\eptime + 1 : \eptime + j} \in \acspace^{j}} \sum_{\eptime' = \eptime}^{\eptime + j} \reward(\state_{\eptime'}, \ac_{\eptime'}).
\end{align*}

Next, note that by the way $\hat{\reward}$ is constructed, $\hat{\reward}(\state_i, \ac_{i : i + \epw - 1}) = \sum_{\eptime = i}^{i + \epw - 1} \reward(\state_\eptime, \ac_\eptime)$, where $\state_{\eptime + 1} = \dynamics(\state_\eptime, \ac_\eptime)$ for $\eptime = i, \hdots, i + \epw - 2$. Thus combining this with the formula proved by induction, (\ref{eq:q_w_eq_max}) clearly holds and the proof is complete.
\end{proof}

\subsection{Other bounds}
\label{sec:other_bounds}

One other work that derives sample complexity bounds for RL with random exploration is \citet{dann_guarantees_2022}. They define an algorithm which maintains at all times a current best policy $\policy$, and acts according to this policy but with some exploration noise, e.g., via an $\epsilon$-greedy policy $\text{expl}_\epsilon(\policy)$. They introduce the notion of a ``myopic exploration gap,'' which is defined as
\begin{gather*}
    \alpha = \sup_{\policy', c \geq 1} \frac{1}{\sqrt{c}} \left( \pret(\policy') - \pret(\policy) \right) \\
    \text{such that for all $(\eptime, \state, \ac) \in [\horizon] \times \statespace \times \acspace$} \\
    \mu_\eptime^{\policy'} \leq c \mu_{\eptime}^{\text{expl}_\epsilon(\policy)}(\state, \ac) \\
    \mu_\eptime^{\policy} \leq c \mu_{\eptime}^{\text{expl}_\epsilon(\policy)}(\state, \ac).
\end{gather*}
This gap is shown to generalize the notion of covering length as well as various others from the literature. However, we find that it is not so useful in many environments in \datasetname.

The problem we find is illustrated in the MDP below:

\begin{tikzpicture}[auto,node distance=6mm,scale=0.8,>=latex,font=\small]
    \tikzstyle{state}=[thick,draw=black,circle]

    \node[state] at (0, 0) (s1) {$\state_1$};
    \node[state] at (-1.5, -1) (s2_1) {$\state_2$};
    \draw[->] (s1) -- node[above left] {$\reward = 1$} (s2_1);
    \node at (0, -1) (s2_ellipsis) {$\dots$};
    \draw[->] (s1) -- node[right] {$0$} (s2_ellipsis);
    \node[state] at (1.5, -1) (s2_2) {$\state_2$};
    \draw[->] (s1) -- node[above right] {$0$} (s2_2);

    \node at (-2, -2) (s3_1) {$\dots$};
    \draw[->] (s2_1) -- node[above left] {$\reward = 1$} (s3_1);
    \node at (-1, -2) (s3_2) {$\dots$};
    \draw[->] (s2_1) -- node[above right] {$0$} (s3_2);
    \node at (1, -2) (s3_3) {$\dots$};
    \draw[->] (s2_2) -- node[above left] {$0$} (s3_3);
    \node at (2, -2) (s3_4) {$\dots$};
    \draw[->] (s2_2) -- node[above right] {$0$} (s3_4);

    \node[state] at (-2, -3) (sT_1) {$\state_\horizon$};
    \draw[->] (s3_1) -- node[left] {$\reward = 1$} (sT_1);
    \node[state] at (-1, -3) (sT_2) {$\state_\horizon$};
    \draw[->] (s3_2) -- node[right] {$0$} (sT_2);
    \node[state] at (1, -3) (sT_3) {$\state_\horizon$};
    \draw[->] (s3_3) -- node[left] {$0$} (sT_3);
    \node[state] at (2, -3) (sT_4) {$\state_\horizon$};
    \draw[->] (s3_4) -- node[right] {$\horizon - 1$} (sT_4);
\end{tikzpicture}

In this MDP, one need simply follow the actions which give rewards of 1 to achieve the optimal return of $\horizon$. One can show $\efhorizon{} = \epw = 1$ in this MDP, which give identical sample complexities of $\horizon^2 \acsize$.

However, the difficulty with the myopic exploration gap is that the analysis in \citet{dann_guarantees_2022} cannot rule out the policy $\policy$ which takes all actions to the right (achieving return $\horizon - 1$) from being chosen at some point while running their RL algorithm. If this happens, then the only way to find a better policy is to completely switch to the policy $\policy'$ which takes all left actions (achieving return $\horizon$). This implies that $\alpha$ is maximized when $c = \acsize^\horizon$, leading to $\alpha = (1 / \acsize)^{\horizon / 2}$. Since the sample complexity bounds in \citet{dann_guarantees_2022} are $O(1 / \alpha^2)$, this gives a bound proportional to $\acsize^\horizon$, which is no better than the worst case.

Thus, whenever there are ``distracting'' rewards, no matter how distant, as in this case, the theory from \citet{dann_guarantees_2022} cannot give good sample complexity bounds. There is also no easy way calculate $\alpha$ directly for an arbitrary environment. For these reasons, we do not include their bounds in our experiments.

\section{Dataset details}
\label{sec:dataset_details}

\begin{wrapfigure}{R}{2.5in}
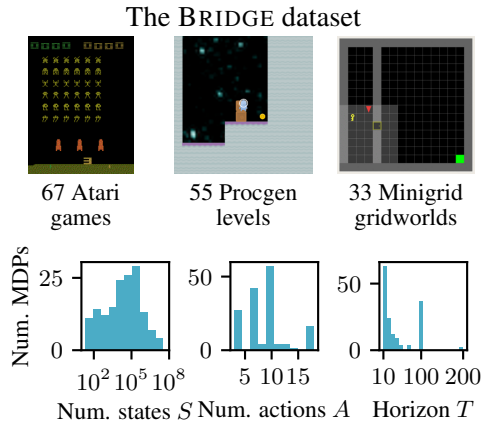

    \centering
    \input{figures/dataset_1.pgf}
    \input{figures/dataset_2.pgf}
    \caption{Our \datasetname dataset consists of 155 deterministic MDPs with full tabular representations. We include MDPs from three popular RL benchmarks which cover a range of state space sizes, action state sizes, and horizons.}
    \label{fig:dataset_overview}
    \vspace{-18pt}
\end{wrapfigure}

In this appendix, we give a detailed explanation of how we chose the MDPs in \datasetname and how we constructed their tabular representations. See Figure \ref{fig:dataset_overview} for an overview of \datasetname.

\subsection{Environments}
\label{sec:dataset_environments}

We limited the horizon of environments for \datasetname to some $\horizon \in \{10, 15, 20, 30, 50, 70, 100, 200\}$, depending on the environment, in order to avoid the state space becoming intractably large. We use subscripts to denote the horizon to which an environment is limited. For instance, $\textsc{Pong}_{50}$ refers to the Atari game Pong limited to 50 timesteps.

\smallparagraph{Frameskip} We carefully used frameskip for each environment. Frameskip is a standard practice in Atari \citep{braylan_frame_2015} in which each action taken in the environment is played for a certain number of frames; the agent only receives the next state after all these frames have completed. We use unusually high frameskips in order to capture episodes with longer wall-clock times in a small number of environment timesteps. The frameskip values we use are listed in Table \ref{tab:frameskips}. For most Atari games, we use a frameskip of 30, corresponding to taking 2 actions per second. The frameskips for Procgen environments vary; we chose ones that tended to align with how long it took the agent to perform various low-level tasks in the environment like moving one space. We did not use frameskip for Minigrid.

\begin{table}[H]
    \centering
    \begin{tabular}{l|r}
        \toprule
        Environment & Frameskip \\
        \midrule
        \textsc{MontezumaRevenge} & 24 \\
        All other Atari games & 30 \\
        \midrule
        \textsc{Bigfish} & 8 \\
        \textsc{Chaser} & 2 \\
        \textsc{Climber} & 6 \\
        \textsc{Coinrun} & 8 \\
        \textsc{Dodgeball} & 8 \\
        \textsc{Fruitbot} & 8 \\
        \textsc{Heist} & 2 \\
        \textsc{Jumper} & 8 \\
        \textsc{Leaper} & 6 \\
        \textsc{Maze} & 1 \\
        \textsc{Miner} & 1 \\
        \textsc{Ninja} & 8 \\
        \textsc{Plunder} & 8 \\
        \textsc{Starpilot} & 8 \\
        \midrule
        All Minigrid gridworlds & 1 \\
        \bottomrule
    \end{tabular}
    \vspace{6pt}
    \caption{Frameskip values used for the MDPs in \datasetname.}
    \label{tab:frameskips}
\end{table}

\smallparagraph{Atari games} For each of the 57 Atari games in the Arcade Learning Environment (ALE) benchmark \citep{bellemare_arcade_2013}, we attempted to construct tabular representations for each horizon $\horizon \in \{10, 20, 30, 50, 70, 100, 200\}$. However, we excluded environments once the state space exceeded 100 million states. We kept multiple horizon-limited versions of games, i.e., \datasetname contains $\textsc{Pong}_{10}$, $\textsc{Pong}_{20}$, $\textsc{Pong}_{30}$, etc. For some games, even $\horizon = 10$ produced too many states, so we did not include these at all. We use the minimal action sets for each MDP rather than all 18 possible Atari actions.

We made one exception to these procedures for Montezuma's Revenge, as it is an environment well-known for being difficult to explore, so we wanted to make sure to include it in \datasetname. We found that with $\horizon = 10$, there was not enough time to get any reward, and with $\horizon = 20$ there were too many states. We found that using $\horizon = 15$ and a frameskip of 24 did allow an agent to receive reward, so we used this version in \datasetname.

We made a couple other modifications to the standard Atari setup. First, we limited agents to one life: as soon as a life is lost, the episode ends. Second, in $\textsc{Skiing}_{10}$, we added an additional 200 frames of \textsc{Noop} actions after the 10 timesteps ($=300$ frames) in each episode. This is necessary to correctly reflect the reward incentives in $\textsc{Skiing}$ with longer horizons.

Finally, we scaled the rewards for many Atari games to make the reward scale more uniform across different games. Often, when deep RL is applied to Atari, rewards are \emph{clipped} to $[-1, 1]$ to avoid instability. However, the MDP with clipped rewards may have a different optimal policy than the unclipped MDP. Thus, instead of clipping, we use scaling. We generally choose the scale factor based on the multiples of points received in the game: for instance, in \textsc{Atlantis}, rewards are always received in multiples of 100 so we scale by $1/100$. Table \ref{tab:reward_factors} lists the reward scaling factors for all games where we apply scaling.

\begin{table}[H]
    \centering
    \begin{tabular}{l|r}
        \toprule
        Game & Reward scaling factor \\
        \midrule
        \textsc{Alien} & $1 / 10$ \\
        \textsc{Amidar} & $1 / 10$ \\
        \textsc{Assault} & $1 / 21$ \\
        \textsc{Asterix} & $1 / 50$ \\
        \textsc{Asteroids} & $1 / 10$ \\
        \textsc{Atlantis} & $1 / 100$ \\
        \textsc{BankHeist} & $1 / 10$ \\
        \textsc{BattleZone} & $1 / 1000$ \\
        \textsc{BeamRider} & $1 / 44$ \\
        \textsc{Centipede} & $1 / 100$ \\
        \textsc{ChopperCommand} & $1 / 100$ \\
        \textsc{CrazyClimber} & $1 / 100$ \\
        \textsc{DemonAttack} & $1 / 10$ \\
        \textsc{Frostbite} & $1 / 10$ \\
        \textsc{Gopher} & $1 / 20$ \\
        \textsc{Hero} & $1 / 25$ \\
        \textsc{Kangaroo} & $1 / 100$ \\
        \textsc{MontezumaRevenge} & $1 / 100$ \\
        \textsc{MsPacman} & $1 / 10$ \\
        \textsc{NameThisGame} & $1 / 10$ \\
        \textsc{Phoenix} & $1 / 20$ \\
        \textsc{PrivateEye} & $1 / 100$ \\
        \textsc{Qbert} & $1 / 25$ \\
        \textsc{RoadRunner} & $1 / 100$ \\
        \textsc{Qeaquest} & $1 / 20$ \\
        \textsc{Skiing} & $1 / 100$ \\
        \textsc{SpaceInvaders} & $1 / 5$ \\
        \textsc{TimePilot} & $1 / 100$ \\
        \textsc{VideoPinball} & $1 / 100$ \\
        \textsc{WizardOfWor} & $1 / 100$ \\
        \bottomrule
    \end{tabular}
    \vspace{6pt}
    \caption{Factors by which Atari games rewards are scaled by in \datasetname, for those where we apply reward scaling.}
    \label{tab:reward_factors}
\end{table}

\smallparagraph{Procgen levels} The Procgen benchmark \citep{cobbe_leveraging_2020} consists of 16 games. For each game, one can generate an arbitrary number of random levels, each of which is identified by a seed. Furthermore, each game has an ``easy'' and ``hard'' difficulty, each with different levels, and some have an additional ``exploration'' level which presents a particularly difficult exploration challenge.

While the benchmark is designed to measure generalization of RL agents trained on some number of levels to unseen levels, we use each level as a separate MDP. For each game, we attempted to construct an MDP for the easy levels with seeds $0$, $1$, and $2$, the hard level with seed $0$, and the exploration level if it exists for that environment. We denote $\textsc{Maze}_{30}^\text{E1}$ to be the easy level with seed $1$ for the \textsc{Maze} game limited to $\horizon = 30$ timesteps; $\textsc{Maze}_{30}^\text{H0}$ is the analogous hard level with seed $0$ and $\textsc{Maze}_{30}^\text{EX}$ is the exploration level.

We generally increased the horizon for each game to the highest value in $\{10, 20, 30, 40, 50, 70, 100, 200\}$ before the number of states was greater than 100 million. The horizon values we ultimately chose can be seen in the table in Appendix \ref{sec:mdp_list}.

\smallparagraph{Minigrid gridworlds} Minigrid \citep{chevalier-boisvert_minimalistic_2018} is an extensible framework for building gridworlds. We considered all the pre-built gridworlds included in Minigrid for inclusion in \datasetname except for those requiring natural language observations for specifying the task to be completed. We also excluded gridworlds with more than 1 million states, since for technical reasons we were unable to parallelize the construction of tabular MDPs for Minigrid. For gridworlds with randomized start states, we chose the start state with seed 0. We use $\horizon = 100$ for all the gridworlds.

\subsection{Constructing tabular representations}
\label{sec:tabular_representations}

For each of the environments described above, we wrote a program to compute a full tabular representation of the transition function $\dynamics$ and reward function $\reward$. Our program uses a search procedure to iteratively explore every state-action pair. We keep a queue of states that need to be explored, which at first is just the initial state. In parallel, a number of worker threads take states from this queue. After popping a state, a worker thread sets the environment to that state and then takes a previously unexplored action, storing the resulting next state and reward. If there are still unexplored actions in the current state, it adds it back to the queue. If the next state is not terminal and has not had all its actions, it can continue this process.

While the search procedure for exhaustively enumerating states is conceptually simple, we experienced difficulties implementing it efficiently due to the massive scale of some of the MDPs in \datasetname. For instance, the full state of the Atari simulator used in the ALE is about 10-12 KB of data. Storing 100 million states by themselves would thus require over one terabye of memory! We avoided this problem by aggressively compressing state data using dictionary compression. Other challenges included efficiently parallelizing the data structures we used to store the queue of states, the transition function, and the reward function. Our final implementation is able to explore more than 20,000 state-action pairs per second in $\textsc{Pong}$ while running on 64 cores.

Once we have enumerated all states and actions that can be reached in the given horizon, we also apply a consolidation step to reduce the number of states. Often, the internal representation of states in the Atari and Procgen environments includes extra or superfluous data, which leads to duplicate states in our tabular representation. We repeatedly consolidate states that (a) have the same screen, (b) have the same rewards for each action, and (c) lead to the same next states for each action. When no more states can be consolidated, we store the resulting transition and reward functions.

We excluded any MDPs for which every sequence of actions results in the same total reward, since these are uninteresting from an RL perspective.

\subsection{Reward shaping}
\label{sec:reward_shaping}

For each Minigrid environment, we constructed one or more versions with shaped rewards for our experiments on the effects of reward shaping. We used three potential functions for shaping:
\begin{enumerate}
    \item $\Phi_\text{dist}(\state)$: the negative distance from the state to the nearest goal. Distance is measured as the minimum number of moves needed to reach the goal, assuming there are no obstacles in the way.
    \item $\Phi_\text{doors}(\state)$: the number of doors that are open.
    \item $\Phi_\text{pickup}(\state)$: the number of objects that have been picked up at least once.
\end{enumerate}
For one or more potential functions $\Phi$, we augment each reward $\reward(\state, \ac)$ with $\Phi(\dynamics(\state, \ac)) - \Phi(\state)$. The potential functions are chosen to incentivize useful behavior in the environments: moving towards goals, picking up objects like keys that could be helpful, and opening doors to reach more parts of the gridworld.

For each Minigrid MDP, we use all potential functions that apply to that MDP. For instance, if an MDP does not have any doors, we do not use $\Phi_\text{doors}$. We also apply the combination of $\Phi_\text{doors}$ and $\Phi_\text{pickup}$ if both are applicable to an MDP.

When analysing the reward shaping results, we only include MDPs for which PPO/DQN converged on both the unshaped and shaped versions.

\subsection{Datasheet for \datasetname}

We provide a datasheet, as proposed by \citet{gebru_datasheets_2021}, for the \datasetname dataset.

\subsubsection{Motivation}

\smallparagraph{For what purpose was the dataset created?}
We have described the purpose extensively in the paper: we aim to bridge the theory-practice gap in RL. \datasetname allows this by providing tabular representations of popular deep RL benchmarks such that instance-dependent bounds can be calculated and compared to empirical RL performance.

\smallparagraph{Who created the dataset (e.g., which team, research group) and on behalf of which entity (e.g., company, institution, organization)?}
Not specified for the double-blind reviewing process.

\smallparagraph{Who funded the creation of the dataset?}
Also not specified for the double-blind reviewing process.

\smallparagraph{Any other comments?}
No.

\subsubsection{Composition}

\smallparagraph{What do the instances that comprise the dataset represent (e.g., documents, photos, people, countries)?}
The instances are Markov Decision Processes (MDPs).

\smallparagraph{How many instances are there in total (of each type, if appropriate)?}
There are 155 MDPs in \datasetname. They include 67 MDPs based on Atari games from the Arcade Learning Environment \citep{bellemare_arcade_2013}, 55 MDPs based on Procgen games \citep{cobbe_leveraging_2020}, and 33 MDPs based on MiniGrid gridworlds \citep{chevalier-boisvert_minimalistic_2018}.

\smallparagraph{Does the dataset contain all possible instances or is it a sample (not necessarily random) of instances from a larger set?}
The MDPs in \datasetname are based on a small subset of the many environments that are used for empirically evaluating RL algorithms. We aimed to cover a range of the most popular environments. To make our analysis possible, we excluded environments that were not deterministic or did not have discrete action spaces. We also reduced the horizon of many of the environments to make it tractable to compute their tabular representations.

\smallparagraph{What data does each instance consist of?}
For each MDP, we provide the following data:
\begin{itemize}
    \item A transition function and a reward function, which are represented as a matrix with an entry for each state-action pair in the MDP.
    \item A corresponding \texttt{gym} environment \citep{brockman_openai_2016} that can be used to train policies for the MDP with various RL algorithms.
    \item Properties of the MDP that are calculated from its tabular representation, including the effective planning window, bounds on the effective horizon, bounds on the covering length, etc.
    \item Results of running RL algorithms (PPO, DQN, and \algac) on the MDP. This includes the empirical sample complexity as well as various metrics logged during training.
    \item For MiniGrid MDPs, there are additional versions of the MDP with shaped reward functions (see Appendix \ref{sec:reward_shaping}) which also include all of the above data.
    \item For Atari and Procgen MDPs, there is additionally a non-uniform exploration policy (see Appendix \ref{sec:exploration_policies}). For Atari games, this is trained via behavior cloning from the Atari-HEAD \citep{zhang_atari-head_2019} dataset; for Procgen games, it is trained on other Procgen levels. We include the above data recalculated using the non-uniform exploration policy in place of the uniformly random exploration policy.
\end{itemize}

\smallparagraph{Is there a label or target associated with each instance?}
In this paper, we aim to bound and/or estimate the empirical sample complexity of RL algorithms, so these could be considered targets for each instance.

\smallparagraph{Is any information missing from individual instances?}
There is no information missing.

\smallparagraph{Are relationships between individual instances made explicit (e.g., users’ movie ratings, social network links)?}
No.

\smallparagraph{Are there recommended data splits (e.g., training, development/validation, testing)?}
No.

\smallparagraph{Are there any errors, sources of noise, or redundancies in the dataset?}
We do not believe there are errors or sources of noise in the dataset. The tabular representations of the MDPs have been carefully tested for correspondence with the environments they are based on. There is some redundancy, as many Atari games are represented more than once with varying horizons.

\smallparagraph{Is the dataset self-contained, or does it link to or otherwise rely on external resources (e.g., websites, tweets, other datasets)?}
The dataset is mostly self-contained, except that the \texttt{gym} environments rely on external libraries. There are archival versions of these available through package managers like PyPI.

\smallparagraph{Does the dataset contain data that might be considered confidential (e.g., data that is protected by legal privilege or by doctor–
patient confidentiality, data that includes the content of individuals’ non-public communications)?}
No.

\smallparagraph{Does the dataset contain data that, if viewed directly, might be offensive, insulting, threatening, or might otherwise cause anxiety?}
No.

\subsubsection{Collection process}

\smallparagraph{How was the data associated with each instance acquired?}
The data was collected using open-source implementations of each environment.

\smallparagraph{What mechanisms or procedures were used to collect the data (e.g., hardware apparatuses or sensors, manual human curation, software programs, software APIs)?}
As described in Appendix \ref{sec:tabular_representations}, we developed a software tool to construct the tabular representations of the MDPs in \datasetname. We validated the correctness of the tabular MDPs through extensive testing to ensure they corresponded exactly with the \texttt{gym} implementations of the environments.

\smallparagraph{If the dataset is a sample from a larger set, what was the sampling strategy (e.g., deterministic, probabilistic with specific sampling probabilities)?}
The MDPs in \datasetname were selected from three collections of commonly used RL environments: the Arcade Learning Environment, ProcGen, and MiniGrid. We chose these three collections to represent a broad set of deterministic environments with discrete action spaces. Within each collection, the environments were further filtered based on the criteria described in Appendix \ref{sec:dataset_environments}.

\smallparagraph{Who was involved in the data collection process (e.g., students, crowdworkers, contractors) and how were they compensated (e.g., how much were crowdworkers paid)?}
Only the authors were involved in the data collection process.

\smallparagraph{Over what timeframe was the data collected?} The dataset was assembled between February 2022 and January 2023. The RL environments from which the MDPs in \datasetname were constructed were created prior to this; see the cited works for each collection of environments for more details.

\smallparagraph{Were any ethical review processes conducted (e.g., by an institutional review board)?} No.

\subsubsection{Preprocessing/cleaning/labeling}

\smallparagraph{Was any preprocessing/cleaning/labeling of the data done (e.g., discretization or bucketing, tokenization, part-of-speech tagging, SIFT feature extraction, removal of instances, processing of missing values)?}
Yes, various preprocessing and analysis was done. See Appendix \ref{sec:tabular_representations} for details.

\smallparagraph{Was the “raw” data saved in addition to the preprocessed/cleaned/labeled data (e.g., to support unanticipated future uses)?} Yes, this is included with the dataset.

\smallparagraph{Is the software that was used to preprocess/clean/label the data available?} Yes, this is available with the rest of our code.

\smallparagraph{Any other comments?} No.

\subsubsection{Uses}

\smallparagraph{Has the dataset been used for any tasks already?}
The dataset has thus far only been used to validate our theory of the effective horizon in this paper.

\smallparagraph{Is there a repository that links to any or all papers or systems that use the dataset?}
There is not. However, we will require that any uses of the dataset cite this paper, allowing one to use tools like Semantic Scholar or Google Scholar to find other papers which use the \datasetname dataset.

\smallparagraph{What (other) tasks could the dataset be used for?}
We hope that the \datasetname dataset is used for further efforts to bridge the theory-practice gap in RL. The dataset could be used to identify other properties or assumptions that hold in common environments, or to calculate instance-dependent sample complexity bounds and compare them to the empirical sample complexity of RL algorithms.

\smallparagraph{Is there anything about the composition of the dataset or the way it was collected and preprocessed/cleaned/labeled that might impact future uses?}
As we have already mentioned, \datasetname is restricted to deterministic MDPs with discrete action spaces and relatively short horizons. This could mean that analyses of the dataset like ours do not generalize to the broader space of RL environments that may have continuous action spaces, stochastic transitions, and/or long horizons. We have included some experiments, like those in Appendix \ref{sec:full_atari}, to show that our theory of the effective horizon generalizes beyond the MDPs in \datasetname. We encourage others to do the same and we hope to address some of these limitations in the future with extensions to \datasetname.

\smallparagraph{Are there tasks for which the dataset should not be used?}
We do not foresee any particular tasks for which the dataset should not be used.

\smallparagraph{Any other comments?}
No.

\subsubsection{Distribution}

\smallparagraph{Will the dataset be distributed to third parties outside of the entity (e.g., company, institution, organization) on behalf of which
the dataset was created?}
Yes, we will distribute the dataset publicly.

\smallparagraph{How will the dataset will be distributed (e.g., tarball on website, API, GitHub)?}
We are still finalizing the method through which the dataset will be distributed.

\smallparagraph{When will the dataset be distributed?}
We plan to make the dataset public in May or June 2023.

\smallparagraph{Will the dataset be distributed under a copyright or other intellectual property (IP) license, and/or under applicable terms of use
(ToU)?}
It will be distributed under CC-BY-4.0.

\smallparagraph{Have any third parties imposed IP-based or other restrictions on the data associated with the instances?}
The Atari ROMs used to construct the Atari MDPs in \datasetname are copyrighted by the original creators of the games. However, they are widely used throughout the reinforcement learning literature and to our knowledge the copyright holders have not complained about this. Since we are not legal experts, we do not know if releasing our dataset violates their copyright, but we do not believe that we are harming them since the tabular representations in \datasetname are only useful for research purposes and cannot be used to play the games in any meaningful way.

\smallparagraph{Do any export controls or other regulatory restrictions apply to the dataset or to individual instances?}
No.

\smallparagraph{Any other comments?}
No.

\subsubsection{Maintenance}

\smallparagraph{Who will be supporting/hosting/maintaining the dataset?}
We (the authors) will support and maintain the dataset.

\smallparagraph{How can the owner/curator/manager of the dataset be contacted
(e.g., email address)?}
Redacted for double-blind review.

\smallparagraph{Is there an erratum?}
We will record reports of any errors in the dataset and release new versions with descriptions of what was fixed as necessary.

\smallparagraph{Will the dataset be updated (e.g., to correct labeling errors, add new instances, delete instances)?}
We will release new versions of the dataset to correct any reported errors as described above. We may also expand the dataset in the future with more MDPs or new kinds of MDPs, such as stochastic or continuous-action-space MDPs. Any updates will be communicated through the service we use to host the dataset (TBD). 

\smallparagraph{If the dataset relates to people, are there applicable limits on the retention of the data associated with the instances (e.g., were the individuals in question told that their data would be retained for a fixed period of time and then deleted)?} No.

\smallparagraph{Will older versions of the dataset continue to be supported/hosted/maintained? If so, please describe how. If not, please describe how its obsolescence will be communicated to dataset consumers.}
We hope to find a host for the dataset that will retain older versions of the dataset. We only plan to maintain the latest version of the dataset, however. We will note this policy in the dataset's description.

\smallparagraph{If others want to extend/augment/build on/contribute to the dataset, is there a mechanism for them to do so?}
There is no predefined mechanism to contribute to the dataset, but we will consider external contributions on a case-by-case basis. We encourage others to extend and build on the dataset.

\smallparagraph{Any other comments?}
No.

\section{Experiment details}
\label{sec:experiment_details}

In this appendix, we describe details of the experiments from Section \ref{sec:experiments}. In particular, we describe how we calculate the empirical sample complexity of PPO and DQN.

We use the implementations of PPO and DQN from Stable-Baselines3 (SB3) \citep{raffin_stable-baselines3_2021}. For the network archictures, we use convolutional neural nets (CNNs) for Atari and Procgen and a fully-connected network for Minigrid. The network architectures are the default CNN and fully-connected architectures chosen by SB3.
We used the below hyperparameters for PPO and DQN, which are mainly taken from the tuned Atari hyperparameters in the RL Baselines3 Zoo \citep{raffin_rl_2020} repository.

We used a discount rate of $\gamma = 1$ for all environments and algorithms except for PPO and DQN in MiniGrid environments, where we used $\gamma = 0.99$. We found that the performance of PPO and DQN significantly degraded when we used $\gamma = 1$ for MiniGrid.

\smallparagraph{PPO} We use the following hyperparameters for PPO:

\begin{table}[H]
    \centering
    \begin{tabular}{l|r}
        \toprule
        Hyperparameter & Value \\
        \midrule
        Training timesteps & 5,000,000 \\
        Number of environments & 8 \\
        Number of steps per rollout & $\{128, 1280\}$ \\
        Clipping parameter ($\epsilon$) & 0.1 \\
        Value function coefficient & 0.5 \\
        Entropy coefficient & 0.01 \\
        Optimizer & Adam \\
        Learning rate & $2.5 \times 10^{-4}$ \\
        Number of epochs per training batch & 4 \\
        Minibatch size & 256 \\
        GAE coefficient ($\lambda$) & 0.95 \\
        Advantage normalization & Yes \\
        Gradient clipping & 0.5 \\
        \bottomrule
    \end{tabular}
    \caption{Hyperparameters we use for PPO.}
    \label{tab:ppo_params}
\end{table}

For each environment, we try rollout lengths of 128 and 1,280, as we find this is the most sensitive hyperparameter to tune.

\smallparagraph{DQN} We use the following hyperparameters for DQN:


\begin{table}[H]
    \centering
    \begin{tabular}{l|r}
        \toprule
        Hyperparameter & Value \\
        \midrule
        Training timesteps & 5,000,000 \\
        Timesteps before learning starts & 0 \\
        Replay buffer size & 100,000 \\
        Target network update interval & 1,000 \\
        Training frequency & 4 \\ 
        Gradient steps per training step & 1 \\
        Optimizer & Adam \\
        Learning rate & $10^{-4}$ \\
        Exploration fraction & $\{0.1, 1\}$ \\
        Final $\epsilon$ & 0.01 \\
        Learning rate & $10^{-3}$ \\
        Gradient clipping & 10 \\
        \bottomrule
    \end{tabular}
    \caption{Hyperparameters we use for DQN.}
    \label{tab:dqn_params}
\end{table}

We try decaying the $\epsilon$ value for $\epsilon$-greedy over the course of either 500 thousand or 5 million timesteps, as we found this was the most sensitive hyperparameter to tune for DQN.

\smallparagraph{Calculating the empirical sample complexity}
In order to compute the empirical sample complexities of PPO and DQN, throughout training we run evaluation episodes and see if the algorithms have discovered an optimal policy yet. During the evaluation, PPO policies take actions according to the argmax over the probabilities they assign to each action, rather than sampling as during training episodes. DQN takes actions greedily with respect to its current Q-function (i.e., with $\epsilon = 0$). If the total episode reward during the evaluation is the optimal return, then we terminate the training run and record the total number of timesteps interacted with the environment as the empirical sample complexity. We take the median sample complexity over 5 random seeds and then the minimum over all hyperparameter settings to get the final empirical sample complexity.

We found that in some environments PPO achieved optimal reward during almost all the training episodes but none of the evaluation episodes. This can happen if a policy does not assign the highest probability to an optimal action in some states but can make up for this by being very likely overall to obtain the highest possible total reward. Thus, if more than half of the training episodes during an iteration achieve the optimal return, we also count this as converging to an optimal policy for the purposes of calculating the empirical sample complexity.

\subsection{\algac vs. deep RL algorithms over longer horizons}
\label{sec:full_atari}

To show that \algac is not just effective over short horizons, we ran additional experiments comparing \algac, PPO, and DQN in more typical Atari benchmark environments with frameskip 4 and a horizon of $\horizon = 27,000$ (corresponding to a maximum of 30 minutes of gameplay). Similarly to the environments in \datasetname, we limit agents to a single life and make the environments entirely deterministic. The hyperparameters for PPO and DQN are identical to those given in Tables \ref{tab:ppo_params} and \ref{tab:dqn_params} except for the following changes: we train for 50 million timesteps; we use a discount rate of $\gamma = 0.99$; and, we set an entropy coefficient of 0.01 for PPO.

We use a training batch size of $10^4$ for PPO and decay $\epsilon$ for DQN over the course of the first 5 million timesteps. For \algac, we use $\numqvi = 1$ and tune $\numep{}$ for each environment.

\subsection{Exploration policies}
\label{sec:exploration_policies}

For the experiments in Section \ref{sec:intervention_experiments} we needed pre-trained policies to initialize PPO with; here, we describe the details of how we trained them. We used two different training methods: one for Atari environments and one for Procgen environments.

In the Atari environments, we trained policies via behavior cloning (BC), i.e., supervised learning, from human data in the Atari-HEAD dataset \citep{zhang_atari-head_2019}. We resampled the actions and processed the screen images from the dataset to align with the frameskip and observation preprocessing of our Atari environments. We trained a BC policy on each environment for 400 batches of 500 timesteps each. We used Adam with a learning rate of $10^{-3}$. We also added an entropy bonus to the loss function with a weight of $0.1$ to avoid the BC policy assigning very little weight to some actions. Our theoretical results in Section \ref{sec:variants} suggest that this should improve the sample complexity. Since not all Atari games are included in Atari-HEAD, we only used a subset for the experiments in Section \ref{sec:intervention_experiments}.

In the Procgen environments, we pre-trained policies on a set of levels not included in \datasetname. In particular, we trained a policy with PPO on 500 easy levels for 25 million timesteps, which in very similar to the methodology in \citet{cobbe_leveraging_2020}. We also use an entropy bonus in PPO with weight $0.1$ for the same reason as above.

Once we have the pre-trained policies, we compute tabular representations of them on the corresponding MDPs in \datasetname. To do so, we feed the observations for every state in the MDP through the pre-trained policy network and record the resulting action distribution. This allows us to compute $Q^{\expolicy}$ and thus obtain bounds on the effective horizon when using the pre-trained exploration policy.

\subsection{Computational resources}
\label{sec:compute}

For deep RL experiments, we used a mix of A100, A4000, and A6000 GPUs from Nvidia. We ran the algorithms either on separate GPUs or sometimes we ran multiple random seeds simultaneously on the same hardware. We used 1-8 CPU threads to run the RL environments. Using this setup, PPO and DQN generally took 2-8 hours to complete 5 million timesteps of training. We used early stopping when the algorithms found an optimal policy before 5 million timesteps, so the amount of compute per experiment was often less than this.

For constructing and analyzing the tabular MDPs in \datasetname, we used up to 128 CPU threads and 500 GB of memory. The amount of time necessary to construct and analyze the MDPs ranged from less than a minute to around 5 days.

\section{Additional experiment results}
\label{sec:additional_results}

Here, we present additional results from the experiments in Section \ref{sec:experiments}.

\subsection{Example of the effective horizon failing to predict generalization}
\label{sec:generalization_example}

As we described in the discussion, the effective horizon cannot model generalization across different states. For instance, in \textsc{Pong}-30 (Pong limited to 30 timesteps/15 seconds), the effective horizon gives a sample complexity of roughly 5 billion timesteps and empirically GORP takes over 80 million timesteps to converge to an optimal policy (Appendix \ref{sec:mdp_list}). However, both PPO and DQN converge in under 500,000 environment steps. We hypothesize this is because they are able to generalize the skill of hitting the ball across the multiple rounds of the Pong game, which the effective horizon cannot capture because it considers learning separately at every timestep.

\subsection{Additional plots}

\begin{figure}[H]
    \centering
    \input{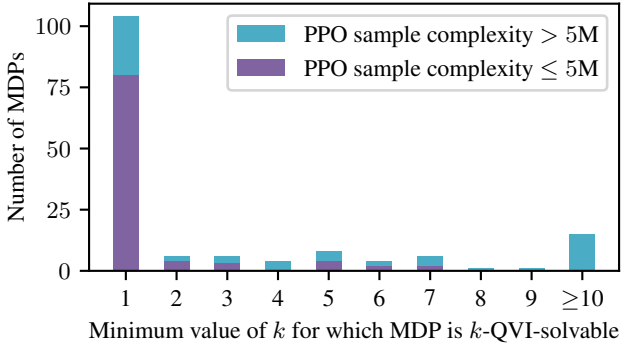}
    \caption{The distribution of the minimum values of $\numqvi$ for which the MDPs in \datasetname are $\numqvi$-QVI solvable. About two thirds are 1-QVI-solvable, meaning they can be solved by simply acting greedily with respect to the Q-function of the random policy. The MDPs are split into those which PPO can and cannot solve in 5 million steps; among those that can be solved efficiently, the values of $\numqvi$ are even lower.}
    \label{fig:numqvi_dist}
\end{figure}

\begin{figure}[H]
    \input{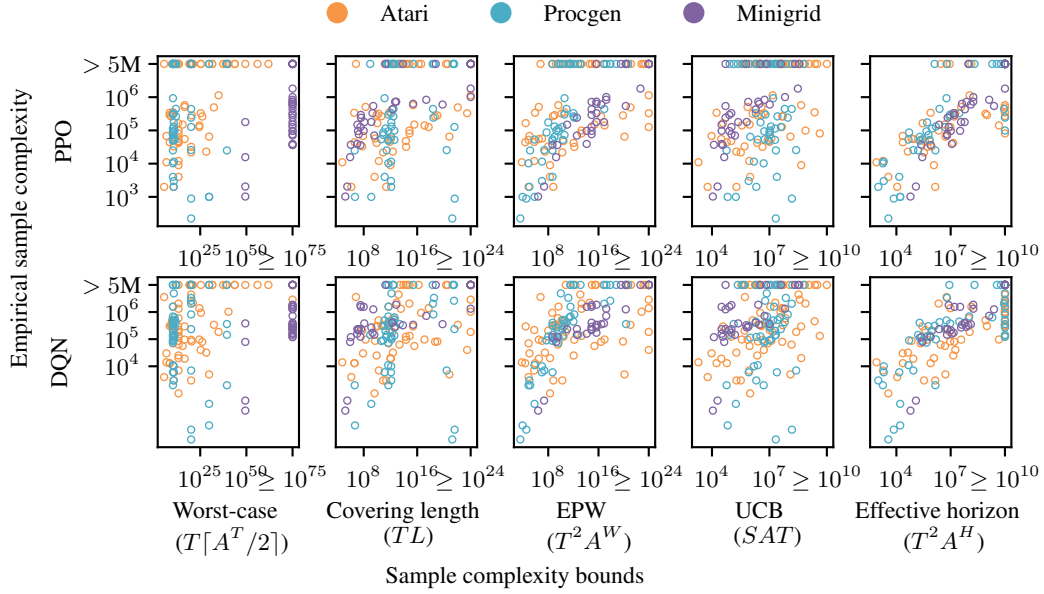}
    \caption{A comparison of sample complexity bounds and the empirical sample complexities of PPO and DQN across the MDPs in \datasetname. In each plot, every dot represents one MDP and its color indicates which benchmark it comes from. Our effective horizon-based bound most closely correlates with empirical sample complexity. See Table \ref{tab:bound_evaluations} for a quantitative comparison of the bounds.}
    \label{fig:bounds_vs_empirical}
\end{figure}

\begin{figure}[H]
    \centering
    \input{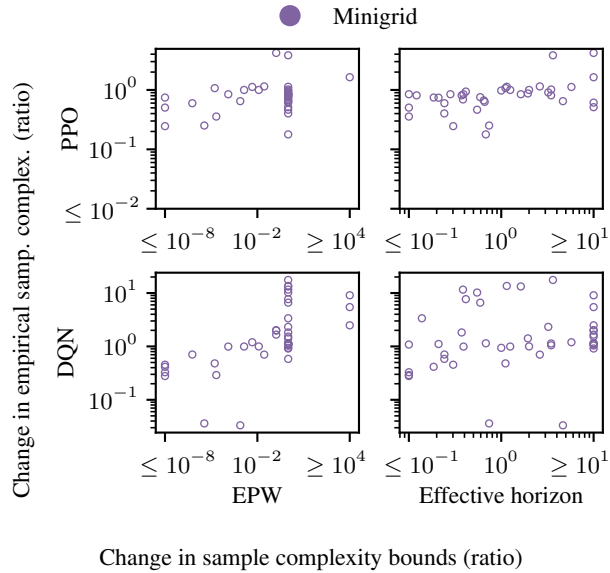}
    \caption{A comparison between the empirical change in sample complexity and the change predicted by sample complexity bounds due to reward shaping. See Table \ref{tab:reward_shaping} for a quantitative comparison.}
    \label{fig:reward_shaping}
\end{figure}

\begin{figure}[H]
    \centering
    \input{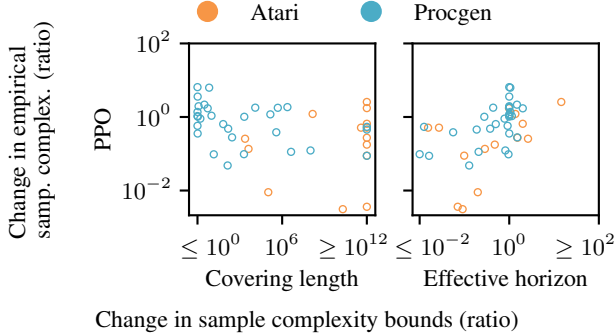}
    \caption{A comparison between the empirical change in the sample complexity of PPO and the change predicted by sample complexity bounds due to initializing with a pre-trained policy. The initial policies for Atari are trained from human data and those for Procgen are trained on other procedurally generated levels. See Table \ref{tab:exploration_policy} for a quantitative comparison.}
    \label{fig:exploration_policy}
\end{figure}

\subsection{List of MDPs in \datasetname with statistics}
\label{sec:mdp_list}

The following table lists all the MDPs in \datasetname, along with various properties: the number of states $\statesize$, number of actions $\acsize$, horizon $\horizon$, minimum $\numqvi$ for which the MDP is $\numqvi$-QVI-solvable, a bound on the effective horizon using the techniques in Appendix \ref{sec:tighter_bounds}, a bound on the covering length $\covlen$ using Lemma \ref{lemma:covlen_bounds}, and the effective planning window $\epw$.

{
\small
\begin{longtable}{l|rrr|rrrr}
\toprule
MDP & $\statesize$ & $\acsize$ & $\horizon$ & Min $\numqvi$ & Bound on $\efhorizon{}$ & Bound on $\covlen$ & $\epw$ \\
\midrule
$\textsc{Alien}_{10}$ & $7.97 \times 10^{5}$ & $18$ & $10$ & $1$ & $3.5$ & $2.27 \times 10^{12}$ & $6$ \\
$\textsc{Amidar}_{20}$ & $1.00 \times 10^{5}$ & $10$ & $20$ & $4$ & $12$ & $4.03 \times 10^{11}$ & $12$ \\
$\textsc{Assault}_{10}$ & $7.21 \times 10^{5}$ & $7$ & $10$ & $3$ & $9.8$ & $4.56 \times 10^{9}$ & $8$ \\
$\textsc{Asterix}_{10}$ & $9.03 \times 10^{4}$ & $9$ & $10$ & $1$ & $3.5$ & $4.99 \times 10^{10}$ & $3$ \\
$\textsc{Asteroids}_{10}$ & $5.11 \times 10^{6}$ & $14$ & $10$ & $7$ & $10$ & $2.72 \times 10^{12}$ & $9$ \\
$\textsc{Atlantis}_{10}$ & $49$ & $4$ & $10$ & $1$ & $1.5$ & $6.12 \times 10^{3}$ & $3$ \\
$\textsc{Atlantis}_{20}$ & $471$ & $4$ & $20$ & $1$ & $4.7$ & $5.53 \times 10^{8}$ & $3$ \\
\bottomrule
\end{longtable}
}

\subsection{Table of bounds and empirical sample complexities}

This table lists all the sample complexity bounds we calculate for each MDP in \datasetname along with the empirical sample complexities of PPO, DQN, and \algac.

{
\tiny \tabcolsep=3pt
\begin{longtable}{l|rrrrr|rrr}
\toprule
 & \multicolumn{5}{|c|}{\bf Sample complexity bounds} & \multicolumn{3}{|c}{\bf Empirical sample complexities} \\
MDP & \multicolumn{1}{|c}{Worst-case} & \multicolumn{1}{c}{Covering length} & \multicolumn{1}{c}{EPW} & \multicolumn{1}{c}{UCB} & \multicolumn{1}{c|}{Effective horizon} & & & \\
 & \multicolumn{1}{|c}{($\horizon \lceil \acsize^\horizon / 2 \rceil$)} & \multicolumn{1}{c}{($\horizon \covlen$)} & \multicolumn{1}{c}{($\horizon^2 \acsize^\epw$)} & \multicolumn{1}{c}{($\statesize \acsize \horizon$)} & \multicolumn{1}{c|}{($\horizon^2 \acsize^{\efhorizon{}}$)} & \multicolumn{1}{|c}{PPO} & \multicolumn{1}{c}{DQN} & \multicolumn{1}{c}{\algac} \\
\midrule
$\textsc{Alien}_{10}$ & $1.79 \times 10^{13}$ & $2.27 \times 10^{13}$ & $3.40 \times 10^{9}$ & $1.43 \times 10^{8}$ & $2.78 \times 10^{6}$ & $> 5 \times 10^6$ & $6.30 \times 10^{4}$ & $2.70 \times 10^{4}$ \\
$\textsc{Amidar}_{20}$ & $1.00 \times 10^{21}$ & $8.06 \times 10^{12}$ & $4.00 \times 10^{14}$ & $2.01 \times 10^{7}$ & $4.00 \times 10^{14}$ & $> 5 \times 10^6$ & $2.21 \times 10^{6}$ & $> 10^8$ \\
$\textsc{Assault}_{10}$ & $1.41 \times 10^{9}$ & $4.56 \times 10^{10}$ & $5.76 \times 10^{8}$ & $5.04 \times 10^{7}$ & $1.95 \times 10^{10}$ & $2.68 \times 10^{5}$ & $3.26 \times 10^{5}$ & $3.35 \times 10^{6}$ \\
$\textsc{Asterix}_{10}$ & $1.74 \times 10^{10}$ & $4.99 \times 10^{11}$ & $7.29 \times 10^{4}$ & $8.13 \times 10^{6}$ & $2.20 \times 10^{5}$ & $5.20 \times 10^{4}$ & $7.00 \times 10^{4}$ & $4.50 \times 10^{3}$ \\
$\textsc{Asteroids}_{10}$ & $1.45 \times 10^{12}$ & $2.72 \times 10^{13}$ & $2.07 \times 10^{12}$ & $7.16 \times 10^{8}$ & $2.89 \times 10^{13}$ & $> 5 \times 10^6$ & $> 5 \times 10^6$ & $> 10^8$ \\
\bottomrule
\end{longtable}
}

\subsection{Table of returns}

This table lists the returns for the random and optimal policies in each MDP as well as the final returns achieved by PPO, DQN, and \algac. We ran PPO and DQN for 5 million timesteps in each environment, and GORP for both 5 million and 100 million timesteps.

{
\small
\begin{longtable}{l|rr|rrrr}
\toprule
 & \multicolumn{5}{|c}{\bf Returns} \\
MDP & \multicolumn{1}{|c}{Random policy} & \multicolumn{1}{c}{Optimal policy} & \multicolumn{1}{|c}{PPO} & \multicolumn{1}{c}{DQN} & \multicolumn{1}{c}{\algac (5M)} & \multicolumn{1}{c}{(100M)} \\
\midrule
$\textsc{Alien}_{10}$ & $74.74$ & $160$ & $160$ & $160$ & $160$ & $160$ \\
$\textsc{Amidar}_{20}$ & $6.77$ & $109$ & $68$ & $68$ & $65$ & $68$ \\
$\textsc{Assault}_{10}$ & $8.66$ & $126$ & $126$ & $126$ & $126$ & $126$ \\
$\textsc{Asterix}_{10}$ & $99.24$ & $400$ & $400$ & $400$ & $400$ & $400$ \\
$\textsc{Asteroids}_{10}$ & $15.08$ & $320$ & $170$ & $190$ & $220$ & $220$ \\
$\textsc{Atlantis}_{10}$ & $24.22$ & $200$ & $200$ & $200$ & $200$ & $200$ \\
\bottomrule
\end{longtable}
}

\subsection{Learning curves for \datasetname MDPs}

\begin{figure}[H]
    \resizebox{\textwidth}{!}{\input{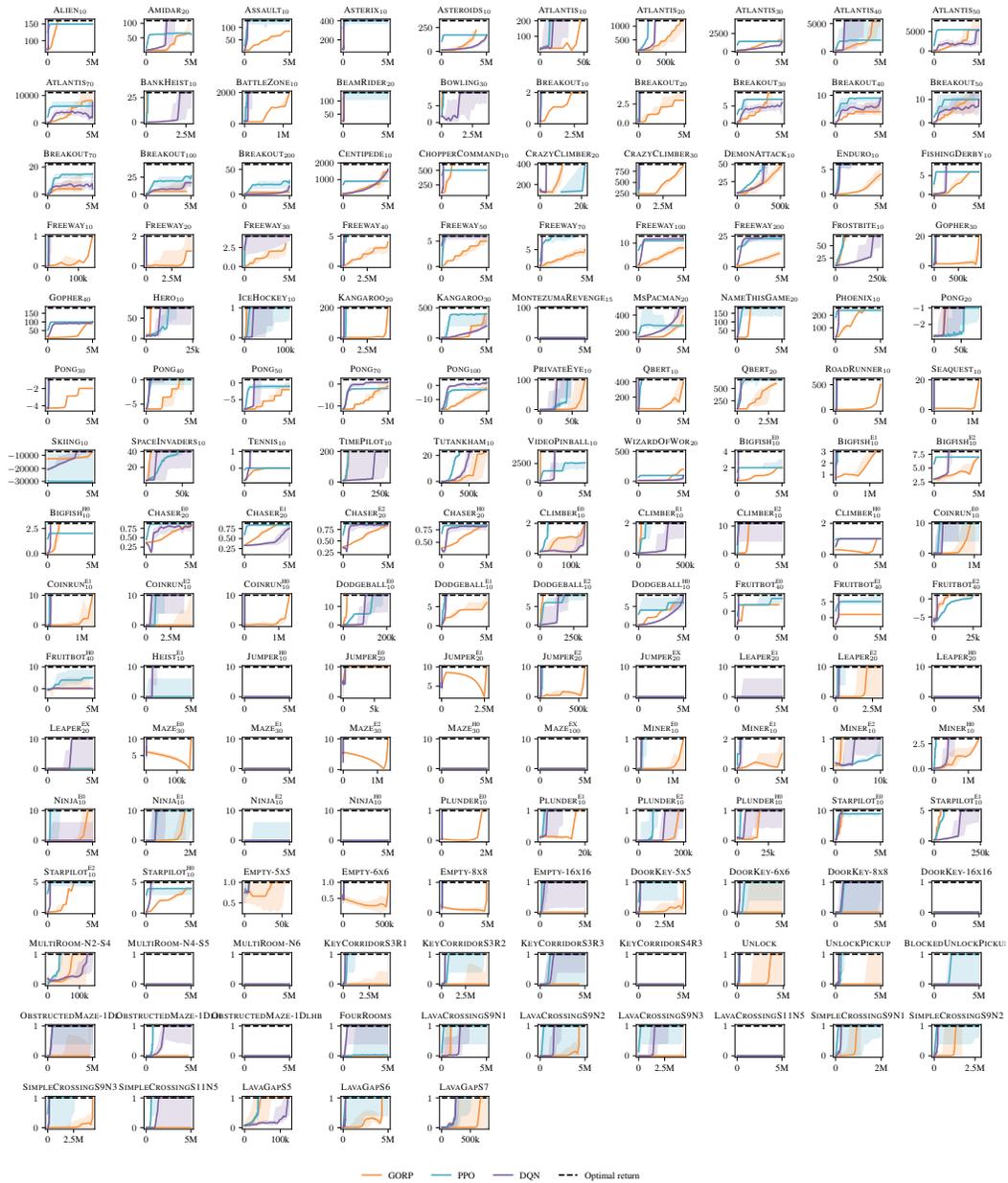}}
    \caption{Learning curves for PPO, DQN, and \algac on all the MDPs in \datasetname. Solid lines show the median return (over multiple random seeds) of the policies learned by each algorithm throughout training. We use 5 random seeds for PPO and DQN and 101 random seeds for \algac. The shaded region shows the range of returns over all random seeds for PPO and DQN, and shows the range from the 10th to 90th percentile over random seeds for \algac. The optimal return in each environment is shown as the dashed black line.}
\end{figure}

\end{document}